\pdfoutput=1

\documentclass[11pt]{article}
\usepackage[most]{tcolorbox}

\usepackage[preprint]{acl}

\usepackage{times}
\usepackage{latexsym}
 \usepackage{amsmath}
\usepackage{amssymb}
\usepackage{algorithm}
\usepackage{algpseudocode}
\usepackage{tabularx}
\usepackage{array}
\usepackage{placeins}
\usepackage{booktabs}
\usepackage[most]{tcolorbox}
\tcbuselibrary{breakable}
\usepackage{hwemoji}
\usepackage{threeparttable}

\newcommand{\CI}[2]{#1{\tiny$\pm$#2}}

 \usepackage[dvipsnames]{xcolor}
\definecolor{randomneg}{HTML}{3CB371} %
\definecolor{hardneg}{HTML}{1F77B4} %
\definecolor{correct}{HTML}{D62728} %
\usepackage[T1]{fontenc}

\usepackage[utf8]{inputenc}

\usepackage{microtype}
\usepackage{multirow}
\usepackage{inconsolata}
\usepackage{graphicx}
\usepackage{subcaption}

\usepackage{graphicx}
\usepackage{emo}
\title{SAGE \raisebox{-0.12em}{\scalebox{1.6}{^^f0^^9f^^8c^^bf}}: Sense-Aligned Multimodal Grounding Evaluation with Gloss-Augmented Reasoning}

\author{
  \textbf{Sunjae Kwon}$^{1}$,
  \textbf{Suhaas Kiran DG}$^{1}$, 
  \textbf{Zonghai Yao}$^{2}$, \\
  \textbf{Talha Chafekar}$^{1}$, 
  \textbf{Deepali Singh}$^{1}$,
  \textbf{Hong Yu}$^{1,2,3,4}$ \\
  $^{1}$UMass Amherst,
  $^{2}$UMass Lowell,
  $^{3}$UMass Chan Medical School,
  $^{4}$VA Bedford Health Care \\
  \texttt{sunjaekwon@umass.edu}
}

\usepackage{tcolorbox}
\begin{document}
\maketitle
\begin{abstract}
We introduce \textbf{S}ense-\textbf{A}ligned \textbf{G}rounding \textbf{E}valuation (\textbf{SAGE}), a comprehensive benchmark for multimodal lexical grounding across noun-, verb-, and entity-centered settings. SAGE reformulates VWSD, VerSe, and WikiDiverse into a shared image-selection protocol with semantically confusable candidates and a sense-preserving image swap for probing verb grounding beyond the original caption--image pairing. Across models, rankings are only partially consistent across subsets, and the VerSe diagnostic reveals substantial differences in the contribution of target-verb meaning. We further introduce \textbf{G}loss-\textbf{A}ugmented \textbf{R}easoning (\textbf{GAR}), a training-free framework for probing which explicit semantic signals transfer across grounding settings. Definition augmentation improves all three subsets, while constraint-guided reasoning benefits noun and verb grounding but is less effective for fine-grained entity discrimination. These results show that lexical grounding behavior and useful semantic evidence vary across grounding settings. The code and data will be made publicly available at \url{https://github.com/soon91jae/SAGE}.
\end{abstract}

\section{Introduction}

Lexical grounding requires more than broad image--text compatibility: a model must identify which meaning of a target expression is intended in context and visually instantiated. For example, \textit{mortar} may denote a building material, a weapon, or a vessel; \textit{serve} may describe providing food or putting a ball into play; and \textit{Shadow} may refer to a drone, a film, or a fictional character. Such distinctions matter in image retrieval, where ambiguous queries may retrieve competing meanings~\citep{loeff-etal-2006-discriminating}. Yet correct image selection may reflect other compatible visual cues rather than discrimination of the intended meaning.

\begin{figure}
    \centering
    \includegraphics[width=.85\linewidth]{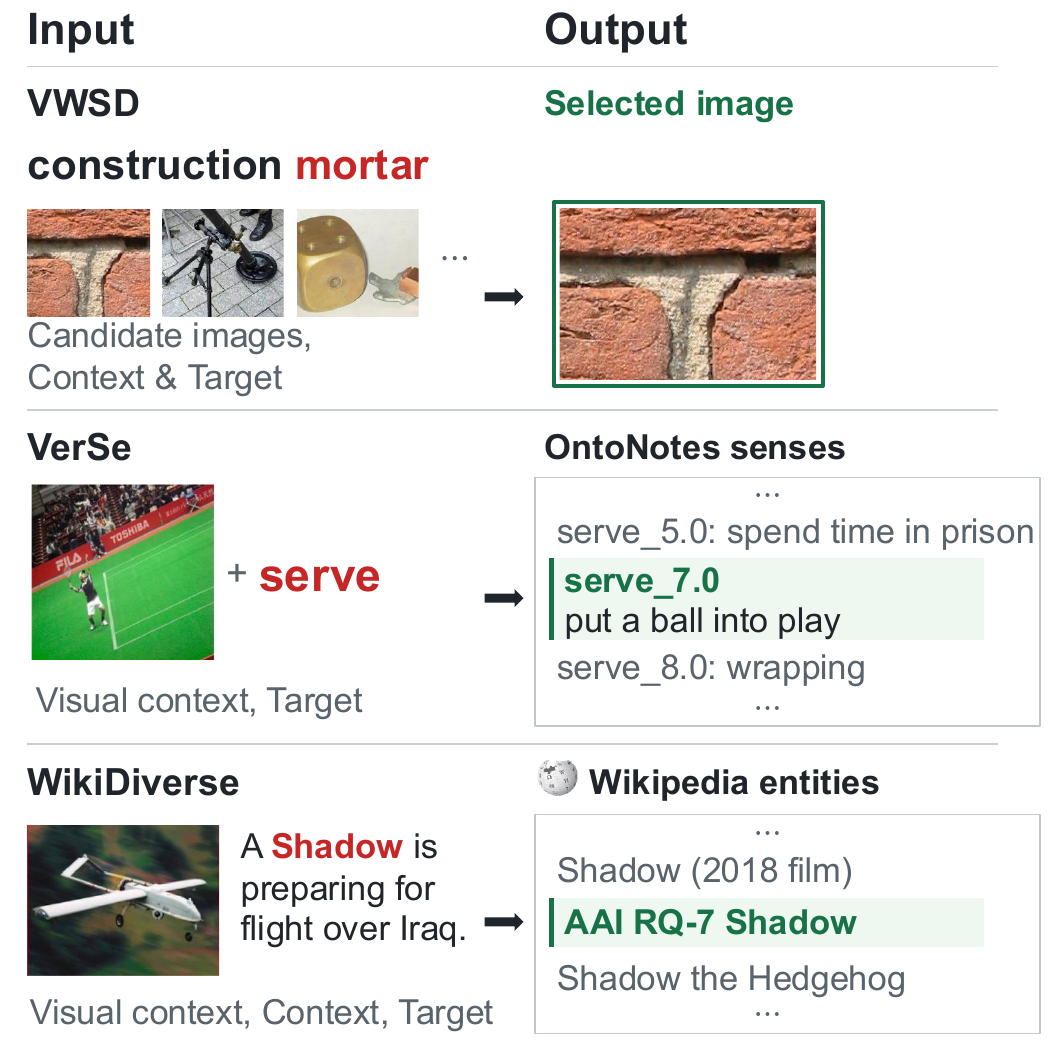}
    \caption{
Original task formulations of the source datasets in SAGE. VWSD selects a candidate image, while VerSe and WikiDiverse map targets to OntoNotes senses and Wikipedia entities, respectively. \textcolor{red}{Red} indicates targets; \textcolor{green!60!black}{green} indicates correct outputs. Alternative entries are illustrative; ellipses indicate omissions.}    
\label{fig:benchmark-comparison}
\end{figure}

Existing multimodal lexical resources differ in focus and task formulation. As shown in Figure~\ref{fig:benchmark-comparison}, Visual Word Sense Disambiguation (VWSD) is mainly noun-centered (Appendix~\ref{apx:pos_distribution}), VerSe targets verb senses, and WikiDiverse focuses on named entities, with outputs ranging from candidate images to OntoNotes senses and Wikipedia entities~\citep{gella2016unsupervised,wang2022wikidiverse,raganato2023semeval}. These differences hinder systematic comparisons across lexical settings, motivating a shared evaluation protocol. We adopt VWSD-style image selection to standardize the decision format without requiring a common sense inventory or knowledge base, while preserving the distinct semantic demands of each setting.

Verb grounding presents an additional evaluation challenge. Prior work identifies difficulties in visual verb understanding~\citep{hendricks2021probing,yao2026towards}. In naturally paired image--caption data such as MS-COCO \citep{chen2015microsoft}, objects, participants, and scene context may support image selection without discrimination of the target verb's meaning. Thus, high accuracy may obscure the verb's contribution. VerSe provides OntoNotes verb-sense annotations over MS-COCO image--caption pairs, enabling evaluation beyond their original pairing.

To address these challenges, we introduce \textbf{S}ense-\textbf{A}ligned \textbf{G}rounding \textbf{E}valuation (\textbf{SAGE}), which reformulates VWSD, VerSe, and WikiDiverse under a shared 10-candidate image-selection protocol with semantically confusable alternatives. Pairwise analysis confirms that hard negatives are substantially more confusable than random alternatives. For VerSe, we introduce \textbf{S}ense-\textbf{P}reserving \textbf{I}mage \textbf{S}wap (\textbf{SPIS}), replacing the original caption-paired image with another image annotated with the same OntoNotes verb sense of VerSe. Combined with target-verb neutralization, SPIS examines verb sensitivity when original image--caption pairing cues are weakened. Model rankings are only partially consistent across SAGE subsets (Kendall's $\tau_b=0.46$--$0.74$). Moreover, three models with similar original VerSe scores (91--92 Hits@1) diverge under SPIS (43.7--74.3), with target-verb neutralization causing performance drops of 3.4--33.0 points. These findings reveal setting-dependent grounding behavior and differences in verb sensitivity obscured by original pairing.

Beyond evaluating grounding performance, SAGE also provides a common setting for investigating which lexical-semantic information supports discrimination among plausible candidates. We therefore investigate both the sources of lexical-semantic information and how that information is used for image discrimination. To this end, we introduce \textbf{G}loss-\textbf{A}ugmented \textbf{R}easoning (\textbf{GAR}), a training-free framework comparing context-generated and lexical-resource-grounded definitions and evaluating two uses of lexical-semantic representations: holistic image ranking and \textbf{C}onstraint-\textbf{G}uided \textbf{R}easoning (\textbf{CGR}). CGR decomposes intended meanings into object, attribute, and state constraints for visual verification.

Our experiments show that lexical-resource-grounded definitions improve all three SAGE subsets, whereas context-generated definitions yield limited and inconsistent gains. CGR improves VWSD and particularly VerSe but degrades performance on WikiDiverse. Combining lexical-definition ranking and CGR achieves the highest macro-average performance, although the best strategy varies across subsets. Object constraints provide the strongest individual signal, while 76\% of examined WikiDiverse CGR errors involve non-discriminative constraints. These findings suggest that lexical-semantic information can improve grounding, but constraint-based verification depends on whether the constraints distinguish competing candidates.

Our main contributions are:
\begin{itemize}
    \item We introduce \textbf{SAGE}, a 10-candidate image-selection benchmark enabling systematic comparison of noun-, verb-, and entity-centered grounding with semantically confusable alternatives.

    \item We introduce \textbf{SPIS}, a verb-specific diagnostic combining sense-preserving image swaps and target-verb neutralization to reveal differences in verb sensitivity obscured by original pairing.

    \item We introduce \textbf{GAR}, a training-free framework examining the sources and uses of lexical-semantic information, revealing consistent benefits from lexical-resource-grounded definitions but setting-dependent effects of constraint-guided visual verification.
\end{itemize}

\section{Related Work}
 \paragraph{Multimodal Lexical Grounding and Evaluation} Multimodal ambiguity has been studied through visual puns, VQA-style tasks, and multilingual or utterance-level reasoning ~\citep{chung2024can,wang2025mma,wang2025mucar}. More directly, VWSD selects an image for a contextualized ambiguous term~\citep{raganato2023semeval}, VerSe evaluates visual grounding of verb senses using OntoNotes labels~\citep{gella2016unsupervised}, and WikiDiverse evaluates multimodal entity linking to Wikipedia entities~\citep{wang2022wikidiverse}. These resources cover complementary noun-, verb-, and entity-centered grounding settings but differ in task formulations and output spaces. SAGE adopts a shared candidate-image selection protocol to enable systematic comparison across these settings. This approach follows prior work on standardizing heterogeneous WSD resources for cross-resource evaluation~\citep{raganato2017word}. 

 \paragraph{Semantic Enrichment and Visual Verification} Enriching textual targets with semantic information is common in vision--language prediction. DCLIP, CuPL, and WaffleCLIP augment class names with generated or auxiliary text for image classification ~\citep{menon2023visual,pratt2023platypus,roth2023waffling}. Related VWSD methods use glosses, lexical knowledge, and generated definitions for context-aware image ranking ~\citep{kwon2023vision,kritharoula2023large,yang2024polclip, papastavrou2024arpa,qiao2025quantum}. While these approaches demonstrate the utility of semantic enrichment, they do not systematically examine its sources and uses across different lexical grounding settings. GAR addresses this gap by comparing context-generated and lexical-resource-grounded definitions and contrasting holistic definition-based ranking with constraint-guided visual verification. This enables analysis of both the contribution of external lexical knowledge and the effectiveness of decomposed visual constraints across grounding settings. 

\section{SAGE Benchmark}
\subsection{Task Setup}

\paragraph{Problem Definition}
Given an input $(C, t, I)$, where $C$ is a textual context, $t$ is a target term or mention, and $I = \{i_1, \ldots, i_{10}\}$ is a candidate image set with exactly one gold image, the task is to select the image that best instantiates the intended sense or entity of $t$ in $C$. The target term is the unit of supervision: an image is correct when it matches the contextualized lexical meaning of $t$, even if not all non-target details in $C$ are evaluated. This distinction is especially important for the SPIS VerSe diagnostic, where the gold image preserves the target verb sense while allowing surrounding objects, participants, and scenes to vary. Formally, models produce a score $r(i; C,t)$ for each candidate image and select the highest scored candidate. This follows the VWSD-style candidate-ranking setup making the lexical grounding target explicit.

\paragraph{Evaluation Criteria}
The main evaluation criteria of the benchmark are Hits@1 and Mean Reciprocal
Rank (MRR). Herein, Hits@1 measures the fraction of examples for which the gold image
is ranked first, while MRR averages its reciprocal rank across examples.
Detailed definitions are provided in Appendix~\ref{apx:metircs}.

\subsection{Benchmark Datasets}
\label{sec:benchmark}
This section describes the three source datasets used to construct SAGE and how each is reformulated into the common 10-way image-selection protocol. We also detail candidate-set construction, including semantically confusable hard negatives and random negatives.

\paragraph{VWSD} The VWSD subset already follows SAGE's image-selection format ~\citep{raganato2023semeval}, with each instance consisting of a textual context $C$, an ambiguous target term $t$, and a set of candidate images. Thus, we use VWSD without modification, preserving its original candidate sets and data splits. Herein, VWSD serves as the noun-centered anchor and maintains direct comparability with the standard VWSD evaluation protocol. Since the English test split is predominantly noun-centered, VerSe and WikiDiverse complement it with verb- and entity-centered grounding settings.

\begin{figure}[t]     \centering     \includegraphics[width=.8\columnwidth]{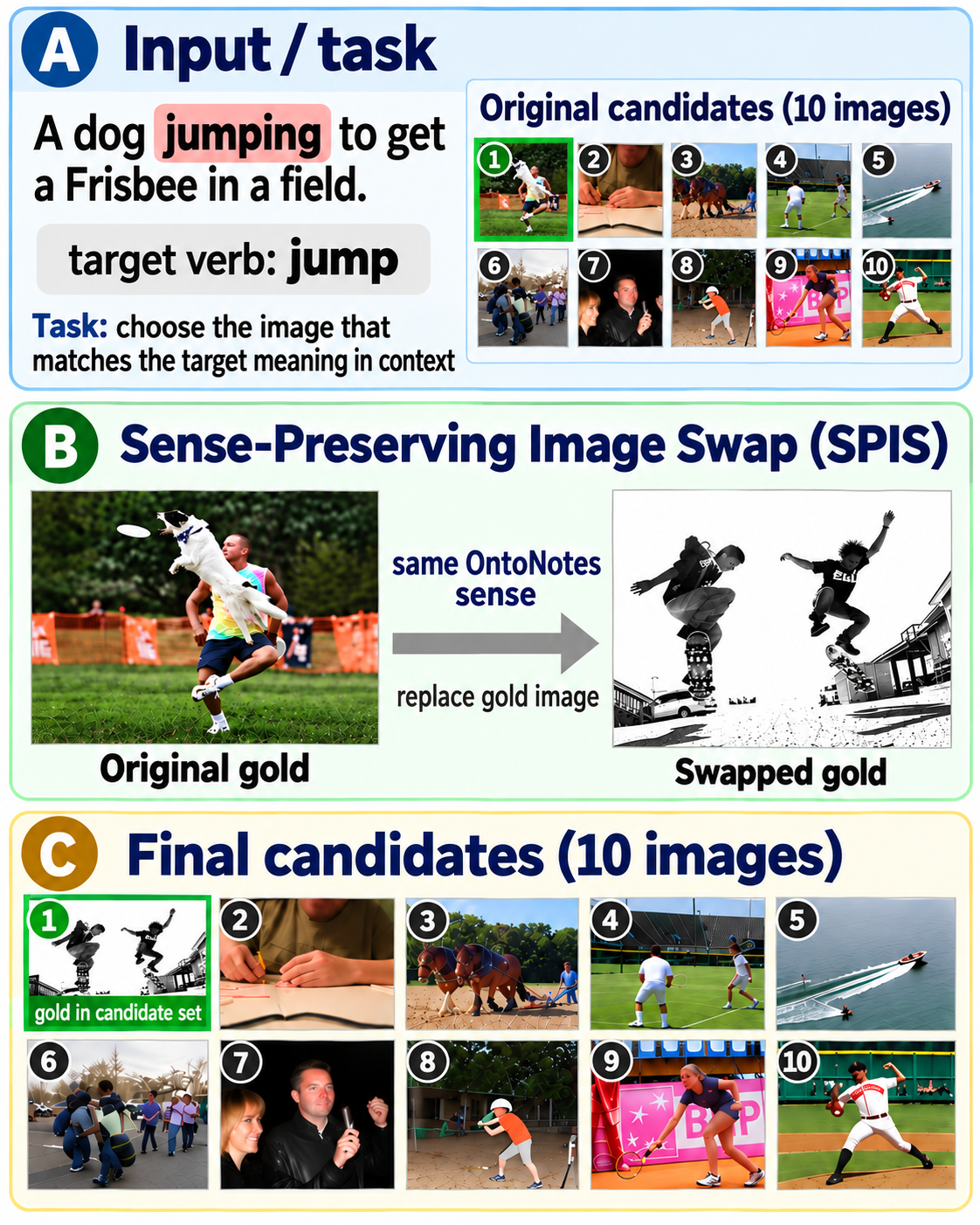}     \caption{Illustration of SPIS. Given a context and target verb, the task is to select the candidate image depicting the intended verb sense. SPIS replaces the original caption-paired gold image with another image annotated with the same OntoNotes verb sense, preserving the target meaning while varying objects, participants, and scene context. The swapped image becomes the gold image in the final 10-candidate set.}     \label{fig:context-swap} \end{figure}

\paragraph{VerSe}
VerSe is a multimodal verb sense disambiguation dataset derived from MS-COCO and TUHOI, containing roughly 3.5K images annotated with 90 verbs and their OntoNotes senses. In SAGE, we use the MS\mbox{-}COCO~\citep{chen2015microsoft} portion, which provides human-annotated image captions. We preserve the original MS-COCO train/test split to remain consistent with prior work and avoid train--test leakage. Additional details and examples are provided in Appendix~\ref{apx:verse_details}.

We reconstruct VerSe as an image-selection task using its verb-sense annotations and the associated MS-COCO captions. Each instance is represented as $(C,t,I)$, where $C$ is the caption, $t$ is the target verb, and $I$ contains 10 candidate images. Hard negatives are sampled from images depicting alternative senses of the same polysemous verb, as identified by OntoNotes annotations, requiring discrimination between competing verb meanings. If fewer than nine hard negatives are available, the remaining candidates are randomly sampled from MS-COCO images that do not match the target.

We further introduce SPIS as a verb-specific diagnostic as shown in Figure~\ref{fig:context-swap}. In the original pairing, the caption and gold image share objects, participants, and scene details beyond the target verb sense. Although useful for ordinary image--text retrieval, these cues may allow correct image selection without discriminating the intended verb meaning. SPIS replaces the original caption-paired gold image with another image annotated with the same OntoNotes verb sense, producing a final candidate set with the swapped image as its gold. This preserves the intended lexical meaning while varying its visual realization and surrounding context. Combined with target-verb neutralization, SPIS examines sensitivity to verb information when original caption--image compatibility is weakened.

\paragraph{WikiDiverse} We reformulate WikiDiverse, a multimodal entity linking (MEL) corpus, into an entity-centric image-selection task while preserving its original target mention, context, and candidate-entity structure. Each instance is represented as \((C,t,e^*,I)\), where \(C\) is the original textual context, \(t\) is the target mention, \(e^*\) is the gold entity, and \(I\) contains images associated with candidate entities. For every candidate entity, we retrieve its Wikipedia page and select a representative image using the same rule: the page thumbnail when available, otherwise the topmost image.  

Distractors are drawn from the official top-10 candidate entities released with WikiDiverse\footnote{\url{https://github.com/wangxw5/wikiDiverse}} for the same mention. We remove aliases or redirects of \(e^*\), identified using Wikidata QIDs, as well as candidates whose Wikipedia pages contain no image. The remaining entities serve as hard negatives because they are plausible MEL competitors for the same mention. If fewer than nine remain, we fill the set with type-matched random entities whose Wikipedia pages contain an image. We enforce that exactly one candidate corresponds to \(e^*\) and preserve the original WikiDiverse split for comparability with the source setting. Detailed examples and explanations are provided in Appendix~\ref{apx:wikidiverse_details}.

\begin{table}[t]
\centering
\scriptsize
\begin{tabular}{lccc}
\hline
& VWSD & VerSe & WikiDiverse \\
\hline\hline
Train & 12,869 & 1,042 & 10,750 \\
Validation & - & - & 1,288 \\
Test & 463 & 1,264 & 1,243 \\
\hline
\end{tabular}
\caption{Summary statistics of the three datasets in SAGE after reformulation.}
\label{tab:data_stat}
\end{table}

\paragraph{Data Statistics}
Table~\ref{tab:data_stat} summarizes the dataset sizes in SAGE after reformulation. VWSD provides the noun-centered anchor set, VerSe provides the verb-focused subset, and WikiDiverse contributes the largest entity-centric component with an official validation split. Across all datasets, each instance is standardized to 10 image candidates with exactly one gold image, enabling direct comparison across datasets.

\paragraph{Human Validation on Reconstructed Benchmark}
We conduct human validation for the newly reformulated VerSe and WikiDiverse
subsets to verify their reconstructed gold-image assignments. We randomly sample 200 instances from the test split of each dataset and obtain independent labels from three annotators. Inter-annotator reliability is strong when computed over all 200 instances (Fleiss'~$\kappa=0.801$ for VerSe and $\kappa=0.823$ for WikiDiverse). Excluding the cases in which all three annotators selected different images—7 instances for VerSe and 5 for WikiDiverse—the majority-vote labels agree with the SAGE gold labels in 91.2\% of VerSe instances and 94.9\% of WikiDiverse instances. These results indicate that the reformulated data are reliable for evaluation and analysis. The remaining disagreements typically involve visually plausible alternatives, indirect but related gold images, or fine-grained sense or referent distinctions requiring additional world knowledge, reflecting the inherent difficulty of multimodal sense and entity grounding rather than systematic annotation error. Detailed analyses are provided in Appendix~\ref{apx:human_analysis}.

\begin{figure*}[!ht]
    \centering
    \includegraphics[width=.85\linewidth]{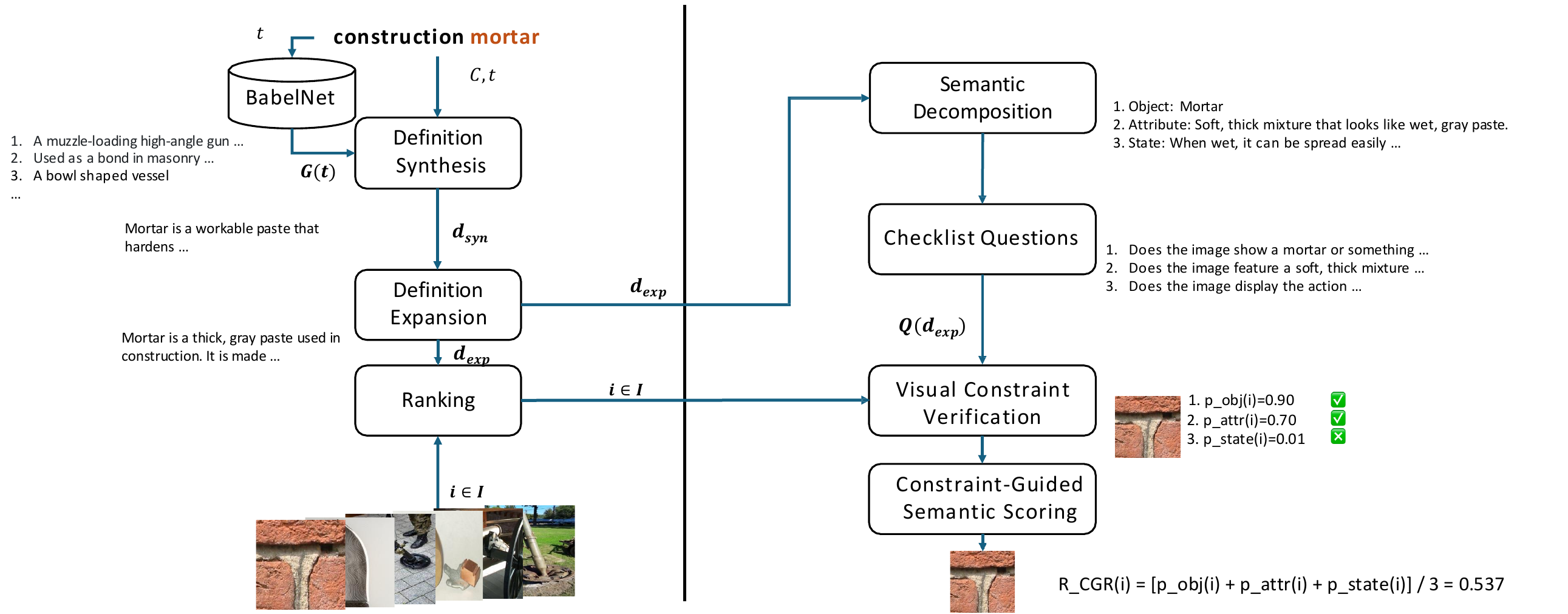}
    \caption{Illustrative overview of the proposed GAR framework. Given a target term $t$ and context $C$, GAR first synthesizes a contextually plausible definition from candidate glosses retrieved from a lexical knowledge base and expands it into visually checkable evidence. The expanded definition is decomposed into semantic constraints, reformulated as binary visual questions, and verified against each candidate image. CGR aggregates the affirmative probabilities into an interpretable semantic score for reranking.}
    \label{fig:illustrative_concept}
\end{figure*}

\section{Gloss-Augmented Reasoning}
\label{sec:gar}

This section introduces GAR, which is designed to examine which forms of explicit lexical-semantic information support grounding across different settings. To separate the contribution of richer semantic descriptions from explicit visual verification, GAR represents the contextually intended meaning both as a holistic definition and as decomposed visually verifiable constraints. GAR consists of three components:
(i) Context-Guided Definition Synthesis,
(ii) Definition Expansion, and
(iii) Constraint-Guided Reasoning (CGR).

\paragraph{Context-Guided Definition Synthesis}
From a lexical knowledge base (LKB)\footnote{Following
\citet{yang2024polclip}, we use BabelNet~\citep{navigli2012babelnet}.},
we retrieve candidate senses or entities and their canonical glosses
for the target term or mention. Given the target, its textual context,
and the candidate glosses, an LLM synthesizes a single concise,
dictionary-style definition of the contextually intended meaning,
denoted by $d_{\mathrm{syn}}$. This stage does not select a
lexical-resource sense index; the glosses serve only as reference
information. Candidate images and benchmark gold labels are not provided. For example, for \textit{mortar} in ``mortar construction,'' the synthesized definition is ``Mortar is a workable paste that hardens to bind building blocks together, filling gaps and distributing weight.'' A human plausibility assessment of the synthesized definitions, including successful recovery from missing retrieved senses and failure cases, is provided in Appendix~\ref{apx:definition_plausibility}.

\paragraph{Definition Expansion}
Dictionary definitions are often too short for multimodal verification.
The expander therefore rewrites $d_{\mathrm{syn}}$ into a compact
plain-language expanded definition. The expansion preserves the same
meaning while making visually checkable cues, such as parts, shape,
material, participants, action, or state, more explicit. The resulting
definition $d_{\mathrm{exp}}$ serves as a semantic specification for
image ranking.

\paragraph{Constraint-Guided Semantic Scoring}
CGR decomposes $d_{\mathrm{exp}}$ into visually checkable binary
constraints over object, attribute, and state; axes without applicable
evidence are omitted. Object constraints identify relevant referents,
attribute constraints capture visible properties or relations, and state
constraints describe actions or conditions. For each candidate image
$i \in I$ and constraint $c \in Q(d_{\mathrm{exp}})$, the VLM assigns a
normalized ``Yes'' probability $p(c \mid i)$:

\vspace{-.1cm}
{\small
\[
R_{\mathrm{CGR}}(i; C,t)=
\frac{1}{|Q(d_{\mathrm{exp}})|}
\sum_{c\in Q(d_{\mathrm{exp}})} p(c\mid i).
\]
}
\vspace{-.1cm}
Then, the highest-scoring candidate is selected.

\section{Experiments}
We evaluate three model groups on SAGE: image--text matching (ITM) baselines, open-source VLMs, and closed-source VLMs. The ITM baselines include CLIP ViT-L/14, SigLIP2 (0.3B, 1B)~\citep{tschannen2025siglip}, and UVWSD~\citep{kwon2023vision}, a CLIP-based method augmented with lexical knowledge. The open-source VLMs include Qwen3-VL (2B, 8B), InternVL 3.5 (2B, 8B), and Gemma3 (4B, 12B), evaluated using their official Hugging Face releases~\citep{bai2025qwen3,team2024gemma,wang2025internvl3}. The closed-source VLMs are GPT-4.1-mini, GPT-5.6 -Luna, -Terra, -Sol and GPT-6 -Luna, -Sol evaluated through their official APIs~\citep{openai2024api}. All generative models use deterministic decoding with temperature 0. Each model ranks the same 10-image candidate sets. For Hits@1, statistical significance is assessed using McNemar's test; for MRR, we use a paired $t$-test. 


\subsection{Experimental Results}

\begin{table*}[t]
\centering
\scriptsize
\begin{tabular}{c|l|cc|cc|cc|cc}
\hline
\multicolumn{1}{c|}{\multirow{2}{*}{Model Type}}
& \multicolumn{1}{c|}{\multirow{2}{*}{Baselines}}
& \multicolumn{2}{c|}{VWSD}
& \multicolumn{2}{c|}{VerSe}
& \multicolumn{2}{c|}{WikiDiverse}
& \multicolumn{2}{c}{Macro Avg.} \\
\cline{3-10}
&
& \multicolumn{1}{c}{Hits@1}
& \multicolumn{1}{c|}{MRR}
& \multicolumn{1}{c}{Hits@1}
& \multicolumn{1}{c|}{MRR}
& \multicolumn{1}{c}{Hits@1}
& \multicolumn{1}{c|}{MRR}
& \multicolumn{1}{c}{Hits@1}
& \multicolumn{1}{c}{MRR} \\
\hline\hline

& CLIP
& \CI{57.7}{4.5} & \CI{72.2}{3.1}
& \CI{43.7}{2.7} & \CI{60.6}{2.0}
& \CI{48.9}{2.8} & \CI{68.7}{1.8}
& \CI{50.1}{2.0} & \CI{67.2}{1.4} \\

& UVWSD
& \CI{60.9}{4.4} & \CI{75.2}{2.9}
& \CI{45.7}{2.8} & \CI{62.2}{2.0}
& \CI{52.3}{2.8} & \CI{70.5}{1.8}
& \CI{53.0}{2.0} & \CI{69.3}{1.3} \\

& SigLIP2-0.3B
& \textbf{\CI{76.0}{3.9}} & \textbf{\CI{84.9}{2.5}}
& \textbf{\CI{51.4}{2.8}} & \textit{\CI{66.2}{2.0}}
& \textit{\CI{51.6}{2.8}} & \textit{\CI{70.5}{1.8}}
& \textit{\CI{59.7}{1.9}} & \textit{\CI{73.9}{1.2}} \\

\multirow{-4}{*}{ITMs}
& SigLIP2-1B
& \textit{\CI{74.7}{4.0}} & \textit{\CI{84.3}{2.5}}
& \textit{\CI{51.2}{2.8}} & \textbf{\CI{66.6}{2.0}}
& \textbf{\CI{53.5}{2.8}} & \textbf{\CI{71.9}{1.8}}
& \textbf{\CI{59.8}{1.9}} & \textbf{\CI{74.3}{1.2}} \\

\hline

& InternVL 3.5-2B
& \CI{55.1}{4.5} & \CI{69.7}{3.2}
&\textbf{ \CI{67.2}{2.6}} & \textbf{\CI{78.9}{1.7}}
& \CI{44.8}{2.8} & \CI{63.8}{1.9}
& \CI{55.7}{2.0} & \CI{70.8}{1.4} \\

& InternVL 3.5-8B
& \CI{66.1}{4.3} & \CI{78.6}{2.8}
& \textit{\CI{59.5}{2.7}} & \textit{\CI{73.1}{1.9}}
& \CI{51.1}{2.8} & \CI{67.8}{1.9}
& \CI{58.9}{2.0} & \CI{73.2}{1.3} \\

& Gemma3-4B
& \CI{57.9}{4.5} & \CI{72.6}{3.0}
& \CI{51.2}{2.7} & \CI{66.0}{2.0}
& \CI{58.5}{2.8} & \CI{73.3}{1.8}
& \CI{55.9}{2.0} & \CI{70.6}{1.4} \\

& Gemma3-12B
& \textit{\CI{75.6}{3.9}} & \textit{\CI{84.7}{2.5}}
& \CI{59.3}{2.7} & \CI{72.7}{1.9}
& \textit{\CI{62.8}{2.7}} & \textit{\CI{76.9}{1.7}}
& \textbf{\CI{65.9}{1.8}} & \textbf{\CI{78.1}{1.2}} \\

& Qwen3-VL-2B
& \CI{62.2}{4.4} & \CI{74.6}{3.1}
& \CI{54.1}{2.7} & \CI{68.2}{2.0}
& \CI{54.9}{2.8} & \CI{71.3}{1.8}
& \CI{57.1}{2.0} & \CI{71.4}{1.4} \\

\multirow{-6}{*}{\shortstack{Open-source\\VLMs}}
& Qwen3-VL-8B
& \textbf{\CI{76.5}{3.9}} & \textbf{\CI{85.4}{2.5}}
& \CI{52.4}{2.8} & \CI{66.1}{2.0}
& \textbf{\CI{64.5}{2.7}} & \textbf{\CI{77.5}{1.8}}
& \textit{\CI{64.5}{1.8}} & \textit{\CI{76.3}{1.2}} \\

\hline

& GPT-4.1-mini
& \CI{78.6}{3.8} & \CI{86.9}{2.4}
& \CI{58.2}{2.7} & \CI{71.1}{2.0}
& \CI{66.6}{2.6} & \CI{79.8}{1.6}
& \CI{67.8}{1.8} & \CI{79.3}{1.2} \\

& GPT-5.6-Luna
& \CI{77.3}{3.9} & \CI{86.6}{2.3}
& \CI{67.6}{2.6} & \CI{78.3}{1.8}
& \CI{66.1}{2.6} & \CI{79.5}{1.7}
& \CI{70.3}{1.8} & \CI{81.5}{1.1} \\

& GPT-5.6-Terra
& \CI{77.1}{3.8} & \CI{86.7}{2.3}
& \CI{73.5}{2.4} & \CI{82.2}{1.7}
& \CI{66.6}{2.6} & \CI{79.9}{1.6}
& \CI{72.4}{1.7} & \CI{82.9}{1.1} \\

& GPT-5.6-Sol
& \textbf{\CI{83.8}{3.4}} & \textbf{\CI{90.8}{1.9}}
& \textbf{\CI{74.3}{2.4}} & \textbf{\CI{83.4}{1.6}}
& \textbf{\CI{73.6}{2.5}} & \textbf{\CI{84.4}{1.5}}
& \textbf{\CI{77.2}{1.6}} & \textbf{\CI{86.2}{1.0}} \\

& GPT-6-Luna
& \CI{73.2}{4.1} & \CI{83.7}{2.5}
& \CI{65.1}{2.6} & \CI{76.8}{1.8}
& \CI{59.1}{2.8} & \CI{74.3}{1.8}
& \CI{65.8}{1.9} & \CI{78.3}{1.2} \\

\multirow{-6}{*}{\shortstack{Closed-source\\VLMs}}
& GPT-6-Sol
& \textit{\CI{82.9}{3.5}} & \textit{\CI{90.1}{2.0}}
& \textit{\CI{70.4}{2.6}} & \textit{\CI{81.2}{1.7}}
& \textit{\CI{70.8}{2.5}} & \textit{\CI{82.5}{1.6}}
& \textit{\CI{74.7}{1.7}} & \textit{\CI{84.6}{1.0}} \\

\hline
\end{tabular}

\caption{
Experimental results on SAGE. Scores are point estimates $\pm$ 95\% bootstrap
confidence-interval half-widths (10,000 item-level resamples). Within each
model type, best and second-best scores are \textbf{bolded} and
\textit{italicized}, respectively. Ranking procedures are in
Appendix~\ref{apx:baseline_ranking}.
}
\label{tab:baseline_performance}
\end{table*}

\begin{table}[t]
\centering
\scriptsize
\setlength{\tabcolsep}{6pt}
\begin{tabular}{l|ccc}
\hline
 & VWSD & VerSe & WikiDiverse \\
\hline\hline
VWSD        & 1.00 & 0.46 & 0.74 \\
VerSe       & 0.46 & 1.00 & 0.47 \\
WikiDiverse & 0.74 & 0.47 & 1.00 \\
\hline
\end{tabular}
\caption{Kendall's $\tau_b$ correlations between model rankings across SAGE subsets in Table~\ref{tab:baseline_performance}. All off-diagonal correlations are significant at $p<0.05$.}
\label{tab:rank_corr}
\end{table}

\paragraph{Experiments on Baseline Models}
Table~\ref{tab:baseline_performance} reveals substantial variation in how model strengths transfer across grounding settings. Model rankings are positively correlated across all three subsets, but their agreement varies considerably: Kendall's $\tau_b$ is 0.74 for VWSD--WikiDiverse, compared with 0.46 for VWSD--VerSe and 0.47 for VerSe--WikiDiverse (Table~\ref{tab:rank_corr}). These results suggest shared grounding capabilities alongside setting-specific differences, with VerSe showing the weakest agreement with the other two subsets.

This variation is also evident across model families. SigLIP2 remains competitive on VWSD, but its relative strength does not transfer equally to VerSe and WikiDiverse. Conversely, InternVL~3.5-2B performs particularly well on VerSe despite weaker results on the other subsets. Qwen3-VL-8B shows the opposite pattern, performing strongly on VWSD and WikiDiverse but less competitively on VerSe. Together, these contrasting profiles indicate that performance on one grounding setting does not necessarily predict performance on another, motivating evaluation across complementary noun-, verb-, and entity-centered settings.

Model scale generally helps but does not fully determine grounding performance. Across within-family scaling comparisons, larger variants either improve or show no statistically significant change in Hits@1, with one exception: InternVL~3.5-8B significantly underperforms its 2B counterpart on VerSe. Newer model generations also do not uniformly outperform earlier ones: GPT-5.6-Sol scores higher than GPT-6-Sol across all three subsets. Substantial headroom remains, particularly on VerSe and WikiDiverse, where the strongest model achieves only 74.3 and 73.6 Hits@1, respectively.

\begin{table}[t]
\centering
\small
\resizebox{\columnwidth}{!}{%
\begin{tabular}{l|rrrr|rr}
\hline
Model & Orig. & Orig. (Neu.) & SPIS & SPIS (Neu.) &
$\Delta_{\mathrm{SPIS}}\downarrow$ &
$\Delta_{\mathrm{verb}}\uparrow$ \\
\hline
CLIP & 91.7 & 90.7 & 43.7 & 40.3 & 48.0 & 3.4 \\
Gemma3-12B & 91.3 & 90.1 & 59.3 & 41.1 & 32.0 & 18.2 \\
GPT-5.6-Sol & 92.0 & 91.2 & 74.3 & 41.3 & 17.7 & 33.0 \\
\hline
\end{tabular}%
}
\caption{VerSe Hits@1 under original pairing and SPIS,
with and without target-verb neutralization.
$\Delta_{\mathrm{SPIS}}=\mathrm{Original}-\mathrm{SPIS}$
measures the performance change under sense-preserving
image replacement, and
$\Delta_{\mathrm{verb}}=\mathrm{SPIS}-\mathrm{SPIS(Neu.)}$
measures sensitivity to target-verb information.}
\label{tab:verse_swap}
\end{table}

\paragraph{SPIS Diagnostic Evaluation}
We compare four VerSe settings to examine target-verb sensitivity relative to pair-specific caption--image compatibility. \textit{Original} uses the original caption-paired gold image, whereas \textit{SPIS}, used in the final VerSe benchmark, replaces it with another image annotated with the same OntoNotes verb sense. \textit{Original-Neutralized} and \textit{SPIS-Neutralized} additionally replace the target verb phrase with a generic expression (e.g., ``doing something''). SPIS thus preserves the intended verb sense while weakening pair-specific visual compatibility, and neutralization probes sensitivity to target-verb information. Details are provided in Appendix~\ref{apx:verb_neutralization}.

As Table~\ref{tab:verse_swap} shows, the original paired setting largely obscures differences in target-verb sensitivity. All three models achieve 91--92 Hits@1, and verb neutralization causes only marginal changes, leaving performance above 90\%. Under SPIS, however, their scores diverge substantially: CLIP, Gemma3-12B, and GPT-5.6-Sol achieve 43.7, 59.3, and 74.3 Hits@1, respectively, with significant pairwise differences ($p<10^{-5}$).

Under SPIS-Neutralized, all three models converge to similar performance (40.3--41.3 Hits@1). The corresponding drops are 3.4, 18.2, and 33.0 points for CLIP, Gemma3-12B, and GPT-5.6-Sol, respectively. These findings indicate that stronger VLMs exhibit greater sensitivity to target-verb information under SPIS, a difference largely obscured by the original image--caption pairing.


\begin{table*}[t]
\centering
\scriptsize
\begin{tabular}{c|l|cc|cc|cc|cc}
\hline
\multicolumn{1}{c|}{\multirow{2}{*}{Model Type}}
& \multicolumn{1}{c|}{\multirow{2}{*}{Method}}
& \multicolumn{2}{c|}{VWSD}
& \multicolumn{2}{c|}{VerSe}
& \multicolumn{2}{c|}{WikiDiverse}
& \multicolumn{2}{c}{Macro Avg.} \\
\cline{3-10}
&
& \multicolumn{1}{c}{Hits@1}
& \multicolumn{1}{c|}{MRR}
& \multicolumn{1}{c}{Hits@1}
& \multicolumn{1}{c|}{MRR}
& \multicolumn{1}{c}{Hits@1}
& \multicolumn{1}{c|}{MRR}
& \multicolumn{1}{c}{Hits@1}
& \multicolumn{1}{c}{MRR} \\
\hline\hline

\multirow{5}{*}{Gemma3-12B}
& Base
& 75.6 & 84.7
& 59.3 & 72.7
& 62.8 & 76.9
& 65.9 & 78.1 \\

& Generated Def.
& 75.8 & 84.9
& 63.2$^{\dagger}$ & 75.6$^{\dagger}$
& 59.5$^{\dagger}$ & 74.7$^{\dagger}$
& 66.2 & 78.4 \\

& Lexical Def.
& 79.9$^{\dagger}$ & 87.7$^{\dagger}$
& 67.2$^{\dagger}$ & 78.9$^{\dagger}$
& \textbf{64.6}$^{*}$ & \textbf{78.0}$^{*}$
& 70.6 & 81.5 \\

& CGR
& 78.6$^{*}$ & 86.2$^{*}$
& \textbf{71.7}$^{\dagger}$ & \textbf{81.7}$^{\dagger}$
& 56.7$^{\dagger}$ & 71.3$^{\dagger}$
& 69.0 & 79.7 \\

& Full GAR
& \textbf{81.2}$^{\dagger}$ & \textbf{88.6}$^{\dagger}$
& 69.8$^{\dagger}$ & 81.1$^{\dagger}$
& 62.7 & 76.8
& \textbf{71.2} & \textbf{82.1} \\

\hline
\end{tabular}

\caption{
Ablation study of GAR.
Generated Def. produces a context-specific definition without lexical-resource
retrieval; Lexical Def. retrieves lexical knowledge and synthesizes an expanded
definition for holistic ranking; CGR decomposes the lexical-semantic
representation into visually verifiable constraints; and Full GAR combines
Lexical Def. and CGR using RRF.
$^{*}p<0.05$ and $^{\dagger}p<0.01$ show significant differences from Base.
}
\label{tab:ablation_results}
\end{table*}

\begin{table}[t]
\centering
\scriptsize
\begin{tabular}{l|cc|cc|cc}
\hline
\multicolumn{1}{c|}{\multirow{2}{*}{\shortstack{Axis}}}
& \multicolumn{2}{c|}{VWSD}
& \multicolumn{2}{c|}{VerSe}
& \multicolumn{2}{c}{WikiDiverse} \\
\cline{2-7}
& \multicolumn{1}{c}{Hits@1} & \multicolumn{1}{c|}{MRR}
& \multicolumn{1}{c}{Hits@1} & \multicolumn{1}{c|}{MRR}
& \multicolumn{1}{c}{Hits@1} & \multicolumn{1}{c}{MRR} \\
\hline\hline

Object
& 75.4 & 84.2 & 69.4 & 79.8 & \textbf{57.0} & \textbf{71.4} \\
Attribute
& 74.3 & 83.5 & 63.9 & 76.1 & 42.0 & 68.4 \\
State
& 67.6 & 77.7 & 67.8 & 79.2 & 32.1 & 50.9 \\
All
& \textbf{78.6} & \textbf{86.2} & \textbf{71.7} & \textbf{81.7} & 56.7 & 71.3 \\

\hline
\end{tabular}
\caption{Ablation study on semantic axes for CGR with Gemma3-12B.
Combining all axes performs best on VWSD and VerSe, while object-level
constraints perform comparably to the all-axis variant on WikiDiverse.}
\label{tab:semantic_axis_ablation}
\end{table}

\paragraph{Experiments on GAR.} Using Gemma3-12B as the base VLM, we compare five ablation settings: \textit{Base}, \textit{Generated Def.}, \textit{Lexical Def.}, \textit{CGR}, and \textit{Full GAR}. \textit{Base} uses the original context. \textit{Generated Def.} generates a context-specific definition directly from the target and context, without external lexical-resource retrieval, and serves as a generic, context-aware prompt-augmentation baseline~\citep{kwon2023vision}. \textit{Lexical Def.} retrieves lexical knowledge and synthesizes an expanded definition of the intended meaning, which is used as a holistic query. \textit{CGR} decomposes the same semantic representation into visually verifiable constraints for candidate ranking (Section~\ref{sec:gar}). \textit{Full GAR} combines \textit{Lexical Def.} and \textit{CGR} using reciprocal rank fusion (RRF; \citealp{cormack2009reciprocal}).

Table~\ref{tab:ablation_results} presents the corresponding ablation results.
\textit{Generated Def.} provides only limited and inconsistent gains across
subsets. In contrast, \textit{Lexical Def.} improves over both Base and
Generated Def. on all three subsets, indicating that the benefit is not
explained by textual augmentation alone and that explicit lexical-semantic
knowledge provides a more consistently useful grounding signal. CGR shows a
different pattern: it improves VWSD and yields particularly strong gains on
VerSe, but degrades performance on WikiDiverse. Combining Lexical Def. and
CGR achieves the strongest macro-average performance and the best result on
VWSD. Overall, these results show that \textit{holistic lexical-semantic
information transfers broadly across grounding settings, whereas decomposed
visual verification is more setting-dependent}.

We analyze the contribution of individual semantic axes in Table~\ref{tab:semantic_axis_ablation}. Object-level constraints are the strongest single axis across all three subsets. On VWSD and VerSe, combining object, attribute, and state constraints further improves performance, suggesting that these cues provide complementary evidence for noun- and
verb-centered grounding. On WikiDiverse, however, the all-axis variant does not improve over object-only reasoning, while attribute and especially state
constraints are substantially weaker. This pattern suggests that \textit{fine-grained entity grounding requires more identity-specific and discriminative evidence than is captured by the current generic semantic-axis decomposition}.

\begin{table}[t]
\centering
\scriptsize
\begin{tabular}{@{ }c@{ }|@{ }l@{ }|@{ }c@{ }c@{ }}
\hline
Setting      & \multicolumn{1}{c@{ }|@{ }}{Model} & Hits@1 & MRR \\
\hline\hline
\multirow{7}{*}{Training-free}  
    & CLIP-ViT-L/14 \citep{yang2024polclip} & 57.5 & 72.6 \\
    & UVWSD \citep{kwon2023vision}& 60.9 & 75.2 \\
    & GPT\_BLIP \citep{kritharoula2023large}    & 72.4 & 82.3 \\
    & QVWSD \citep{qiao2025quantum}        & 62.4 & 76.6 \\    
    & DCTP \citep{bhattacharya2026visual}        & 62.2 & 76.9 \\
    & Qwen2.5-VL-72B \citep{musacchio2025assessing} & 79.1 & -- \\
    & GAR-full (Ours) & 81.2 & 88.6 \\
\hline
\multirow{4}{*}{Fine-tuned} 
    & CLIP-ViT-L/14 \citep{yang2024polclip} & 76.0 & 83.9 \\
    & FCLL \citep{yang2023tam}          & 81.4 & 87.7 \\
    & ARPA \citep{papastavrou2024arpa}          & \textbf{84.7} & 88.7 \\
    & PolCLIP \citep{yang2024polclip}       & 83.6 & \textbf{90.1} \\
\hline
\end{tabular}
\caption{Comparison with previous VWSD models. Details of the models are provided in Appendix~\ref{apx:comparison_models}.}
\label{tab:comparison_SOTA}
\end{table}

\paragraph{Comparison with Prior VWSD Methods} Table~\ref{tab:comparison_SOTA} compares GAR-full with previously VWSD studies. \textit{GAR-full outperforms the compared training-free approaches}, reaching 81.2 Hits@1 and 88.6 MRR without task-specific training. Notably, it also surpasses the Hits@1 of the larger Qwen2.5-VL-72B model (79.1), despite using Gemma3-12B as its base VLM. Although GAR-full still trails the strongest fine-tuned systems, these results show that explicit semantic grounding can improve training-free VWSD performance.

\section{Discussion}

\begin{table}[t]
\centering
\small
\begin{tabular}{l|cc}
\hline
 & Hard Conf. & Rand. Conf. \\
\hline
WikiDiverse & 14.56\% & 2.98\% \\
VerSe       & 26.38\% & 11.59\% \\
\hline
\end{tabular}
\caption{Pairwise confusion analysis of hard and random negatives on Gemma3-12B. Each negative is compared individually against the gold image. Hard Conf. and Rand. Conf. denote the percentage of hard--gold and random--gold pairs, respectively, in which the negative is ranked above the gold image.}
\label{tab:hard_soft_confusion}
\end{table}

\paragraph{Are Hard Negatives More Confusable?}
Table~\ref{tab:hard_soft_confusion} validates the hard-negative construction strategy through pairwise comparisons against the gold image. Hard negatives are substantially more likely than random negatives to outrank the gold in both VerSe (26.38\% vs.\ 11.59\%) and WikiDiverse (14.56\% vs.\ 2.98\%). Because each negative is evaluated independently in a one-to-one comparison with the gold image, these rates directly measure the confusability of the two negative types. The results confirm that the semantic selection procedure produces genuinely more confusable alternatives.

\begin{table}[t]
\centering
\scriptsize
\begin{tabular}{l|ccc}
\hline
Error Type & VWSD & VerSe & WikiDiverse \\
\hline
Upstream Semantic / Referent & 26\% & 12\% & 10\% \\
Decomposition Drift          & 20\% & 18\% & 2\%  \\
Cross-Axis Conflict          & 6\%  & 6\%  & 6\%  \\
Non-Discriminative Constraint& \textbf{32\%} & \textbf{48\%} & \textbf{76\%} \\
Visual Verification          & 16\% & 16\% & 6\%  \\
\hline
\end{tabular}
\caption{Distribution (\%) of manually categorized CGR errors
(50 per subset; 150 total).}
\label{tab:cgr_error_analysis}
\end{table}

\paragraph{Error Analysis.}
To better understand CGR failures, we manually analyze 50 errors from each
grounding setting, considering the semantic representation, the
three constraint axes, and the gold and top-ranked images. We identify five
failure types: upstream semantic/referent errors, decomposition drift,
cross-axis conflicts, non-discriminative constraints, and visual verification
errors (Table~\ref{tab:cgr_error_analysis}).

Failure profiles differ across datasets. WikiDiverse is dominated by \textit{non-discriminative constraints} (76\%), where plausible questions lack the identity-specific evidence needed to distinguish close entities. VerSe shows a more mixed profile, although 48\% of errors likewise
reflect insufficiently discriminative cues for separating competing event
realizations. VWSD is more broadly distributed across upstream errors (26\%),
decomposition drift (20\%), and non-discriminative constraints (32\%). Cross-axis conflicts occur in 6\% of errors across all three subsets, suggesting occasional limitations of uniform axis averaging. Representative examples are provided in Appendix~\ref{apx:cgr_errors}.

\paragraph{Implications and Future Directions}
Our findings distinguish semantic compatibility from fine-grained discrimination. While lexical definitions consistently improve grounding, constraint-guided verification depends on whether the generated evidence distinguishes competing candidates. This is particularly evident in WikiDiverse, where generic constraints often apply to closely related entities. Future work could develop candidate-aware, contrastive constraints that emphasize distinguishing visual attributes rather than shared semantic properties. The varying contributions of object, attribute, and state constraints also motivate adaptive semantic-axis weighting based on their discriminative value. Furthermore, the complementary strengths of lexical definitions and CGR suggest more adaptive integration of holistic ranking and structured visual verification, beyond the fixed rank fusion used in GAR.

\section{Conclusion}

We presented SAGE, a benchmark spanning noun, verb, and entity-centered multimodal grounding, and GAR, a training-free framework combining definition grounding with structured semantic verification. SAGE shows that performance on noun-centered VWSD alone does not fully characterize model behavior in verb- and entity-centered grounding settings. Definition augmentation provides consistent gains across all three subsets, while constraint-guided reasoning improves VWSD, yields strong gains on VerSe, and is less effective for fine-grained entity grounding on WikiDiverse.

\section*{Ethical Considerations}
SAGE is constructed from publicly available resources, 
including VWSD, VerSe, MS-COCO, and 
WikiDiverse/Wikimedia-derived metadata.\footnote{
VWSD: CC-BY-NC 4.0; VerSe annotations: CC-BY 4.0; 
WikiDiverse: CC BY-SA 4.0.}
We will not redistribute image files; MS-COCO, Flickr, 
Wikipedia, and Wikimedia images remain subject to their 
original source-specific licenses and terms.\footnote{
Our release will include only reformulated metadata, 
candidate lists, gold indices, prompts, evaluation 
scripts, and dataset loaders.}

For human evaluation, three annotators were recruited 
and paid at or above the local minimum wage. We sampled 
a subset of LLM-generated definitions and did not 
observe any statements violating the ACL anti-harassment 
policy. However, as with any large language model 
outputs, definitions that were not explicitly vetted 
may still contain potentially harmful or offensive 
content (e.g., racist, xenophobic, or toxic language).

\paragraph{AI Assistance Disclosure} We used AI-assisted tools to support the drafting, editing, and refinement of manuscript text. The authors reviewed and verified the resulting content and retained full responsibility for the study design, experiments, analyses, interpretations, and final manuscript.

\section*{Limitations}

As with established WSD, retrieval, and VLM benchmarks \citep{raganato2017word,thakur2021beir, yue2024mmmu,yue2025mmmu}, SAGE provides a shared evaluation protocol without fully equalizing difficulty across heterogeneous settings. Although SAGE standardizes the 10-candidate image-selection format, candidate confusability, contextual information, and semantic granularity may still vary across datasets. Thus, cross-setting performance differences should not be interpreted as intrinsic differences in grounding difficulty. Nevertheless, the shared protocol supports systematic comparisons of model behavior across complementary lexical grounding settings.

SAGE also inherits some characteristics of its source datasets. WikiDiverse representative images and SPIS VerSe examples may occasionally admit more than one plausible visual interpretation. However, our human analysis indicates that such cases are uncommon: potential representation ambiguity was identified in 4.0\% of sampled WikiDiverse instances and multiple plausible same-sense realizations in 3.0\% of sampled VerSe instances. Detailed annotation statistics and examples are provided in Appendix~\ref{apx:human_analysis}.

Although SPIS reveals differences in model sensitivity to target-verb information obscured by original caption--image pairing, it does not constitute a fully controlled causal evaluation. Replacing the gold image also changes objects, participants, and scene context, potentially influencing model predictions. Future work could extend this diagnostic through more controlled interventions that systematically vary verb, noun, and scene information, enabling finer-grained analysis of their respective contributions to multimodal grounding.

SAGE is currently limited to English, and lexical ambiguity, sense inventories, and entity expressions may differ across languages and cultures. Extending the benchmark to multilingual settings remains future work.

 GAR depends on the quality of generated semantic representations and visual constraints. Our analysis shows that overly generic constraints can be less effective for fine-grained entity discrimination, motivating more candidate-aware and contrastive constraint generation. GAR also incurs additional inference cost from constraint-based visual verification; in our VWSD runtime analysis, CGR takes approximately 3.7$\times$ as long as base pointwise scoring. Detailed runtime results are provided in Appendix~\ref{apx:runtime_comparison}. 

\section*{Acknowledgment}
We thank Dr. Katrin Erk for her valuable feedback and insightful suggestions that helped improve this work.
 
\bibliography{custom}

\newpage
\appendix
\onecolumn

\section{Lexical Category Distribution in VWSD}
\label{apx:pos_distribution}

\begin{table}[!h]
\centering
\small
\begin{tabular}{@{}l|ccccc@{}}
\hline
     & VERB & NOUN & PROPN & Others & All \\
\hline\hline
Count   & 12   & 392  & 54    & 5  & 463 \\
Ratio   & 2.59\% & 84.67\% & 11.66\% & 1.08\% & 100.00\% \\
\hline
\end{tabular}
\caption{Distribution of POS categories in the VWSD test split.}
\label{tab:pos_distribution}
\end{table}

Table~\ref{tab:pos_distribution} reports the part-of-speech distribution of the VWSD test split used in our analysis. The distribution shows that VWSD is heavily noun-centered, with nouns accounting for 84.67\% of the test instances and verbs accounting for only 2.59\%.

\section{Evaluation Metrics}
\label{apx:metircs}
Our benchmark follows the evaluation protocol of the VWSD task, which formulates the problem as a multimodal image retrieval task. 
Unlike traditional word sense disambiguation (WSD), which is typically evaluated using classification metrics such as F1 score \citep{dongsuk2018word}, our setting requires ranking a set of candidate images according to their compatibility with a target word and its contextual description. 
Therefore, retrieval-oriented ranking metrics are more appropriate for this task. 
Following prior work on VWSD, we report \textbf{Hits@1} (equivalent to Accuracy) and \textbf{Mean Reciprocal Rank (MRR)} as our primary evaluation metrics.

\paragraph{Hits@1} measures whether the correct image is ranked at the top position among the candidate images. 
Let $N$ denote the number of evaluation instances and let $r_i$ denote the rank of the correct image for the $i$-th instance. 
Hits@1 is defined as:

\begin{equation}
\text{Hits@1} = \frac{1}{N} \sum_{i=1}^{N} \mathbb{I}(r_i = 1)
\end{equation}

where $\mathbb{I}(\cdot)$ is an indicator function that equals 1 when the condition holds and 0 otherwise. 
In this context, Hits@1 is equivalent to the standard Accuracy metric used in retrieval-based benchmarks.

\paragraph{Mean Reciprocal Rank (MRR)} evaluates the average inverse rank of the correct image across all evaluation instances, capturing how highly the correct image is placed in the ranked list even when it is not ranked first. 
MRR is defined as:

\begin{equation}
\text{MRR} = \frac{1}{N} \sum_{i=1}^{N} \frac{1}{r_i}
\end{equation}

where $r_i$ denotes the rank of the correct image for the $i$-th instance. 
Higher MRR values indicate that the correct images tend to appear closer to the top of the ranked candidate list.

\section{Details on SAGE Benchmark Dataset}
\label{apx:SAGE_details}

\subsection{Examples on VerSe}
\label{apx:verse_details}
\begin{figure*}[!ht]
    \centering
    \includegraphics[width=.7\linewidth]{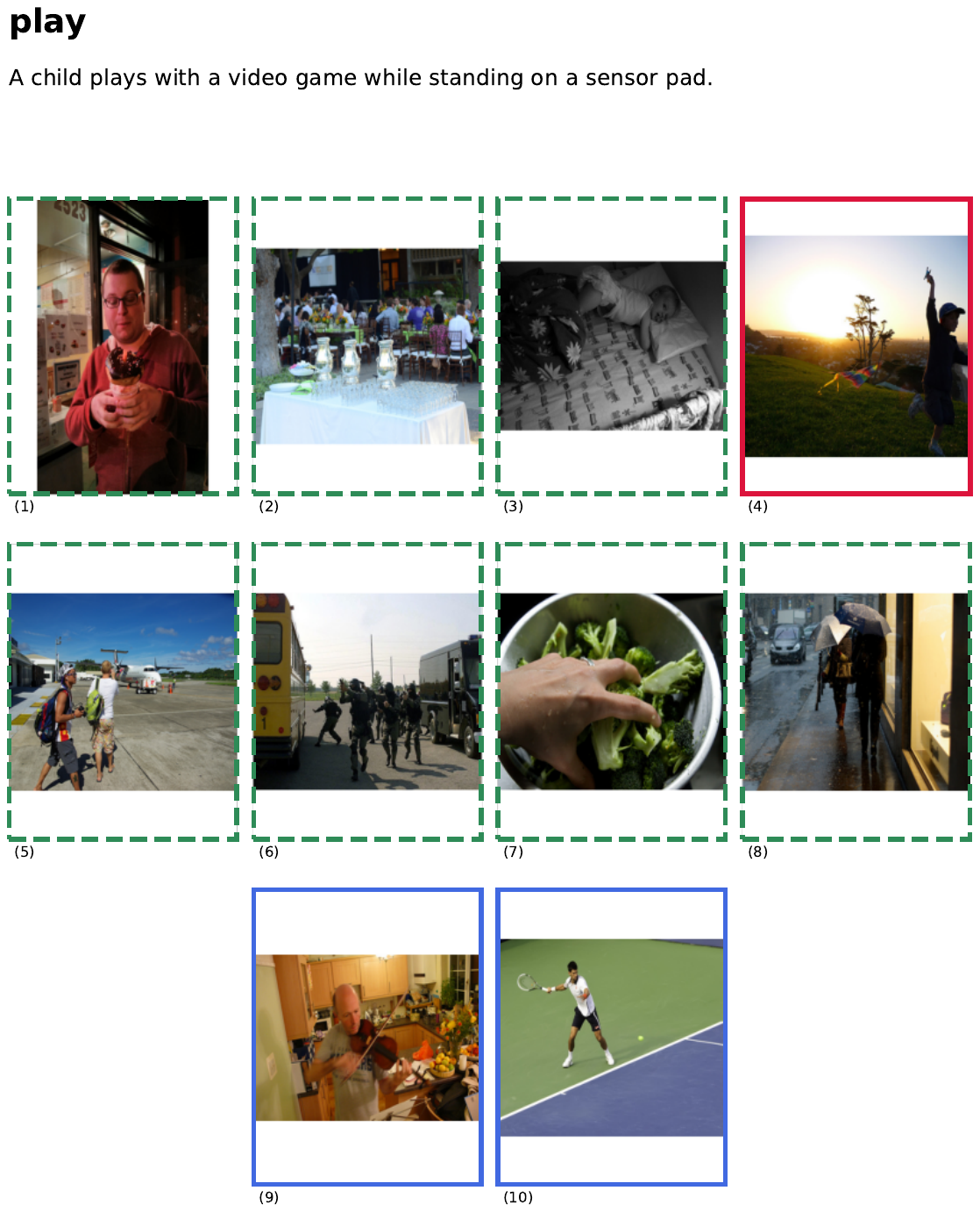}
    \caption{An example of ``play'' in VerSe. 
Borders indicate labels: \textcolor{red}{red = correct}, 
\textcolor{hardneg}{blue = hard negatives}, 
\textcolor{randomneg}{green = random negatives}.}
    \label{fig:play}
    
\end{figure*}

\begin{figure*}[!ht]
    \centering
    \includegraphics[width=.7\linewidth]{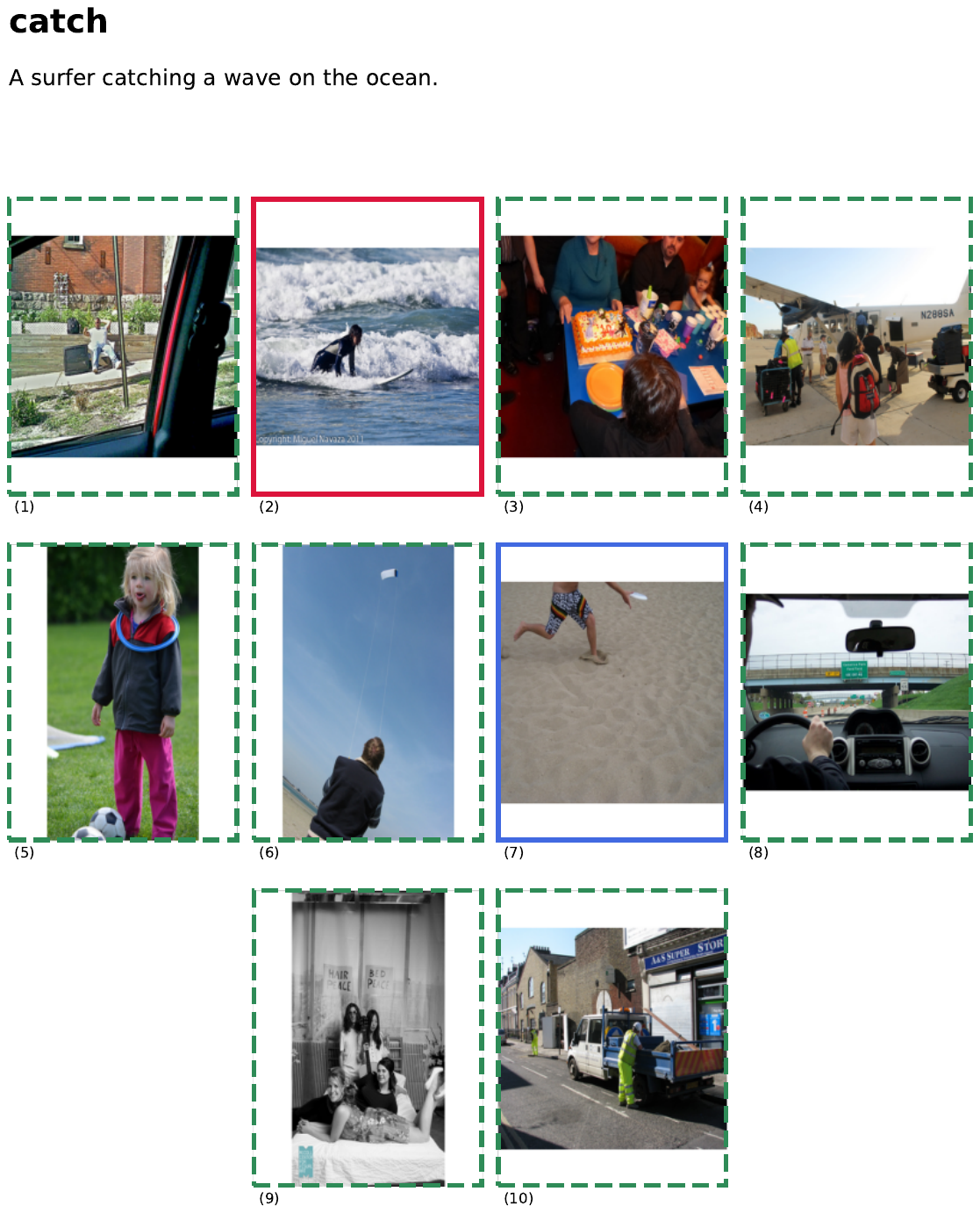}
    \caption{An example of ``catch'' in VerSe. 
Borders indicate labels: \textcolor{red}{red = correct}, 
\textcolor{hardneg}{blue = hard negatives}, 
\textcolor{randomneg}{green = random negatives}.}
    \label{fig:catch}
    
\end{figure*}

This section illustrates VerSe's candidate design through two
per-topic panels: \textit{play} (Figure~\ref{fig:play}) and \textit{catch} (Figure~\ref{fig:catch}). Each panel bundles the correct target with distractors, and uses border colors to denote labels: \textcolor{correct}{\textbf{red = correct}}, \textcolor{hardneg}{\textbf{blue = hard negatives}}, \textcolor{randomneg}{\textbf{green = random negatives}}. A correct item uniquely (or within a small correct set) satisfies the constraints of the meaning of the target word in a query.

\paragraph{Hard vs.\ Random Negatives}
\textcolor{hardneg}{\textbf{Hard negatives}} are confusable because they reuse the same surface verb (\emph{polysemy}) or exhibit close event semantics while violating the intended \emph{sense}. 
In the \textit{play} panel, \textbf{(9) violin performance} and \textbf{(10) tennis} use the same verb ``play'' but in the \emph{music} and \emph{sport/competition} senses, not the recreational/childlike sense indicated by the context. 
In the \textit{catch} panel, \textbf{(7)} visually suggests the ``grab'' sense of \emph{catch} but lacks both the \emph{surfboard} (instrument) and the \emph{ocean wave} (setting); hence it is a hard negative for ``catch a \emph{wave}.''
By contrast, \textcolor{randomneg}{\textbf{random negatives}} (e.g., buffet tables, airport ramp, cleaning, salad mixing, indoor party, street-cleaning truck, car interior) are semantically distant and provide little visual or conceptual evidence for the queried action; they are easy to reject.

\paragraph{Grounded Event-Level Reasoning is Essential}
Solving verb panels cannot rely on keyword overlap or isolated object tags alone. Correct decisions require: (i) \emph{verb-sense disambiguation} (e.g., \textit{play} as childlike recreation vs.\ music/sport; \textit{catch} as hand-grab vs.\ catch a wave), (ii) \emph{event-role constraints} that align agent, patient, instrument, and setting, (iii) \emph{affordances, pose, and motion cues} indicating feasible actions, and (iv) \emph{scene and phase cues} that bias plausible readings. In short, these tasks require integrated reasoning over textual context, visual evidence, and grounded event semantics.

\newpage

\noindent\textbf{Correct.}
(4) A child flying a kite on a grassy hill — a prototypical \emph{childlike recreational activity}, matching the intended sense of \textit{play}.

\noindent\textbf{Hard negatives (\textcolor{hardneg}{blue}).}
\begin{itemize}
  \item (9) \textcolor{hardneg}{Playing the violin} — verb “play” in the \emph{music} sense (\emph{perform or transmit music}); not the childlike/recreational sense.
  \item (10) \textcolor{hardneg}{Playing tennis} — verb “play” in the \emph{sport/competition} sense (\emph{engage in or make moves related to competition}); not the target sense.
\end{itemize}

\noindent\textbf{Random negatives (\textcolor{randomneg}{green}).}
\begin{itemize}
  \item (1) Person holding food — not gameplay or childlike recreational activity.
  \item (2) Outdoor buffet tables — no play context.
  \item (3) Child lying — creative/academic activity, not the target sense.
  \item (5) People are walking — travel/logistics, not play.
  \item (6) Tactical officers moving: law-enforcement action, not play.
  \item (7) Grabbing salad: food preparation.
  \item (8) People are walking.
\end{itemize}

\noindent\textbf{Correct.}
(2) A surfer riding a breaking wave — clearly “catching” a wave on the ocean; matches the action.

\noindent\textbf{Hard negatives (\textcolor{hardneg}{blue}).}
\begin{itemize}
  \item (7) \textcolor{hardneg}{Person reaching to a disk}: appears to ``catching'' \emph{by hand}, but there is no surfboard or wave; not \emph{catching a wave}.
\end{itemize}

\noindent\textbf{Random negatives (\textcolor{randomneg}{green}).}
\begin{itemize}
  \item (1) A person is sitting: not ocean/surfing.
  \item (3) Indoor birthday party with cake: unrelated scene.
  \item (4) Airport gate with luggage: no ocean or surfer.
  \item (5) A child is playing: sports, not surfing.
  \item (6) Kite flying: outdoor but not surfing.
  \item (8) Driving: no ocean context.
  \item (9) Photobooth family portrait: unrelated.
  \item (10) Street cleaning truck: unrelated to surfing.
\end{itemize}

\newpage
\subsection{Examples on WikiDiverse}
\label{apx:wikidiverse_details}

\begin{figure*}[!ht]
    \centering
    \includegraphics[width=.7\linewidth]{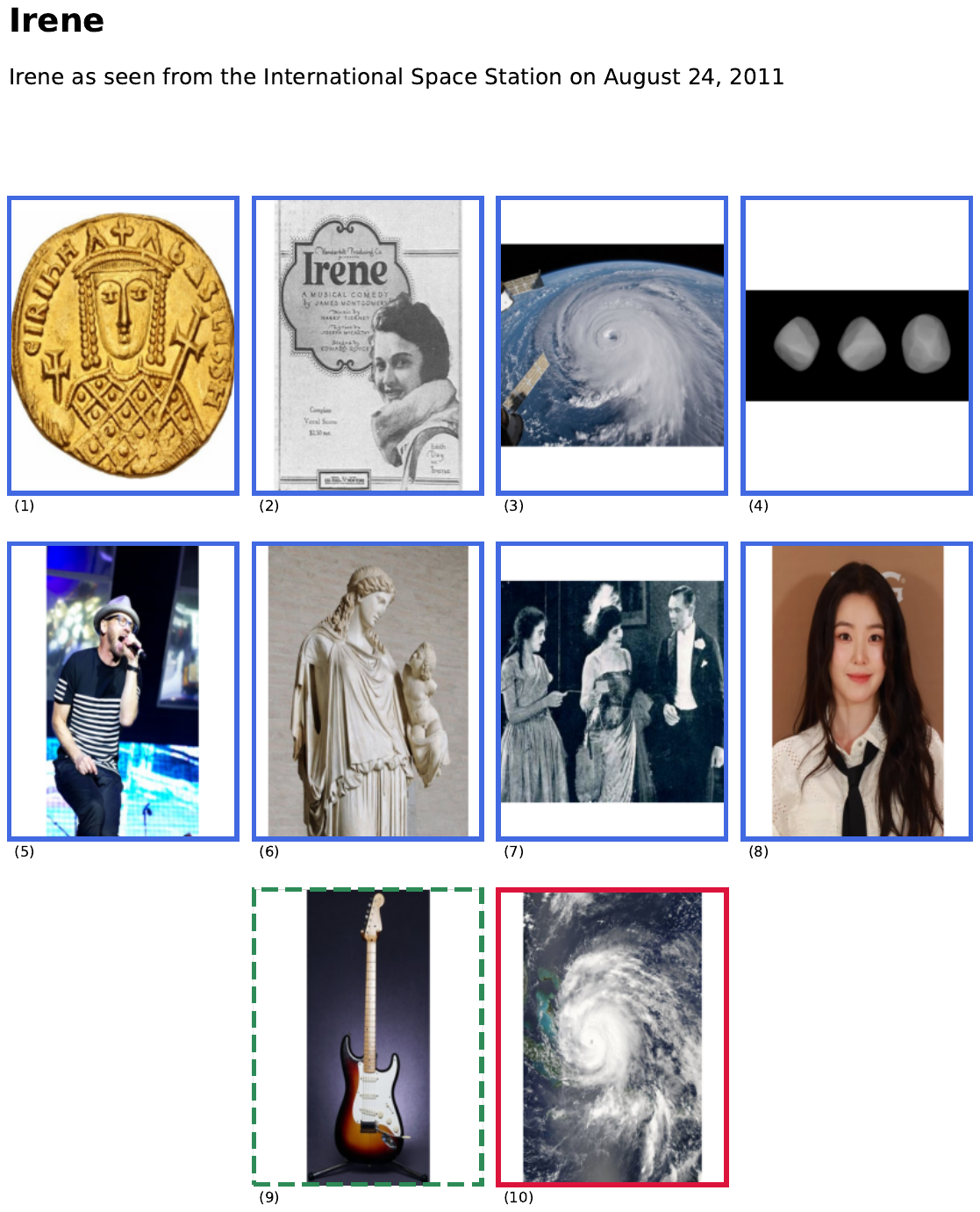}
    \caption{An example of ``Irene'' in WikiDiverse. 
Borders indicate labels: \textcolor{red}{red = correct}, 
\textcolor{hardneg}{blue = hard negatives}, 
\textcolor{randomneg}{green = random negatives}.}
    \label{fig:irene}
    
\end{figure*}

\begin{figure*}[!ht]
    \centering
    \includegraphics[width=.7\linewidth]{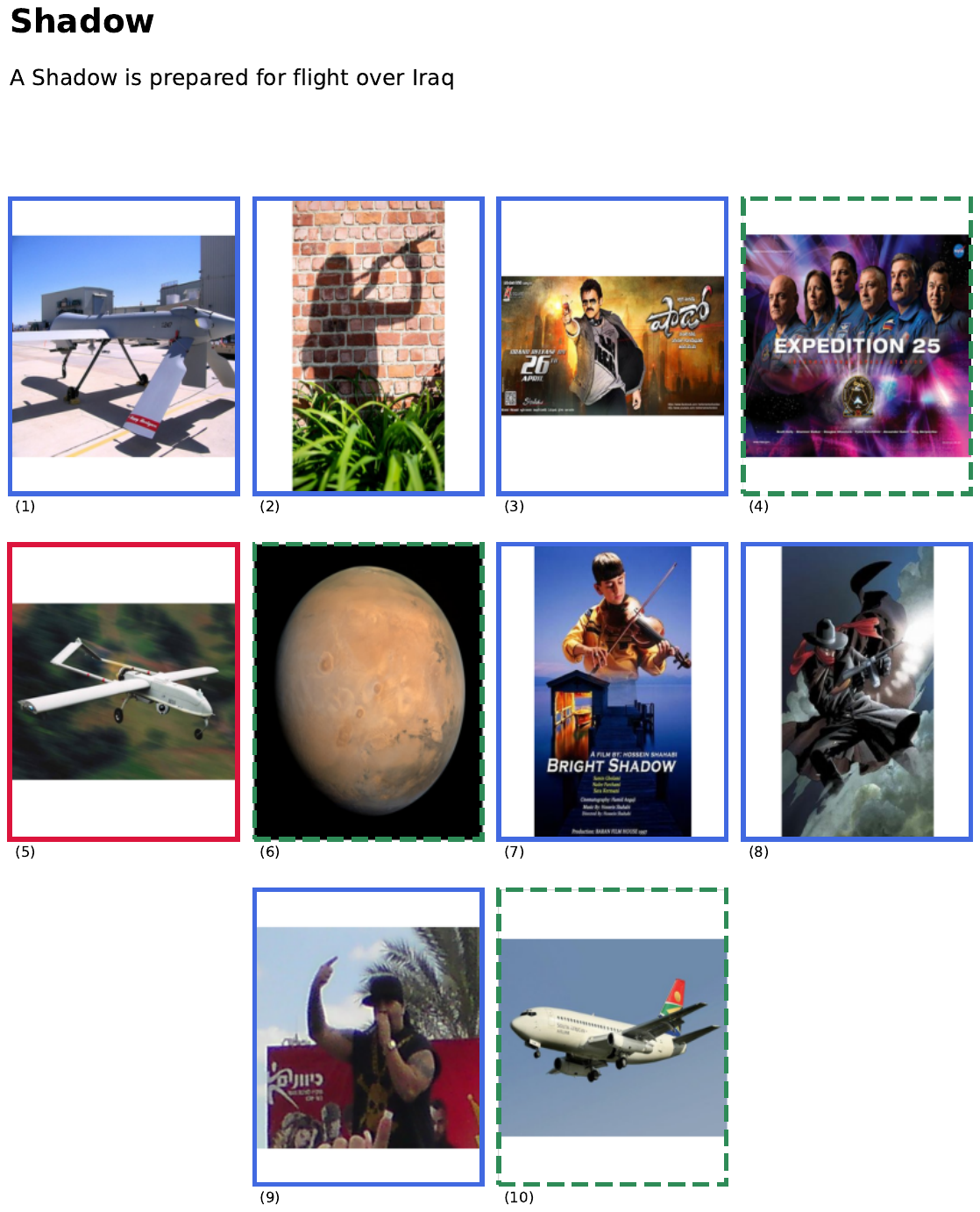}
    \caption{An example of ``Shadow'' in WikiDiverse.  
Borders indicate labels: \textcolor{red}{red = correct}, 
\textcolor{hardneg}{blue = hard negatives}, 
\textcolor{randomneg}{green = random negatives}.}
    \label{fig:shadow}
\end{figure*}

This section illustrates WikiDiverse's candidate design through two
per-topic panels: \textit{Irene} (Figure~\ref{fig:irene}) and \textit{Shadow} (Figure~\ref{fig:shadow}). Each panel bundles the correct target with distractors, and uses border colors to denote labels: \textcolor{correct}{\textbf{red = correct}}, \textcolor{hardneg}{\textbf{blue = hard negatives}}, \textcolor{randomneg}{\textbf{green = random negatives}}. A \textit{correct} item uniquely (or within a small correct set) satisfies the query with respect to entity or phenomenon identity and, when relevant, time.

\paragraph{Hard vs.\ Random Negatives}
\textcolor{hardneg}{\textbf{Hard negatives}} are confusable distractors due to surface name overlap, categorical proximity (same family of phenomena/objects), or strong visual similarity. In the \textit{Irene} panel, for instance, \textit{Hurricane Florence} shares the same phenomenon type as \textit{Hurricane Irene} but differs in storm identity and date. Likewise, items named \textit{Irene} that are actually an \emph{asteroid (14~Irene)} or \emph{creative works/people} belong to different categories and are thus hard negatives. In the \textit{Shadow} panel, non-target UAVs (e.g., Predator family) or homonymous film/character/artist entries are treated as hard negatives. In contrast, \textcolor{randomneg}{\textbf{random negatives}} are semantically distant items with little visual or conceptual evidence for the query (e.g., planetary imagery, mission posters, commercial airliners).

\paragraph{World Knowledge is Essential}
Solving these panels cannot rely on keyword matching or low-level visual cues alone. Correct decisions require: (i) meteorological concepts such as the \emph{eye/eyewall/rainbands} of a hurricane, and (ii) aeronautical knowledge to distinguish airframes (\textbf{RQ-7 Shadow} twin-boom vs.\ \textbf{MQ-1/Predator} single-boom with inverted V-tail), (iii) etymology and naming variants (Greek goddess \emph{Eirene}; Latinized form \emph{Irene}), and (iv) cultural/media knowledge that \emph{The Shadow} may denote films, characters, or artists rather than aircraft. In short, these tasks demand \emph{integrated reasoning over text–vision signals and world knowledge}.
\newpage

\noindent\textbf{Correct.}
(10) The image of \textcolor{red}{Hurricane \textit{Irene}} (Entity type: `Other').  
This image uniquely matches the query in phenomenon (tropical cyclone Irene) considering its name and date.

\noindent\textbf{Hard negatives (\textcolor{hardneg}{blue}).}
These items are strongly confusable because they share the name “Irene” or exhibit visual/semantic similarity to the target, despite not satisfying the query:
\begin{itemize}
  \item (3) Hurricane \textit{Florence} from the ISS in 2018: \textcolor{hardneg}{same phenomenon and viewpoint}, but \emph{different storm and year}.
  \item (1), (2), (4), (7), (8) name matches: Empress \textcolor{hardneg}{\textit{Irene}} on a Byzantine coin; the 1919 Broadway musical \textcolor{hardneg}{\textit{Irene}}; \textcolor{hardneg}{Asteroid 14~Irene} - identical name but a celestial body; a film titled \textcolor{hardneg}{\textit{Irene}}; and the Korean pop singer \textcolor{hardneg}{\textit{Irene}}. None are a satellite/ISS view of a storm.
  \item (5) \textcolor{hardneg}{TobyMac performance} - included due to his 2003 song ``Irene.''
  \item (6) Classical statue of the Greek goddess \textcolor{hardneg}{\textit{Eirene}} (Latin: \textit{Irene}).
\end{itemize}

\noindent\textbf{Random negative (\textcolor{randomneg}{green}).}
(9) \textcolor{randomneg}{Fender Stratocaster} (Entity type: `Other') - semantically unrelated to the query; easy to reject (no name or visual similarity to a hurricane/ISS scene).

\newpage

\noindent\textbf{Correct.}
(5) \textcolor{correct}{AAI RQ-7 \textit{Shadow}} (Entity type: `Other') UAV in flight — matches the query “Shadow (drone)” and the caption context (“A Shadow is prepared for flight over Iraq”).

\noindent\textbf{Hard negatives (\textcolor{hardneg}{blue}).}
Visual/lexical confounders that look related or share the name “Shadow” but do not satisfy the query:
\begin{itemize}
  \item (1) \textcolor{hardneg}{MQ-1B/Predator-style UAV on ramp}: a different GA-ASI airframe with a single fuselage boom and inverted V-tail, larger wingspan, and SATCOM/EO–IR turret layout typical of the Predator line; \emph{not} the twin-boom AAI RQ-7 \textit{Shadow}.
  \item (2) \textcolor{hardneg}{Literal human \emph{shadow} on a wall}: a cast silhouette (person holding an object, e.g., a flute) used as a visual motif; matches the word “shadow” only, with no aircraft or military context.
  \item (3) \textcolor{hardneg}{Film poster \emph{Shadow}}: commercial cinema artwork (live-action film); lexical overlap with the query but no UAV imagery, airframe, or operational cues.
  \item (7) \textcolor{hardneg}{Film poster \emph{Bright Shadow}}: title includes “Shadow” as part of a longer phrase; again a movie promotion graphic, not a drone or ISR platform.
  \item (8) \textcolor{hardneg}{\emph{The Shadow} (fictional character) artwork}: pulp/superhero vigilante depiction (cape/hat), a named character unrelated to aviation; confusion arises purely from the homonymous name.
  \item (9) \textcolor{hardneg}{\emph{The Shadow} (Yoav Eliasi) performing}: Israeli rapper’s stage name ``The Shadow''; performance/photojournalism context with no UAV or military hardware present.
\end{itemize}

\noindent\textbf{Random negatives (\textcolor{randomneg}{green}).}
These semantically distant items are easy to reject; all random negatives have the same entity type, ``Other'':
\begin{itemize}
  \item (4) \textcolor{randomneg}{NASA Expedition~25 poster}: space mission artwork, not a drone.
  \item (6) \textcolor{randomneg}{Mars global view}: planetary image, unrelated to aircraft.
  \item (10) \textcolor{randomneg}{Boeing 737-200 airliner}: manned commercial jet, not a UAV.
\end{itemize}






\newpage
\section{Human Validation of Benchmark Construction and Error Analysis}
\label{apx:human_analysis}

\subsection{Annotation Protocol and Interface}
\label{apx:annotation_protocol}

We conducted a human audit to assess the reliability of the automatically constructed gold labels and candidate sets for the VerSe and WikiDiverse reformulations.

\paragraph{Annotators.}
Three graduate students independently annotated the sampled instances.

\paragraph{Interface and procedure.}
Annotators used a standardized interface presenting the textual context, the target word or mention, its corresponding sense or entity description, and the complete set of 10 candidate images. For each instance, annotators selected the image that best represented the intended meaning of the target in context. For VerSe, annotators selected the candidate image that best instantiated the specified verb sense. For WikiDiverse, they selected the image corresponding to the specified target entity. Example annotation panels are shown in Figures~\ref{fig:verse_mismatch_lean} and~\ref{fig:wiki_human_knowledge}. This annotation procedure was designed to validate the automatically constructed gold images rather than to estimate human performance under the model input setting.

\paragraph{Inter-annotator agreement.}
Table~\ref{tab:annotation_agreement_composition} summarizes the agreement composition among the three independent annotators. Fleiss'~$\kappa$ was computed directly from the original, unaggregated annotations over all 200 instances per dataset. Agreement was strong for both VerSe ($\kappa=0.801$) and WikiDiverse ($\kappa=0.823$).

\begin{table}[h]
\centering
\small
\begin{tabular}{lcccc}
\toprule
\textbf{Dataset}
& \textbf{Unanimous}
& \textbf{2-of-3 agreement}
& \textbf{All different}
& \textbf{Fleiss' $\kappa$} \\
\midrule
VerSe
& 150/200 (75.0\%)
& 43/200 (21.5\%)
& 7/200 (3.5\%)
& 0.801 \\
WikiDiverse
& 155/200 (77.5\%)
& 40/200 (20.0\%)
& 5/200 (2.5\%)
& 0.823 \\
\bottomrule
\end{tabular}
\caption{
Agreement composition among the three independent annotators.
Unanimous agreement indicates that all three annotators selected the same image; 2-of-3 agreement indicates that exactly two annotators selected the same image; and all different indicates that the three annotators selected three distinct images. Fleiss'~$\kappa$ was computed from the original annotations before aggregation.
}
\label{tab:annotation_agreement_composition}
\end{table}

\paragraph{Aggregation and agreement with the SAGE gold.}
For analyses requiring a single aggregate human label, we aggregated the three independent annotations by majority vote. When at least two annotators selected the same image, the majority choice was used as the aggregate label. When all three annotators selected different images, one of the three annotated choices was sampled uniformly at random, following \citet{snow2008cheap}.

To avoid making the reported human--gold agreement dependent on random
tie-breaking, we exclude the all-different cases from this calculation. The resulting majority-vote labels agreed with the SAGE gold labels for 176 of 193 VerSe instances (91.2\%) and 185 of 195 WikiDiverse instances (94.9\%). Together with the inter-annotator agreement results, these findings indicate that the reformulated datasets provide reliable gold labels for evaluation and analysis.

The remaining disagreements typically involved visually plausible
alternatives, indirect but semantically related gold images, or fine-grained sense and referent distinctions requiring additional world knowledge. These cases primarily reflect the inherent difficulty of multimodal sense and entity grounding rather than systematic errors in the reformulation process.

\newpage
\subsection{VerSe Mismatch Analysis}
\label{apx:mismatch_verse}

Annotators matched the SAGE gold labels for 177 out of 200 sampled VerSe instances (88.5\%). We manually inspected the remaining 23 human--gold mismatches. Unlike WikiDiverse, where mismatches often involve referent or entity-representation ambiguity, VerSe mismatches primarily reflect event-level ambiguity under the SPIS setting. In this setting, the SAGE gold image is selected to preserve the same OntoNotes verb sense, rather than to reproduce the original caption-paired participants, objects, or scene details. As a result, some disagreements reflect a tension between caption-specific visual matching and sense-level verb grounding.

We group the VerSe mismatches into four broad categories: caption-specific versus sense-level match, competing sense/event cue, multiple plausible same-sense realizations, and weak or ambiguous visual evidence. These categories should be interpreted as a qualitative analysis of the residual disagreement cases, rather than as a claim that the categories are exhaustive. Overall, the high human--gold agreement suggests that the SPIS VerSe reformulation is reliable for evaluation, while the remaining mismatches reveal expected ambiguities introduced by evaluating verb-sense grounding beyond exact caption--image compatibility.

\begin{table}[h]
\centering
\small
\begin{tabular}{p{0.28\linewidth} p{0.52\linewidth} c}
\toprule
\textbf{Mismatch category} & \textbf{Description} & \textbf{Support} \\
\midrule
Caption-specific vs. sense-level match
& Annotators select a candidate that better matches specific participants, objects, or scene details in the original caption, while the SAGE gold preserves the same OntoNotes verb sense under a different visual realization.
& 7/23 (30.4\%) \\
\midrule
Competing sense/event cue
& A non-gold candidate evokes a different sense or event interpretation of the same target verb, creating sense-level competition.
& 4/23 (17.4\%) \\
\midrule
Multiple plausible same-sense realizations
& More than one candidate plausibly instantiates the same coarse OntoNotes verb sense or a closely related event configuration.
& 6/23 (26.1\%) \\
\midrule
Weak or ambiguous visual evidence
& The intended verb event is visually underspecified, weakly depicted, or difficult to verify from the available candidate images.
& 6/23 (26.1\%) \\
\bottomrule
\end{tabular}
\caption{Breakdown of the 23 human--gold mismatches in the VerSe validation sample. Support is computed over the mismatched cases.}
\label{tab:verse_mismatch_categories}
\end{table}

\begin{figure*}[!ht]
    \centering
    \includegraphics[width=.7\linewidth]{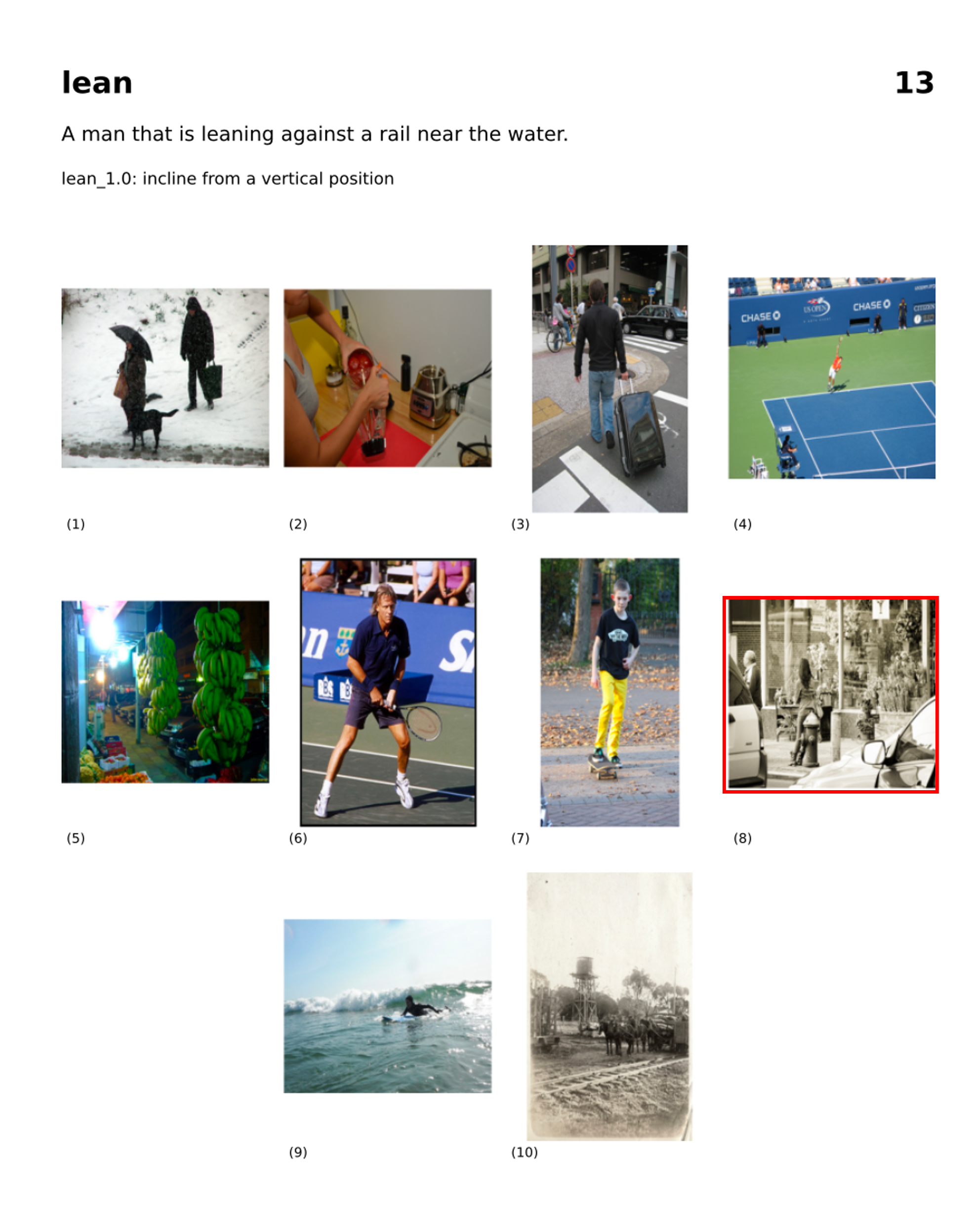}
    \caption{
Example of caption-specific versus sense-level matching in VerSe.
The target verb is ``lean,'' defined as inclining from a vertical position.
Both the gold and alternative candidates can evoke the broad sense of leaning or being inclined, but they differ in the visual realization of the event and its participants.
This illustrates a residual tension in the SPIS setting: the gold preserves the verb sense, while an alternative may appear closer to the original caption details.
The red border indicates the SAGE gold label.
}
    \label{fig:verse_mismatch_lean}
\end{figure*}

\begin{figure*}[!ht]
    \centering
    \includegraphics[width=.7\linewidth]{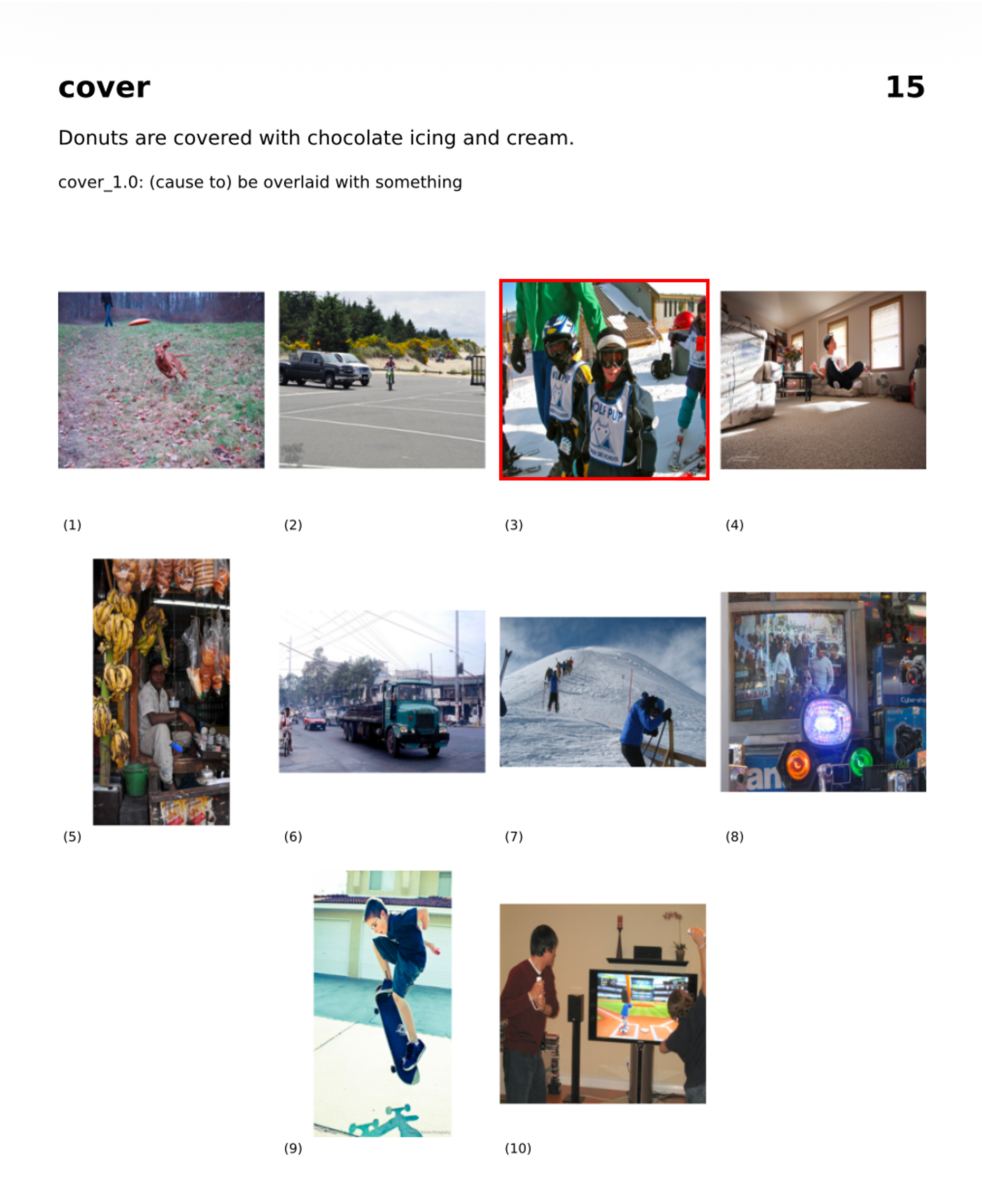}
    \caption{
Example of a competing sense/event cue in VerSe.
The target verb is ``cover'' in the sense of being overlaid with something, as in donuts covered with icing.
A competing candidate can evoke a different protective or shielding sense of ``cover,'' making the sense distinction visually ambiguous.
The red border indicates the SAGE gold label.
}
    \label{fig:verse_mismatch_cover}
\end{figure*}

\begin{figure*}[!ht]
    \centering
    \includegraphics[width=.7\linewidth]{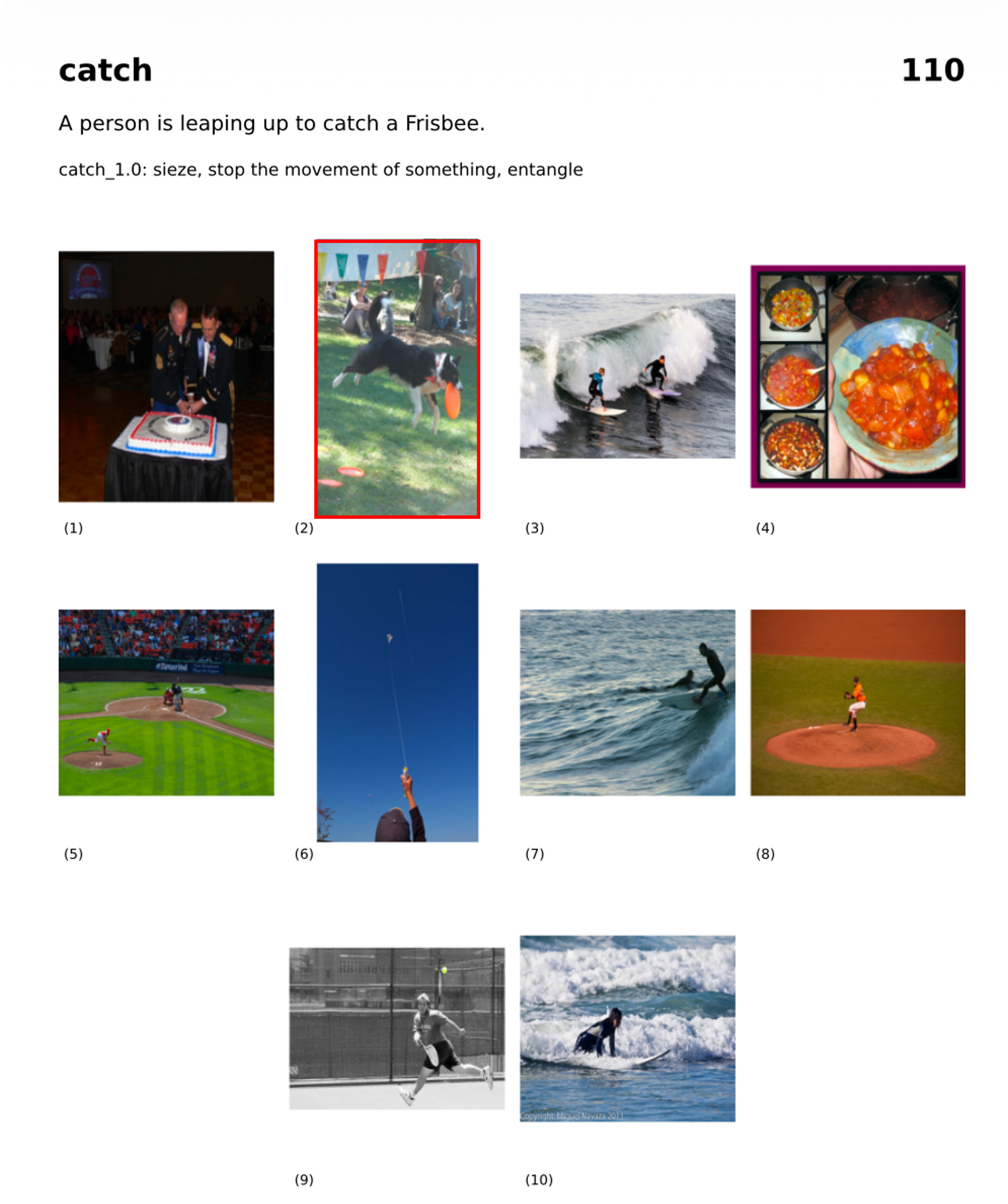}
    \caption{
Example of multiple plausible same-sense realizations in VerSe.
The target verb is ``catch'' in the sense of seizing or stopping the movement of something.
Several candidates depict related catching or intercepting events with different participants and settings, making the verb sense visually plausible in more than one image.
The red border indicates the SAGE gold label.
}
    \label{fig:verse_mismatch_catch}
\end{figure*}

Figures~\ref{fig:verse_mismatch_lean}--\ref{fig:verse_mismatch_catch}
show representative examples from mismatch categories. Together, these examples illustrate that VerSe disagreements are primarily event-level: annotators must determine not only whether a related object or action appears, but whether the depicted event instantiates the intended verb sense under the SPIS evaluation setting.

\newpage

\newpage


\newpage

\newpage
\subsection{WikiDiverse Mismatch Analysis}
\label{apx:mismatch_wikidiverse}

Annotators matched the SAGE gold labels for 188 out of 200 sampled WikiDiverse instances (94.0\%). To better understand the remaining disagreements, we manually inspected all 12 human--gold mismatches. As shown in Table~\ref{tab:wikidiverse_mismatch_categories}, the mismatches fall into three broad categories: fine-grained knowledge requirement, benchmark label ambiguity, and representation ambiguity.

These categories reflect the specific challenges of entity-centric image selection. Fine-grained knowledge requirement refers to cases where the gold image is valid, but selecting it requires geographic, domain-specific, or relational world knowledge. Benchmark label ambiguity covers cases where the selected gold image is difficult to reconcile with the textual context, or where another candidate is also clearly valid under the target mention and context. Representation ambiguity captures cases where the same entity can be visualized in multiple plausible ways, such as a physical building, an institution, a logo, a team, an event, or an indirectly related scene. This last category also reflects a limitation of using Wikipedia page images as entity representations: thumbnails or top images are not always the uniquely most representative visual grounding for an entity.

Overall, the 94.0\% agreement suggests that the WikiDiverse reformulation is broadly reliable for evaluation, while the residual mismatches highlight expected ambiguity in referent-level entity grounding and limitations of single-image entity representation.

\begin{table}[h]
\centering
\small
\begin{tabular}{p{0.28\linewidth} p{0.52\linewidth} c}
\toprule
\textbf{Mismatch category} & \textbf{Description} & \textbf{Support} \\
\midrule
Fine-grained knowledge requirement
& The gold image is plausible, but correct identification requires fine-grained geographic, domain-specific, or relational world knowledge.
& 4/12 (33.3\%) \\
\midrule
Benchmark label ambiguity
& The gold image conflicts with the textual context, or another candidate image is also clearly valid under the given target and context.
& 3/12 (25.0\%) \\
\midrule
Representation ambiguity
& The target entity can be visually represented in multiple plausible ways, such as a country, team, organization, logo, building, event, or related scene.
& 5/12 (41.7\%) \\
\bottomrule
\end{tabular}
\caption{Breakdown of the 12 human--gold mismatches in the WikiDiverse validation sample. Support is computed over the mismatched cases.}
\label{tab:wikidiverse_mismatch_categories}
\end{table}

\begin{figure*}[!ht]
    \centering
    \includegraphics[width=.7\linewidth]{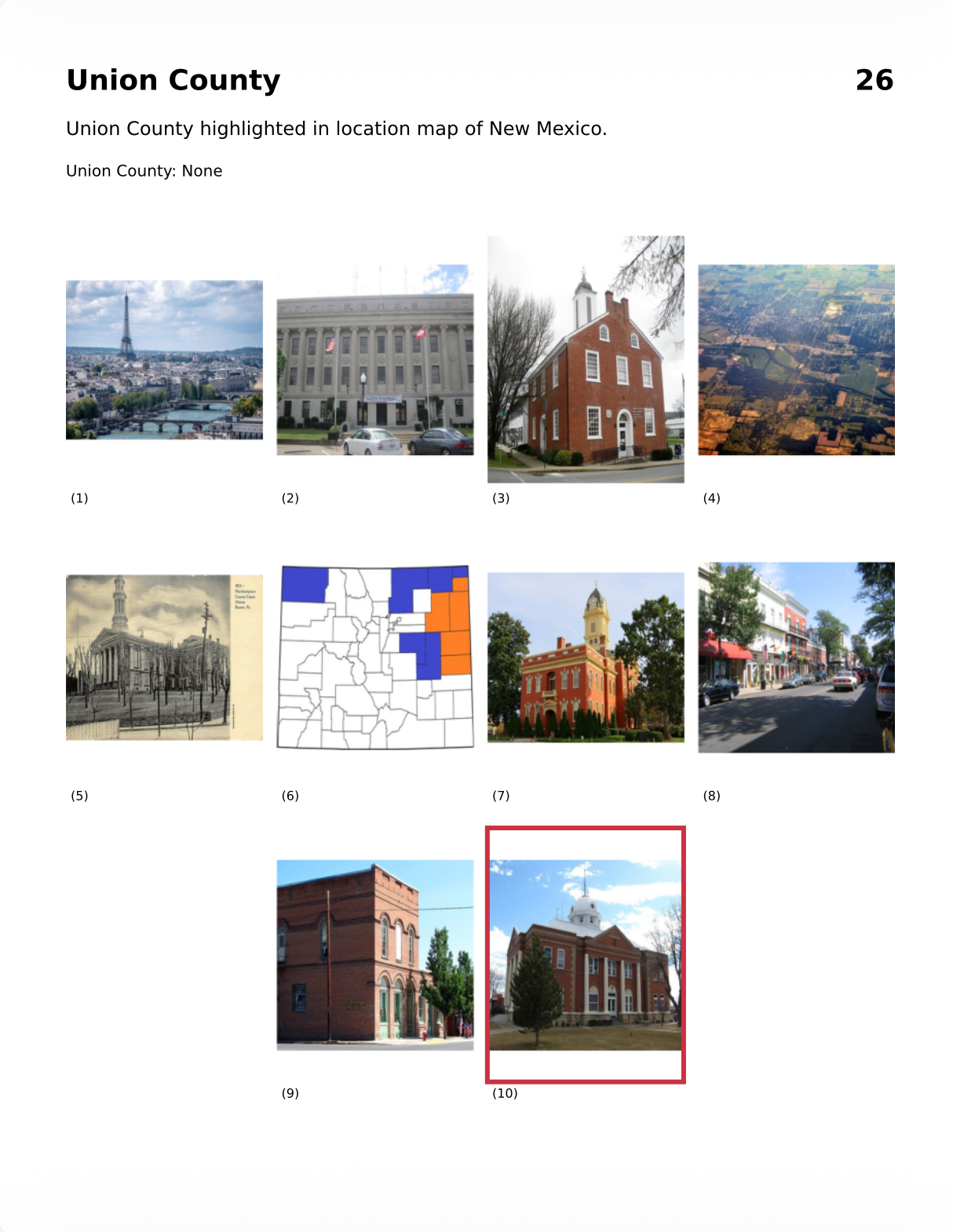}
    \caption{
Example of fine-grained knowledge requirement in WikiDiverse.
Although several candidates depict plausible county buildings or maps, the context specifies Union County in New Mexico, and the gold image is the corresponding New Mexico referent.
The red border indicates the SAGE gold label.
}
    \label{fig:wiki_human_knowledge}
\end{figure*}

\begin{figure*}[!ht]
    \centering
    \includegraphics[width=.7\linewidth]{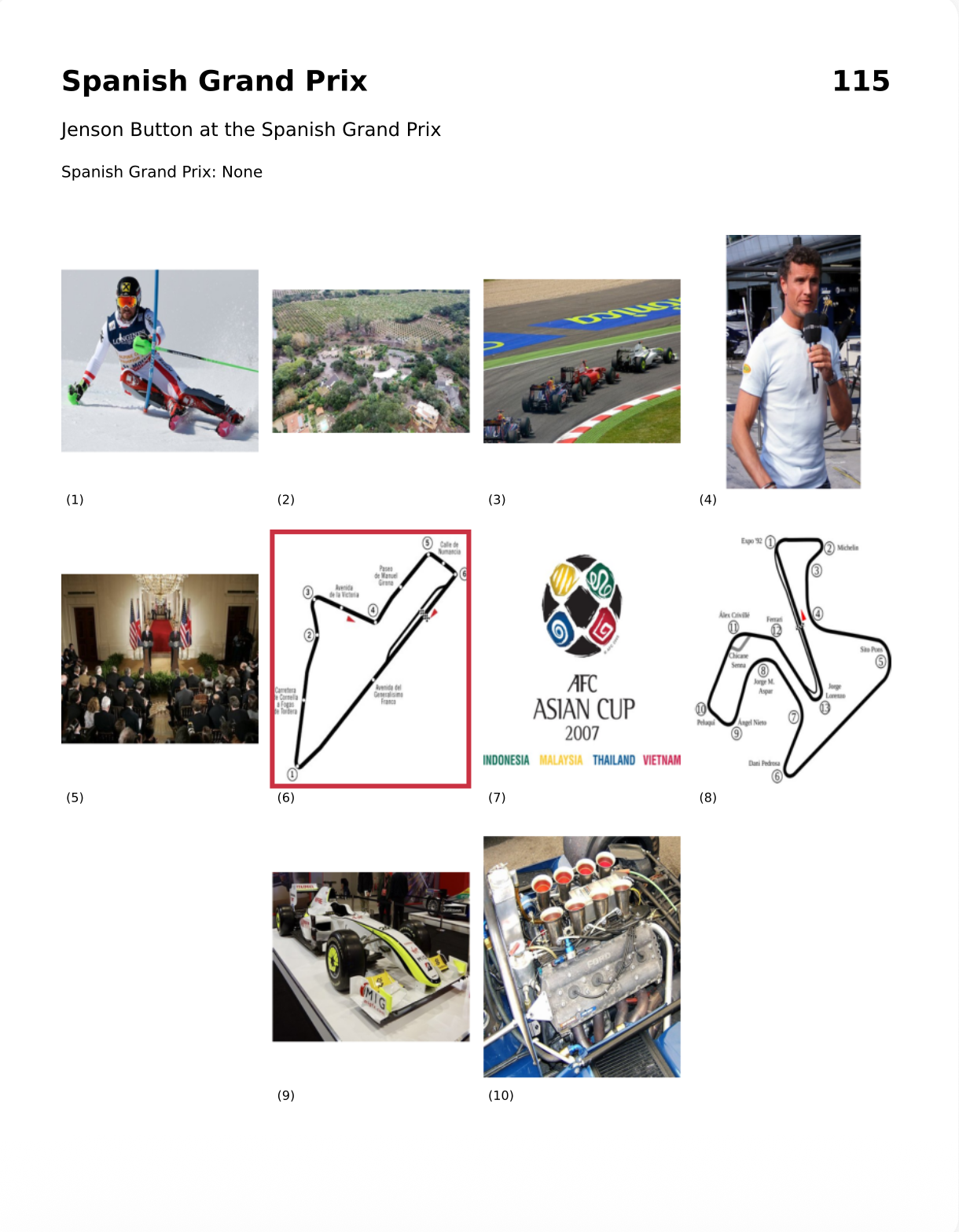}
    \caption{
Example of benchmark label ambiguity in WikiDiverse.
The target mention is ``Spanish Grand Prix'' in the context of Jenson Button at the Spanish Grand Prix.
Although the benchmark gold image corresponds to a circuit-map representation, another candidate depicts an actual race scene from the Spanish Grand Prix, making more than one candidate plausibly valid under the given context.
The red border indicates the SAGE gold label.
}
    \label{fig:wiki_err}
\end{figure*}

\begin{figure*}[!ht]
    \centering
    \includegraphics[width=.7\linewidth]{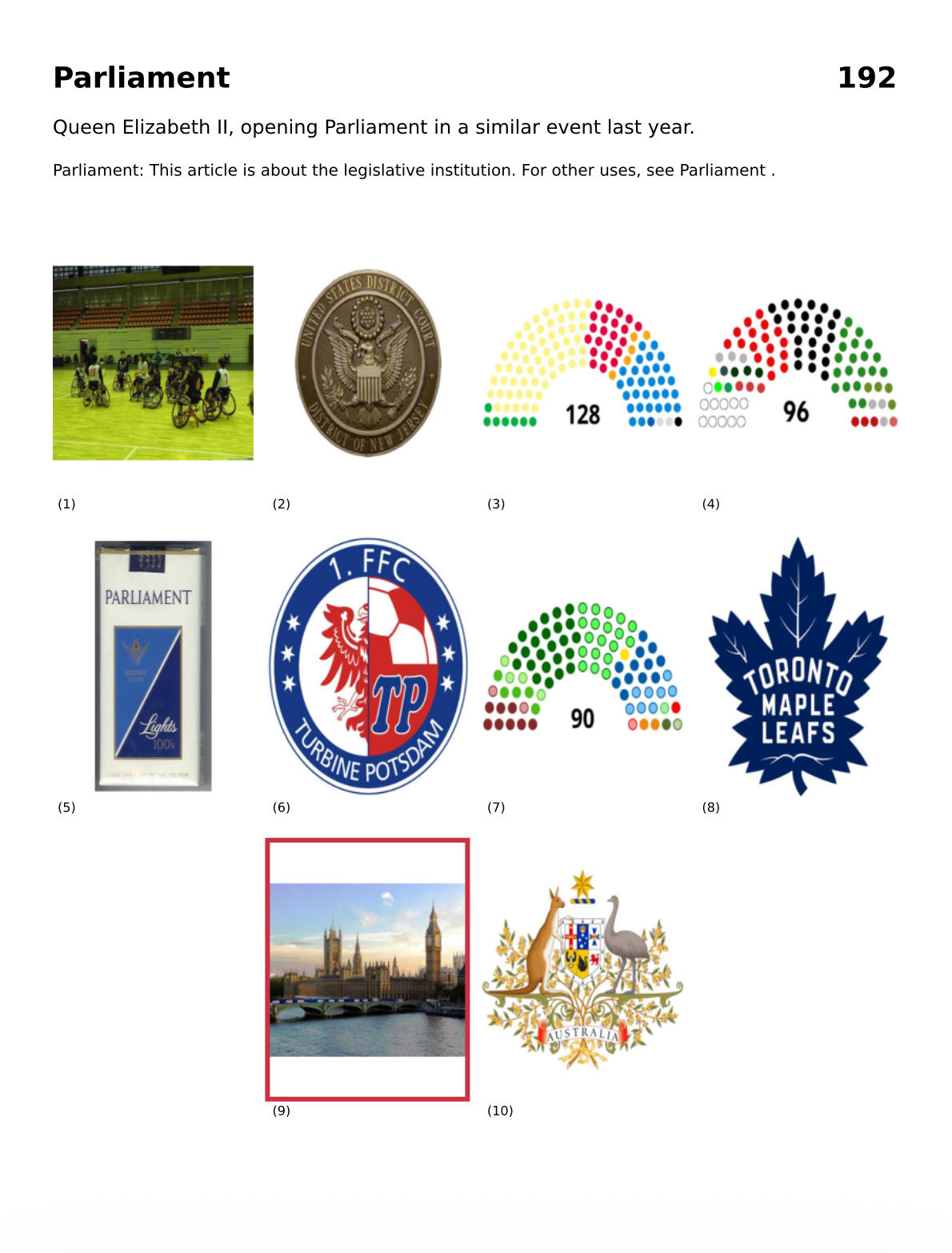}
    \caption{
Example of representation ambiguity in WikiDiverse.
The target mention ``Parliament'' can be visually grounded as a physical building, an institution, or an abstract parliamentary structure.
The benchmark gold image depicts the Palace of Westminster, while other candidates provide plausible institutional representations.
The red border indicates the SAGE gold label.
}
    \label{fig:wiki_amb}
\end{figure*}

Figures~\ref{fig:wiki_human_knowledge}--\ref{fig:wiki_amb} show representative examples from each category. These examples illustrate why the residual mismatches are expected in a named-entity grounding benchmark: unlike common object recognition, WikiDiverse often requires identifying a specific referent and deciding which visual representation of that referent is most appropriate.

\newpage

\newpage

\newpage

\newpage
\section{Hard-Negative Composition}
\label{apx:hard_negative_stats}

The composition of hard negatives differs between VerSe and WikiDiverse
because the two source datasets provide different structures for constructing
semantically confusable alternatives. In both subsets, each SAGE instance
contains one gold image and nine negatives; eligible hard negatives are
included first, and the remaining positions are filled with random negatives.

For \textbf{VerSe}, hard negatives are images associated with alternative
senses of the same target verb. Their availability is therefore limited by the
number of alternative senses and corresponding visual examples present in the
source data, resulting in relatively few hard negatives per query.

For \textbf{WikiDiverse}, we use entity candidates returned by the original
candidate-retrieval stage as the primary source of hard negatives, excluding
the gold entity and retaining candidates with an available representative
image. Because multiple retrieved entity candidates are available for many
mentions, WikiDiverse naturally yields a denser set of hard negatives.

\begin{table}[h]
\centering
\small
\begin{tabular}{lcc}
\hline
 & Avg. Hard & Avg. Random \\
\hline
WikiDiverse & 4.55 & 4.45 \\
VerSe       & 0.85 & 8.15 \\
\hline
\end{tabular}
\caption{Average numbers of hard and random negatives per query. Each
candidate set contains nine negatives in total; available hard negatives are
included first and remaining positions are filled with random negatives.}
\label{tab:hard_negative_stats}
\end{table}

These differences reflect the subset-specific candidate-generation mechanisms
rather than a fixed hard-negative proportion across grounding settings.

\newpage
\section{Details on the Lexical Knowledge Base}
\label{apx:lexical_kb}

\begin{table}[t]
\centering
\small
\begin{tabular}{l|ccc|ccc}
\hline
 & \multicolumn{3}{c|}{BabelNet} & \multicolumn{3}{c}{WordNet} \\
\cline{2-7}
 & VWSD & VerSe & WikiDiverse & VWSD & VerSe & WikiDiverse \\
\hline\hline
Coverage (\%) & 100.00 & 100.00 & 95.01 & 97.84 & 100.00 & 31.50 \\
Mean Ambiguity & 27.76 & 19.07 & 11.11 & 6.94 & 27.11 & 0.84 \\
\hline
\end{tabular}
\caption{Coverage and ambiguity statistics of lexical knowledge bases across datasets. Coverage is measured as the percentage of instances with at least one available sense definition. Ambiguity is measured as the average number of candidate senses per instance.}
\label{tab:coverage}
\end{table}

We provide additional analysis of the lexical knowledge bases used for definition augmentation, focusing on their coverage and ambiguity characteristics. In particular, we compare BabelNet and WordNet on the test splits across the VWSD, VerSe, and WikiDiverse datasets, as summarized in Table~\ref{tab:coverage}.

\paragraph{Background.}
BabelNet \citep{navigli2012babelnet} is a large-scale lexical knowledge base that integrates WordNet and Wikipedia, providing broad coverage over both common concepts and named entities. Prior work on visual word sense disambiguation has employed different lexical resources, including BabelNet \citep{yang2024polclip} and WordNet \citep{kwon2023vision}. In this work, we adopt BabelNet\footnote{Specifically, we utilize the BabelNet 5.3 online API} due to its broader coverage and richer sense inventories.

\paragraph{Coverage.}
As shown in Table~\ref{tab:coverage}, BabelNet achieves consistently high coverage across all datasets (100\% on VWSD and VerSe, and 95.01\% on WikiDiverse). In contrast, WordNet exhibits substantially lower coverage on WikiDiverse (31.5\%), while maintaining high coverage on VWSD (97.84\%) and VerSe (100\%). This suggests that coverage depends not only on the lexical resource itself, but also on how dataset instances align with the resource.

\paragraph{Ambiguity.}
Table~\ref{tab:coverage} further shows that BabelNet generally provides a richer set of candidate senses, with higher mean ambiguity on VWSD (27.76) and WikiDiverse (11.11), compared to WordNet (6.94 and 0.84, respectively). On VerSe, WordNet exhibits higher ambiguity (27.11 vs. 19.07), indicating that the distribution of candidate senses varies significantly across datasets and lexical resources.

\paragraph{Discussion.}
Overall, Table~\ref{tab:coverage} highlights that lexical resources differ in both coverage and the granularity of sense inventories. While higher coverage and richer ambiguity provide more opportunities for semantic grounding, they may make it more difficult to infer and synthesize the contextually intended meaning from a larger candidate inventory. 

\newpage
\section{Human Assessment of Synthesized Definition Plausibility}
\label{apx:definition_plausibility}

Given the risk of error propagation \citep{kwon2019effective} to downstream components, accurately estimating the quality of synthesized definitions is essential. We randomly sampled 150 instances, with 50 instances from each SAGE
subset. Two annotators independently assessed whether each synthesized
definition plausibly captured the meaning of the target term in its
textual context. They were shown the target, context, retrieved
candidate glosses, and synthesized definition, but not the candidate
images or benchmark gold labels. 

Annotators A and B judged 138/150 (92.0\%) and 139/150 (92.7\%)
definitions as contextually plausible, respectively. They agreed on
143/150 instances (95.3\%), with Gwet's AC1 of 0.946. Because the
judgments were strongly concentrated in the plausible category, we
report AC1, which is less affected by category-prevalence imbalance
than Cohen's $\kappa$ \citep{zec2017high}.

\begin{table}[t]
\centering
\footnotesize
\setlength{\tabcolsep}{3.5pt}
\begin{tabular}{lccc}
\toprule
Subset & Annotator A & Annotator B & Agreement / AC1 \\
\midrule
VWSD
& 48/50 (96.0\%)
& 47/50 (94.0\%)
& 47/50 (94.0\%) / 0.934 \\
VerSe
& 46/50 (92.0\%)
& 46/50 (92.0\%)
& 48/50 (96.0\%) / 0.953 \\
WikiDiverse
& 44/50 (88.0\%)
& 46/50 (92.0\%)
& 48/50 (96.0\%) / 0.951 \\
\midrule
Overall
& 138/150 (92.0\%)
& 139/150 (92.7\%)
& 143/150 (95.3\%) / 0.946 \\
\bottomrule
\end{tabular}
\caption{Human assessment of the contextual plausibility of synthesized
definitions.}
\label{tab:definition_plausibility}
\end{table}

\paragraph{Recovery beyond the retrieved inventory.}
Within the WikiDiverse subset, we identified three clear cases in which the contextually intended meaning was absent from the retrieved candidate glosses, yet both annotators judged the synthesized definition
plausible. Table~\ref{tab:missing_sense_recovery} summarizes these
cases. They show that synthesis is not restricted to copying one of
the retrieved glosses: the glosses provide reference information,
while the context and the model's parametric knowledge can recover a
meaning missing from the retrieved inventory.

\begin{table*}[t]
\centering
\scriptsize
\setlength{\tabcolsep}{3pt}
\renewcommand{\arraystretch}{1.08}
\begin{tabularx}{\textwidth}{
    @{}
    p{0.10\textwidth}
    p{0.20\textwidth}
    >{\raggedright\arraybackslash}X
    >{\raggedright\arraybackslash}X
    @{}
}
\toprule
Subset & Target and context & Retrieved-gloss gap &
Synthesized definition \\
\midrule
WikiDiverse
& \textit{jets}: ``One of the jets shadowing the bomber''
& Retrieved entries referred to the New York Jets, a musical gang, and
sports clubs, but not aircraft.
& Jets are fast, powerful aircraft. \\

\addlinespace
WikiDiverse
& \textit{McClellan}: McClellan shakes hands with President Bush ...
& Retrieved entries provided only a surname and an airport.
& McClellan refers to a person who served as the White House Press
Secretary. \\

\addlinespace
WikiDiverse
& \textit{AFC Asian Cup Group A match}: Thailand about to take a corner kick against Oman in their AFC Asian Cup Group A match.
& No candidate gloss was retrieved.
& A competition game between teams in the same group during the AFC
Asian Cup tournament. \\

\bottomrule
\end{tabularx}
\caption{Examples in which definition synthesis recovered a plausible
meaning absent from the retrieved candidate glosses.}
\label{tab:missing_sense_recovery}
\end{table*}

\paragraph{Jointly implausible outputs.}
The two annotators jointly judged eight outputs as contextually
implausible. As shown in
Table~\ref{tab:definition_implausible_examples}, the failures differed
across subsets. The VWSD error involved domain disambiguation, the
VerSe errors primarily involved overly narrow or incorrect action and
posture senses, and the WikiDiverse errors involved fine-grained entity
identification or insufficiently informative definitions.

\begin{table*}[t]
\centering
\scriptsize
\setlength{\tabcolsep}{3pt}
\renewcommand{\arraystretch}{1.08}
\begin{tabularx}{\textwidth}{
    @{}
    p{0.14\textwidth}
    >{\raggedright\arraybackslash}X
    >{\raggedright\arraybackslash}X
    p{0.14\textwidth}
    @{}
}
\toprule
Target & Context (abridged) & Synthesized definition & Issue \\
\midrule

\multicolumn{4}{@{}l}{\textbf{VWSD}} \\
\addlinespace[1pt]

\textit{expression}
& expression maths
& In computer science, an expression is a syntactic entity in a
programming language that may be evaluated to determine its value.
& Wrong domain \\

\midrule
\multicolumn{4}{@{}l}{\textbf{VerSe}} \\
\addlinespace[1pt]

\textit{perform}
& A man performing a trick on a skateboard.
& To do something in front of an audience, often to entertain.
& Wrong action sense \\

\textit{perform}
& A person with skis performing a jump.
& To do something in front of an audience, such as acting or music,
often in order to entertain.
& Wrong action sense \\

\textit{sit}
& Two people sitting on top of a truck.
& Be located or situated somewhere.
& Wrong posture sense \\

\midrule
\multicolumn{4}{@{}l}{\textbf{WikiDiverse}} \\
\addlinespace[1pt]

Smithsonian Museum of Natural History
& A woolly mammoth on display at the museum.
& Smithsonian Museum of Natural History: the intended sense of the word.
& Tautological \\

Kirchner
& Kirchner speaking to the Mexican Senate.
& Kirchner was a German expressionist painter and printmaker, and one
of the founders of the artists group Die Brücke.
& Wrong entity \\

Le TCD Foudre
& The Foudre at the naval arsenal in Toulon.
& A large wooden vat used for aging wine or spirits.
& Wrong lexical sense \\

De Winne
&  Crew of the Soyuz TMA-15 From left to right; Thirsk, Romanenko and De Winne.
& De Winne is a person who is part of a crew.
& Underspecified \\

\bottomrule
\end{tabularx}
\caption{Synthesized definitions jointly judged contextually implausible
by both annotators, grouped by SAGE subset.}
\label{tab:definition_implausible_examples}
\end{table*}

\newpage
\section{Model-Specific Ranking Approach}
\label{apx:baseline_ranking}

Different model families expose image--text compatibility through different
inference interfaces. We convert each model's native output into a scalar
compatibility score for each candidate image and use these scores to rank the
10 candidates within an instance.\footnote{All locally hosted experiments were
conducted on NVIDIA A100 GPUs.}

Across all methods, models receive the same multimodal input consisting of a
target word or mention, its textual context, and one candidate image. The
resulting candidate rankings are evaluated using the same Hits@1 and MRR
metrics. Because the scores are used only for within-instance ranking,
cross-example score calibration is not required.

\subsection{Embedding-Similarity Ranking for Image--Text Matching Models}

For image--text matching (ITM) models, we follow the standard pointwise
similarity-based ranking approach used for VWSD
\citep{raganato2023semeval}, consistent with the task definition
$(C,t,I)$ in Section~3.1.

\begin{tcolorbox}[
    colback=gray!8,
    colframe=blue!70!black,
    title=ITM Text Encoder Input,
    fonttitle=\bfseries
]

\textbf{Text input:}

\texttt{<context> <target\_term>}

\end{tcolorbox}

Given a textual context $C$ and a target term $t$, we construct the textual
query $q(C,t)$ by concatenating the context and target term using the model's
default text-encoder input format.

Let $f_T(\cdot)$ and $f_I(\cdot)$ denote the text and image encoders,
respectively. For each candidate image $i_k \in I$, we compute the pointwise
compatibility score

\[
s_k
=
f_T\bigl(q(C,t)\bigr)^{\top} f_I(i_k).
\]

We normalize the scores within the candidate set using

\[
\hat{s}_k
=
\frac{\exp(s_k)}
{\sum_{j=1}^{|I|}\exp(s_j)}.
\]

Candidates are ranked in descending order of $\hat{s}_k$. The resulting
ranking is used to compute Hits@1 and MRR.

\subsection{Token-Likelihood Ranking for Open-Source Generative VLMs}
\label{sec:logprob_ranking}

For open-source generative vision--language models, we use binary constrained
prompting and derive a scalar compatibility score from the relative token
likelihoods of affirmative and negative responses
\citep{abdallah-etal-2025-good}.

\begin{tcolorbox}[
    colback=gray!8,
    colframe=blue!70!black,
    title=Token-Likelihood Ranking Prompt for Open-Source VLMs,
    fonttitle=\bfseries
]

\textbf{System Message:}

You are a strict binary judge for visual lexical disambiguation.
Given a TARGET WORD, a short TEXT CONTEXT, and ONE candidate image,
decide whether the image correctly matches the meaning of the TARGET WORD
as used in the TEXT CONTEXT.

\textbf{Answer Rules:}
\begin{itemize}
    \item If the image clearly matches the intended sense, answer: YES
    \item If the image does NOT match or expresses a different sense, answer: NO
    \item Answer with a single token: YES or NO
    \item Do NOT output any other words or explanation
\end{itemize}

\textbf{User Message:}

\begin{itemize}
    \item \texttt{<candidate\_image>}
    \item \texttt{TARGET WORD: <target\_word>}
    \item \texttt{TEXT CONTEXT: <context>}
\end{itemize}

Question: Does this image correctly match the meaning of the TARGET WORD as
used in the TEXT CONTEXT?

Answer with a single token: YES or NO.

\end{tcolorbox}

Given an input $X_k$ consisting of the target word, textual context, and the
$k$-th candidate image, the model is prompted to generate a single-token
answer from the label space

\[
\mathcal{Y}
=
\{\texttt{YES},\texttt{NO}\}.
\]

We use deterministic inference with temperature $0$ and obtain the
full-vocabulary output logits at the first generated position. Let
$\ell(u\mid X_k)$ denote the logit assigned to token $u$. Because the final
score is normalized only over affirmative and negative response tokens,
using logits is equivalent to using full-vocabulary log probabilities up to
a shared normalization constant.

We define the affirmative and negative token sets as

\[
\mathcal{T}_{\texttt{YES}}
=
\{\texttt{YES},\texttt{Yes},\texttt{yes}\},
\]

and

\[
\mathcal{T}_{\texttt{NO}}
=
\{\texttt{NO},\texttt{No},\texttt{no}\}.
\]

The corresponding unnormalized probability masses are

\[
m_{\texttt{YES}}(X_k)
=
\sum_{u\in\mathcal{T}_{\texttt{YES}}}
\exp\bigl(\ell(u\mid X_k)\bigr),
\]

and

\[
m_{\texttt{NO}}(X_k)
=
\sum_{u\in\mathcal{T}_{\texttt{NO}}}
\exp\bigl(\ell(u\mid X_k)\bigr).
\]

We compute the normalized affirmative probability as

\[
P(\texttt{YES}\mid X_k)
=
\frac{
m_{\texttt{YES}}(X_k)
}{
m_{\texttt{YES}}(X_k)
+
m_{\texttt{NO}}(X_k)
}.
\]

The candidate ranking score is

\[
s_k
=
P(\texttt{YES}\mid X_k),
\]

and candidates are ranked in descending order of $s_k$.

This formulation provides a normalized compatibility signal between
affirmative and negative responses. Summing over capitalization variants
reduces sensitivity to surface-form and tokenization differences in the
response labels.

\begin{table}[t]
\centering
\small
\begin{tabular}{lccc}
\hline
Model & Params & Hits@1 & MRR \\
\hline\hline
InternVL-3.5-2B & 2B  & 55.1 & 69.7 \\
InternVL-3.5-8B & 8B  & 66.1 & 78.6 \\
Gemma3-4B       & 4B  & 57.9 & 72.6 \\
Gemma3-12B      & 12B & 75.6 & 84.7 \\
Qwen3-VL-2B     & 2B  & 62.2 & 74.6 \\
Qwen3-VL-8B     & 8B  & \textbf{76.5} & \textbf{85.4} \\
\hline
\end{tabular}
\caption{Open-source VLM performance under pointwise token-likelihood
ranking on the VWSD subset. Scores are reported on a 0--100 scale.}
\label{tab:open_source_vlms}
\end{table}

\begin{table}[t]
\centering
\small
\begin{tabular}{lccc}
\hline
Model & Params & Hits@1 & MRR \\
\hline\hline
Qwen2.5-VL-3B-Instruct  & 3B  & 63.7 & -- \\
Qwen2.5-VL-7B-Instruct  & 7B  & 71.3 & -- \\
Qwen2.5-VL-32B-Instruct & 32B & 76.5 & -- \\
Qwen2.5-VL-72B-Instruct & 72B & \textbf{79.1} & -- \\
Aya-Vision-8B            & 8B  & 20.5 & -- \\
Aya-Vision-32B           & 32B & 13.8 & -- \\
Gemma-3-4B               & 4B  & 29.8 & -- \\
Gemma-3-12B              & 12B & 29.8 & -- \\
Gemma-3-27B              & 27B & 74.7 & -- \\
Llama-4-Scout-17B        & 17B & 72.8 & -- \\
Llama-4-Maverick-17B     & 17B & 73.4 & -- \\
\hline
\end{tabular}
\caption{Published open-source VLM results from
\citet{musacchio2025assessing} on VWSD using joint multiple-choice
prediction. This method produces a single top-1 selection rather than a
complete candidate ranking; MRR is therefore not reported.}
\label{tab:prior_vwsd}
\end{table}

To contextualize our open-source baselines, we compare them with results
reported by \citet{musacchio2025assessing} on the same VWSD dataset under a
different zero-shot prediction method. Their method presents all 10
candidate images jointly and elicits a single multiple-choice prediction,
whereas our approach scores each candidate independently and produces a
complete ranking.

Under independent candidate ranking, Qwen3-VL-8B and Gemma3-12B achieve
76.5 and 75.6 Hits@1, respectively, broadly comparable to previously
reported open-source VLM results. Our approach additionally supports
ranking-based evaluation, yielding MRR scores of 85.4 and 84.7 for the two
models.

\subsection{Scalar Compatibility Ranking for Proprietary VLMs}

For proprietary VLMs, we elicit a generated scalar compatibility score for
each candidate image \citep{gu2026survey}. GPT-5.6 and 6 families did not expose the
token-level output log probabilities required by the likelihood-based approach
through the API configuration used in our experiments. We therefore use a
0--100 scalar compatibility interface and apply the same method uniformly
to all proprietary VLMs, providing a consistent comparison within the
proprietary-model family.

\begin{tcolorbox}[
    enhanced,
    colback=gray!10,
    colframe=blue!85!black,
    coltitle=white,
    fonttitle=\bfseries,
    title=Scalar Compatibility Ranking Prompt for Proprietary VLMs,
    boxrule=1pt
]

\textbf{System Message}

You are a strict judge for visual lexical disambiguation.
Given a TARGET WORD, a short TEXT CONTEXT, and ONE candidate image,
output a confidence score (0--100) indicating how well the image
matches the meaning of the TARGET WORD as used in the TEXT CONTEXT.

\textbf{Scoring Rules}

\begin{itemize}
    \item 100: perfect match, unmistakable
    \item 70--90: clear match with minor uncertainty
    \item 40--60: ambiguous or partial match
    \item 10--30: likely mismatch
    \item 0: clearly wrong sense
\end{itemize}

\textbf{Output Format}

\begin{itemize}
    \item Return ONLY a JSON object: \texttt{\{"confidence": 0\}}
    \item The value must be an integer between 0 and 100
    \item Do NOT output any other text or explanation
\end{itemize}

\textbf{User Message}

Evaluate this candidate image.

\texttt{<candidate\_image>}

\texttt{TARGET WORD: <target\_word>}

\texttt{TEXT CONTEXT: <context>}

Question: How confidently does this image match the meaning of the
TARGET WORD in the TEXT CONTEXT?

Return ONLY JSON: \texttt{\{"confidence": 0-100\}}

\end{tcolorbox}

Given input $X_k$, the model generates an integer compatibility score

\[
s_k
\in
\{0,1,\ldots,100\},
\]

representing how well the candidate image instantiates the contextualized
meaning of the target. We use $s_k$ directly to rank the 10 candidate images
within the instance. The resulting ranking is evaluated using the same
Hits@1 and MRR metrics as the rankings produced by the other model families.

\newpage
\subsection{Sensitivity to the Scoring Interface}
\label{apx:gpt_scoring_sensitivity}

In the main experiments, GPT-4.1-mini and GPT-5.6 and 6 use the same
pointwise 0--100 compatibility interface because token-level output log
probabilities are unavailable for GPT-5.6 and GPT-6 under our API configuration.
In contrast, Gemma3-12B uses binary YES/NO token-likelihood scoring. To
assess whether these interface choices bias the comparison, we evaluate
GPT-4.1-mini under both interfaces and additionally evaluate Gemma3-12B
using scalar compatibility scores. For GPT-4.1-mini, the API returns at
most 20 candidate-token log probabilities.

Table~\ref{tab:gpt_scoring_sensitivity} shows that YES/NO token-likelihood
scoring matches or exceeds scalar scoring for both models across all three
subsets. In particular, scalar scoring substantially lowers Gemma3-12B
performance, whereas token-likelihood scoring modestly improves
GPT-4.1-mini. Thus, scalar scoring does not selectively favor
GPT-4.1-mini. Conversely, enforcing a uniform scalar interface across
models may underestimate open-source VLM performance, as illustrated by the
substantial decrease for Gemma3-12B. We therefore retain model-native
scoring interfaces in the main experiments and report this sensitivity
analysis to make cross-family comparisons transparent. Although individual
candidate rankings vary across interfaces, the aggregate conclusions remain
consistent.

\begin{table}[t]
\centering
\scriptsize
\setlength{\tabcolsep}{2.5pt}
\begin{tabular}{llrrrr}
\toprule
Model & Dataset
& \multicolumn{2}{c}{Scalar score}
& \multicolumn{2}{c}{YES/NO logprob} \\
\cmidrule(lr){3-4}\cmidrule(lr){5-6}
& & Hits@1 & MRR & Hits@1 & MRR \\
\midrule
\multirow{3}{*}{GPT-4.1-mini}
& VWSD        & 78.6 & 86.9 & 80.8 & 88.1 \\
& VerSe       & 58.2 & 71.1 & 62.1 & 74.4 \\
& WikiDiverse & 66.6 & 79.8 & 67.9 & 80.0 \\
\midrule
\multirow{3}{*}{Gemma3-12B}
& VWSD        & 58.8 & 74.5 & 75.6 & 84.7 \\
& VerSe       & 47.0 & 62.3 & 59.3 & 72.7 \\
& WikiDiverse & 47.9 & 67.1 & 62.8 & 76.9 \\
\bottomrule
\end{tabular}
\caption{Sensitivity to scoring interfaces. GPT-4.1-mini and Gemma3-12B
are evaluated using both scalar compatibility scores and binary YES/NO
token likelihoods. Token-likelihood scoring matches or exceeds scalar
scoring for both models.}
\label{tab:gpt_scoring_sensitivity}
\end{table}
\newpage
\section{Details on GAR templates}
\label{apx:GAR_details}

\subsection{Context-Guided Definition Synthesis}
\label{sec:definition-synthesis-prompt}

We first synthesize a concise definition of the target word's
contextually intended sense from the provided candidate dictionary
definitions. The model does not select a definition index; instead, it
produces a single dictionary-style definition.

\begin{tcolorbox}[
    enhanced,
    colback=gray!10,
    colframe=blue!85!black,
    coltitle=white,
    fonttitle=\bfseries,
    title=Context-Guided Definition Synthesis Prompt,
    boxrule=1pt
]

\textbf{System Message}

You output ONLY valid JSON. No extra text.

\textbf{Output Format}

Return the following schema exactly:

\texttt{\{"definition": "<definition>"\}}

\textbf{Task}

Use the TEXT CONTEXT to identify the intended sense of the KEYWORD.
Then write a concise, general, dictionary-style definition of that
sense.

\textbf{Instructions}

\begin{itemize}
    \item The definition should describe the core lexical meaning of
    the KEYWORD.
    \item Prefer wording close to the candidate definitions when
    possible.
    \item Do NOT quote or mention the candidate definitions.
    \item Do NOT output an index.
    \item The definition must contain at most one sentence.
    \item The definition must be non-empty and match the intended
    sense indicated by the TEXT CONTEXT.
    \item Keep the definition concise and dictionary-like.
    \item Do not use bullet points, markdown, or headings.
    \item Avoid unrelated information not supported by the candidate
    definitions.
\end{itemize}

\textbf{User Message}

\texttt{KEYWORD: <target\_word>}

\texttt{CANDIDATE DEFINITIONS (reference only):}

\texttt{- <candidate\_definition\_1>}

\texttt{- <candidate\_definition\_2>}

\texttt{- \ldots}

\texttt{TEXT CONTEXT:}

\texttt{<context>}

Return ONLY JSON:

\texttt{\{"definition": "<definition>"\}}

\end{tcolorbox}

\newpage
\subsection{Definition Expansion}
\label{sec:definition-expansion-prompt}

We expand the synthesized sense definition with additional clarifying
details while constraining the model to preserve the original sense.

\begin{tcolorbox}[
    enhanced,
    colback=gray!10,
    colframe=blue!85!black,
    coltitle=white,
    fonttitle=\bfseries,
    title=Definition Expansion Prompt,
    boxrule=1pt
]

\textbf{System Message}

You output ONLY valid JSON. No extra text.

\textbf{Output Format}

Return the following schema exactly:

\texttt{\{"expanded\_definition": "<string>"\}}

\textbf{Task}

Expand the DEFINITION of the TARGET WORD.

\textbf{Instructions}

\begin{itemize}
    \item The \texttt{expanded\_definition} MUST keep the SAME SENSE
    as the provided definition.
    \item The \texttt{expanded\_definition} MUST be entailed by the
    provided definition; do not introduce sense drift.
    \item Use simple words and add clarifying details, especially
    visual or physical traits when applicable.
    \item Expand the provided meaning in terms of what the correct image is expected to depict.
    \item When describing what an image would show, keep the target
    meaning itself as the visual anchor.
    \item Do not replace the target meaning with a likely surrounding
    object, ingredient, symbolic namesake, or example scene.
    \item Do not over-emphasize source, origin, etymology, production
    process, symbolic namesake, or background facts when they are not
    the visual target.
    \item Preserve the context-specific phrase when it identifies the
    visual referent.
    \item Use at most four sentences.
    \item Do NOT repeat sentences or restate the same information.
    \item The output must be non-empty.
\end{itemize}

\textbf{User Message}

\texttt{TARGET WORD: <target\_word>}

\texttt{DEFINITION: <synthesized\_definition>}

Return ONLY JSON:

\texttt{\{"expanded\_definition": "<expanded\_definition>"\}}

\end{tcolorbox}

\newpage
\subsection{Semantic Decomposition}
\label{sec:semantic-decomposition-prompt}

We decompose the expanded definition into three complementary
visual-semantic axes: object, attribute, and state/event.
\begin{tcolorbox}[
    enhanced jigsaw,
    breakable,
    colback=gray!10,
    colframe=blue!85!black,
    coltitle=white,
    fonttitle=\bfseries,
    title=Semantic Decomposition Prompt,
    boxrule=1pt
]

\textbf{System Message}

You are extracting structured semantic information for multimodal
lexical grounding.

\textbf{User Message}

\texttt{Target word: <target\_word>}

\texttt{Expanded definition:}

\texttt{<expanded\_definition>}

Decompose the target meaning into three visual-semantic axes:
Object, Attribute, and State/Event.

\textbf{Output Format}

Return JSON only in this format:

\begin{verbatim}
{
  "object": "...",
  "attribute": "...",
  "state_event": "..."
}
\end{verbatim}

\textbf{Goal}

Describe how the target meaning would be grounded in an image.
Preserve the intended sense as specifically as possible, while using
only information that could plausibly be checked visually.

\textbf{Axis Meanings}

\textbf{Object}

\begin{itemize}
    \item Identify the primary visual anchor of the intended meaning.
    \item The object should name the target-specific referent, not
    merely a surrounding scene, source ingredient, symbolic namesake,
    or generic category.
    \item Preserve specific names, entities, compounds, subtypes,
    model names, titles, and context-specific senses whenever they
    are visually meaningful.
    \item If the target is a named entity, keep the entity name in
    the object and add a visible type or realization.
    \item If the target is a common noun or compound, keep the
    intended subtype or sense, not only the broad headword.
    \item If the image may show surrounding objects, ingredients,
    examples, or scenes, mention them only as supporting cues, not
    as the main object.
\end{itemize}

\textbf{Attribute}

\begin{itemize}
    \item Describe visible properties or relations that help
    recognize the target meaning.
    \item Include shape, color, material, parts, markings, layout,
    text, logo, clothing, posture, spatial relation, or other visible
    cues when useful.
    \item For actions, prefer relational cues between participants,
    tools, and objects.
    \item For named entities, prefer distinctive visible signatures
    rather than broad category-level descriptions.
    \item Avoid non-visual facts such as origin, history, role,
    purpose, ownership, nationality, smell, taste, or hidden
    properties unless they correspond to visible evidence.
\end{itemize}

\textbf{State/Event}

\begin{itemize}
    \item Describe a visible action, interaction, posture,
    configuration, use, motion, operation, or process.
    \item For action-like meanings, capture the main visible event.
    \item For object or entity meanings, include this only when a
    clearly visible action or configuration is part of the meaning.
    \item Avoid hidden causes, mental states, biological mechanisms,
    social intentions, institutional roles, or other facts that
    cannot be judged from a single image.
    \item Describe what can be seen happening in a still image;
    otherwise, use \texttt{not clearly specified}.
\end{itemize}

\textbf{Guidelines}

\begin{itemize}
    \item Use only the target word and the given definition.
    \item Prefer specific and discriminative visual cues over generic
    ones.
    \item Do not invent details that are not supported.
    \item If an axis is not visually checkable, write
    \texttt{not clearly specified}.
    \item Keep each value concise but informative.
    \item Use semicolon-separated phrases when helpful.
    \item When the definition mentions both a visible final target
    and its source, origin, production process, or symbolic name,
    choose the final visually grounded target as the object.
    \item Output valid JSON only.
\end{itemize}

\end{tcolorbox}

\newpage
\subsection{Verification Question Generation}
\label{sec:verification-question-generation-prompt}

We convert each decomposed semantic component into a concise yes/no
question that can be evaluated using directly visible evidence in a
single image.

\begin{tcolorbox}[      enhanced jigsaw,      breakable,      colback=gray!10,      colframe=blue!85!black,      coltitle=white,      fonttitle=\bfseries,      title=Verification Question Generation,      boxrule=1pt,      arc=1mm,      left=2mm,      right=2mm,      top=1.5mm,      bottom=1.5mm,      before skip=6pt,      after skip=6pt  ]

\textbf{System Message}

You are generating visual verification questions for image-based
lexical disambiguation.

\textbf{User Message}

\texttt{TARGET WORD: <target\_word>}

\texttt{DEFINITION: <expanded\_definition>}

\texttt{SEMANTIC ATTRIBUTES:}

\texttt{\{}

\texttt{\ \ "object": "<object>",}

\texttt{\ \ "attribute": "<attribute>",}

\texttt{\ \ "state\_event": "<state\_event>"}

\texttt{\}}

Convert each semantic attribute into one short yes/no question that
can be answered from directly visible evidence in a single image.

\textbf{Output Format}

Return JSON only in the following format:

\texttt{\{}

\texttt{\ \ "object": "...",}

\texttt{\ \ "attribute": "...",}

\texttt{\ \ "state\_event": "..."}

\texttt{\}}

\textbf{Object}

\begin{itemize}
    \item Use the \texttt{object} field.
    \item Ask whether the most specific target referent is visible or
    present.
    \item Preserve specific names, entities, compounds, subtypes, and
    context-specific expressions when useful.
    \item Do not reduce the target to a generic category.
    \item Ask only about the referent itself, not its attributes,
    materials, actions, functions, or states.
\end{itemize}

\textbf{Attribute}

\begin{itemize}
    \item Use the \texttt{attribute} field.
    \item Ask about visible properties, parts, shapes, colors,
    materials, markings, layouts, text, logos, clothing, posture, or
    spatial relations.
    \item For relational attributes, ask directly about the visible
    relation.
    \item Exclude attributes that cannot be checked visually.
\end{itemize}

\textbf{State/Event}

\begin{itemize}
    \item Use the \texttt{state\_event} field.
    \item Ask about visible actions, interactions, motion, posture,
    use, operation, contact, configuration, or physical processes.
    \item If the state or event is mostly non-visual, retain only its
    visible component.
\end{itemize}

\textbf{Guidelines}

\begin{itemize}
    \item Do not reinterpret the target meaning from scratch.
    \item Faithfully convert the provided semantic attributes into
    verification questions.
    \item Ask exactly one yes/no question for each visually checkable
    field.
    \item Every non-empty output must be a complete yes/no question.
    \item Keep each question short, natural, specific, and
    unambiguous.
    \item Do not use vague terms such as ``suggest,'' ``imply,''
    ``possibly,'' or ``seems.''
    \item If an axis is not visually checkable, write
    \texttt{not clearly specified}.
    \item Output valid JSON only.
\end{itemize}

\end{tcolorbox}

\newpage
\subsection{Checklist Filling}
\label{sec:candidate-image-verification-prompt}

For each candidate image, we independently evaluate the verification
questions associated with the object, attribute, and state/event
axes. The probability assigned to \texttt{YES} is used as the
axis-specific compatibility score.

\begin{tcolorbox}[      enhanced jigsaw,      breakable,      colback=gray!10,      colframe=blue!85!black,      coltitle=white,      fonttitle=\bfseries,      title=Checklist Filling,      boxrule=1pt,      arc=1mm,      left=2mm,      right=2mm,      top=1.5mm,      bottom=1.5mm,      before skip=6pt,      after skip=6pt  ]

\textbf{System Message}

You are a strict binary judge for visual lexical disambiguation.

Given a TARGET WORD, its definition, a short TEXT CONTEXT, a semantic
question, and ONE candidate image, determine whether the image
satisfies the semantic question.

\textbf{Decision Rules}

\begin{itemize}
    \item Answer YES if the image clearly supports the semantic
    question.
    \item Answer NO otherwise.
    \item If the evidence is ambiguous or uncertain, answer NO.
\end{itemize}

\textbf{Output Format}

\begin{itemize}
    \item Output exactly one token: \texttt{YES} or \texttt{NO}.
    \item Do NOT provide an explanation or any additional text.
\end{itemize}

\textbf{User Message}

Evaluate this candidate image.

\texttt{<candidate\_image>}

\texttt{TARGET WORD: <target\_word>}

\texttt{TARGET WORD DEFINITION: <expanded\_definition>}

\texttt{TEXT CONTEXT: <context>}

\texttt{SEMANTIC AXIS: <object|attribute|state\_event>}

\texttt{SEMANTIC QUESTION: <semantic\_question>}

Does the candidate image satisfy the semantic question?

Answer with YES or NO.

\end{tcolorbox}

\newpage
\section{Experimental Settings on Comparison Models}
\label{apx:comparison_models}
\subsection{Unsupervised VWSD}
\paragraph{CLIP}
We report the CLIP ViT-L/14 performance on the original VWSD dataset as introduced by \citet{raganato2023semeval}. CLIP ranks candidate images using cosine similarity between textual and visual embeddings.

\paragraph{UVWSD}
\citet{kwon2023vision} proposes a Bayesian-style inference framework that combines CLIP embeddings with lexical knowledge from WordNet to perform sense disambiguation. 
As the original work reports results primarily on the training split, we additionally evaluate UVWSD on the official test split under our unified setting. 
For consistency with our experimental setting, we use BabelNet as the external LKB.

\paragraph{GPT\_BLIP}
\citet{kritharoula2023large} introduces a zero-shot VWSD framework that leverages large language models to enrich gloss representations through various prompting strategies (e.g., exact gloss, “what is,” descriptive reformulation). 
The enriched textual representations are then used in a multimodal retrieval framework with CLIP-style encoders to rank candidate images. 
The method remains fully unsupervised and does not involve task-specific fine-tuning.

\paragraph{Q-VWSD}
Q-VWSD models multiple candidate glosses as a quantum superposition state and applies quantum-inspired interference mechanisms during inference to mitigate gloss bias \citep{qiao2025quantum}. 
As no official implementation is publicly available, we re-implement the generalized classical variant, referred to as Quantum-Inspired Inference (Q-VWSDQI). 
Consistent with other experiments, we use BabelNet as the lexical knowledge source.

\paragraph{Qwen2.5-VL}
\citet{musacchio2025assessing} systematically evaluate encoder-based and
generative vision--language models on the original SemEval-2023 VWSD benchmark,
along with a multilingual benchmark constructed using co-hyponym candidate
images. For generative VLMs, the target expression and all 10 candidate images
are presented jointly, and the model is prompted to select the image that best
matches the intended meaning in a zero-shot multiple-choice setting. We report
their strongest result on the original English VWSD benchmark,
Qwen2.5-VL-72B-Instruct, which achieves 79.05 Hits@1 without task-specific
fine-tuning. Because the approach produces a single top-1 prediction rather
than a complete candidate ranking, MRR is not available.

\paragraph{DCTP}
DCTP proposes a training-free CLIP-based VWSD framework that ensembles semantic and photorealistic prompts in a dual-channel architecture \citep{bhattacharya2026visual}. 
It further applies test-time image augmentation to enhance text–image alignment during inference.

\subsection{Supervised VWSD}

\paragraph{CLIP (Fine-tuned)}
We report the fine-tuned CLIP ViT-L/14 results as implemented in \citet{yang2024polclip}. 
The model is trained on VWSD training data under a supervised contrastive objective and evaluated on the official test split.

\paragraph{FCLL}
FCLL \citep{yang2023tam} proposes a fine-grained contrastive language–image learning framework for VWSD. 
It learns sense-aware textual representations aligned with visual features through supervised contrastive training. 
The model aims to improve discrimination among visually similar candidate images by explicitly modeling sense-level distinctions.

\paragraph{ARPA}
ARPA \citep{papastavrou2024arpa} introduces a hybrid framework that incorporates large language models for semantic enrichment of word senses together with transformer-based vision–language encoders. 
The enriched textual representations are integrated with visual features to improve multimodal alignment under supervised training.

\paragraph{PolCLIP}
PolCLIP \citep{yang2024polclip} presents a supervised extension of CLIP tailored for VWSD. 
It learns complementary multimodal representations from textual and visual inputs and optimizes a contrastive objective to enhance sense-level discrimination among candidate images.
\newpage

\section{Reciprocal Rank Fusion Details}
\label{apx:rrf}

GAR-full combines the rankings produced by Def-only and CGR using
reciprocal rank fusion (RRF). For each candidate image $i$, we compute
\[
R_{\mathrm{RRF}}(i)
=
\frac{1}{k+r_{\mathrm{Def}}(i)}
+
\frac{1}{k+r_{\mathrm{CGR}}(i)},
\]
where $r_{\mathrm{Def}}(i)$ and $r_{\mathrm{CGR}}(i)$ are the
one-based ranks assigned by Def-only and CGR, respectively. We set
$k=60$ and apply fusion over all 10 candidate images.

\section{Target Verb Neutralization}
\label{apx:verb_neutralization}
To evaluate the extent to which performance depends on the lexical identity of the target verb, we construct a neutralized-caption variant of VerSe. In this setting, the target verb is first masked using a lightweight lemmatization-based procedure, and the masked span is then replaced with a grammatically plausible but lexically uninformative expression. Given a target verb lemma and caption, we parse the caption with spaCy\footnote{https://spacy.io} and mask the verbal span whose lemma matches the target, including adjacent auxiliaries when applicable (e.g., \textit{is moving}, \textit{has been moved}). The masked caption is then filled by Qwen2.5-3B-Instruct~\citep{qwen2025qwen25technicalreport} using a restricted set of generic placeholders such as \textit{do something}, \textit{does something}, \textit{doing something}, or \textit{has something done}. We apply simple validation checks to prevent reintroducing the target verb or its inflected forms, and fall back to a generic replacement when validation fails. This variant is intended as a diagnostic ablation rather than a fluency-preserving paraphrase procedure. 

\begin{algorithm}[!h]
\caption{Lemma-based target verb neutralization}
\label{alg:neutralization}
\begin{algorithmic}[1]
\Require target verb lemma $v$, caption $c$
\State parse $c$ using POS tagging and lemmatization
\State $m \gets \emptyset$
\For{each token $t_i$ in $c$}
    \If{$\mathrm{lemma}(t_i) = v$ or $\mathrm{text}(t_i) = v$}
        \State $m \gets t_i$
        \State expand $m$ leftward to include adjacent auxiliaries, if any
        \State $c_{\mathrm{mask}} \gets$ replace $m$ in $c$ with \texttt{[MASK]}
        \State $\hat{c} \gets$ fill \texttt{[MASK]} with a generic placeholder using an instruction-tuned LM
        \If{$\hat{c}$ contains an allowed neutral placeholder and does not contain $v$ or its inflected forms}
            \State \Return $\hat{c}$
        \Else
            \State \Return fallback generic replacement for \texttt{[MASK]}
        \EndIf
    \EndIf
\EndFor
\State \Return $c$
\end{algorithmic}
\end{algorithm}

\paragraph{Examples.}
Example outputs from the neutralization procedure are shown below. We show both the intermediate masked caption and the final neutralized caption.

\begin{itemize}
    \item \textbf{Original:} ``A person pointing to a section of their food.'' \\
    \textbf{Masked:} ``A person [MASK] to a section of their food.'' \\
    \textbf{Neutralized:} ``A person does something to a section of their food.''

    \item \textbf{Original:} ``A man in a blue cap is pointing on a street corner.'' \\
    \textbf{Masked:} ``A man in a blue cap [MASK] on a street corner.'' \\
    \textbf{Neutralized:} ``A man in a blue cap does something on a street corner.''

    \item \textbf{Original:} ``A man is pointing toward the court as a boy holds the ball.'' \\
    \textbf{Masked:} ``A man [MASK] toward the court as a boy holds the ball.'' \\
    \textbf{Neutralized:} ``A man does something toward the court as a boy holds the ball.''

    \item \textbf{Original:} ``A table that is set with plates and cups.'' \\
    \textbf{Masked:} ``A table that [MASK] with plates and cups.'' \\
    \textbf{Neutralized:} ``A table that has something done with plates and cups.''

    \item \textbf{Original:} ``A young boy on a skateboard is being helped by a man.'' \\
    \textbf{Masked:} ``A young boy on a skateboard [MASK] by a man.'' \\
    \textbf{Neutralized:} ``A young boy on a skateboard has something done by a man.''
\end{itemize}

\newpage

\section{Detailed on Qualitative Analysis on Error Cases}
\label{apx:cgr_errors}

We organize the observed failures of constraint-guided semantic scoring (CGR)
into five recurring categories.  Each subsection presents two representative
examples.  The reported score is the arithmetic mean over valid Object,
Attribute, and State questions; invalid or unspecified axes are omitted.

\subsection{Upstream Semantic or Referent Errors}
\label{app:cgr-upstream}

These failures occur before visual verification: the synthesized definition
selects the wrong sense or referent, so the resulting questions consistently
evaluate an unintended concept.  In VWSD \#129, the context \emph{wild torch}
refers to a plant, but the definition substitutes the portable-light sense of
\emph{torch}.  Consequently, the Object question assigns the flashlight
distractor $P(\mathrm{YES})=0.9997$, compared with $0.0000$ for the gold image.

\begin{figure*}[t]
  \centering
  \includegraphics[width=0.96\textwidth]{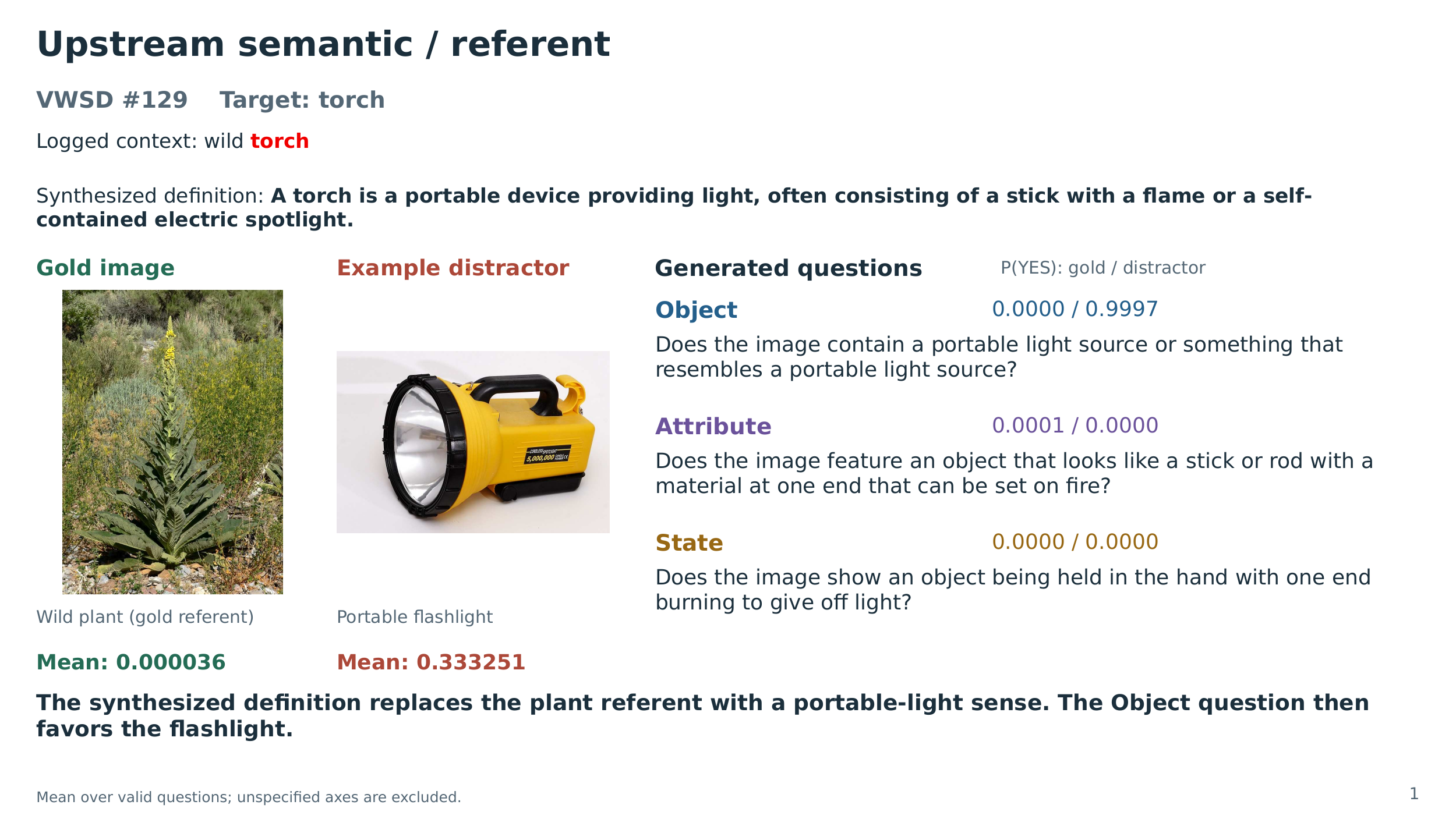}
  \caption{Upstream semantic error for VWSD \#129 (\emph{torch}).  The
  synthesized definition changes the intended plant referent into a portable
  light source, causing the generated Object constraint to favor the
  flashlight distractor.}
  \label{fig:cgr-upstream-torch}
\end{figure*}

WikiDiverse \#604 shows the same mechanism for a named entity.  The context
refers to Garuda Indonesia returning to European airspace, whereas the
definition resolves \emph{Garuda} as the Hindu deity.  Both valid constraints
therefore favor the mythological depiction, producing means of $0.001587$ for
the airline image and $0.999702$ for the distractor.

\begin{figure*}[t]
  \centering
  \includegraphics[width=0.96\textwidth]{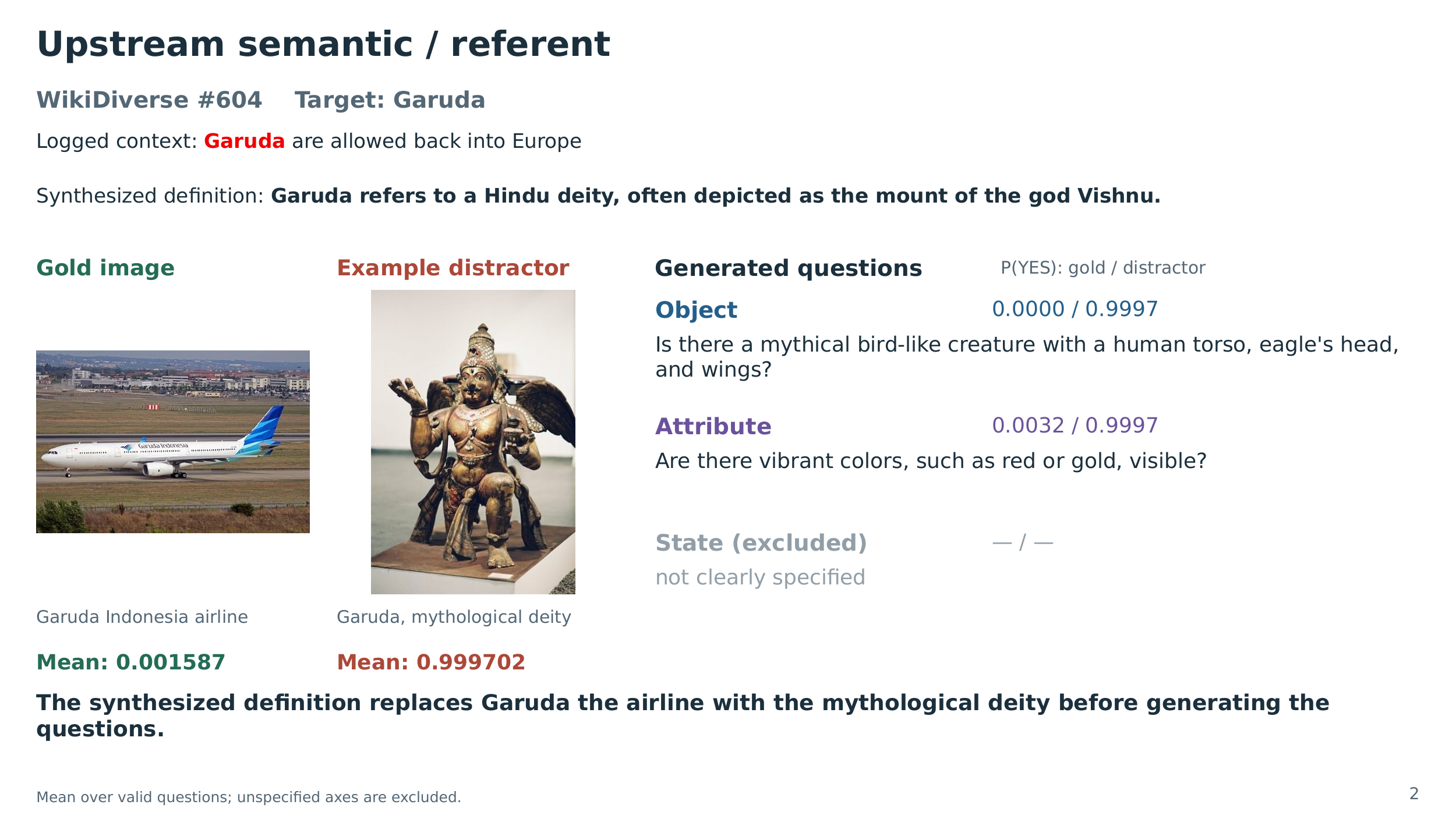}
  \caption{Upstream referent error for WikiDiverse \#604 (\emph{Garuda}).
  The airline referent is replaced by the mythological deity before question
  generation.}
  \label{fig:cgr-upstream-garuda}
\end{figure*}

\clearpage

\subsection{Decomposition Drift}
\label{app:cgr-decomposition}

Decomposition drift occurs when the synthesized definition is adequate or
partly adequate, but the generated questions shift toward another lexical
sense or an overly literal visual prototype.  For VWSD \#30, the definition
correctly describes facelift surgery, yet all three questions ask about
physically lifting an object.  The helicopter image consequently receives a
mean of $0.990417$, compared with $0.000956$ for the operating-room image.

\begin{figure*}[t]
  \centering
  \includegraphics[width=0.96\textwidth]{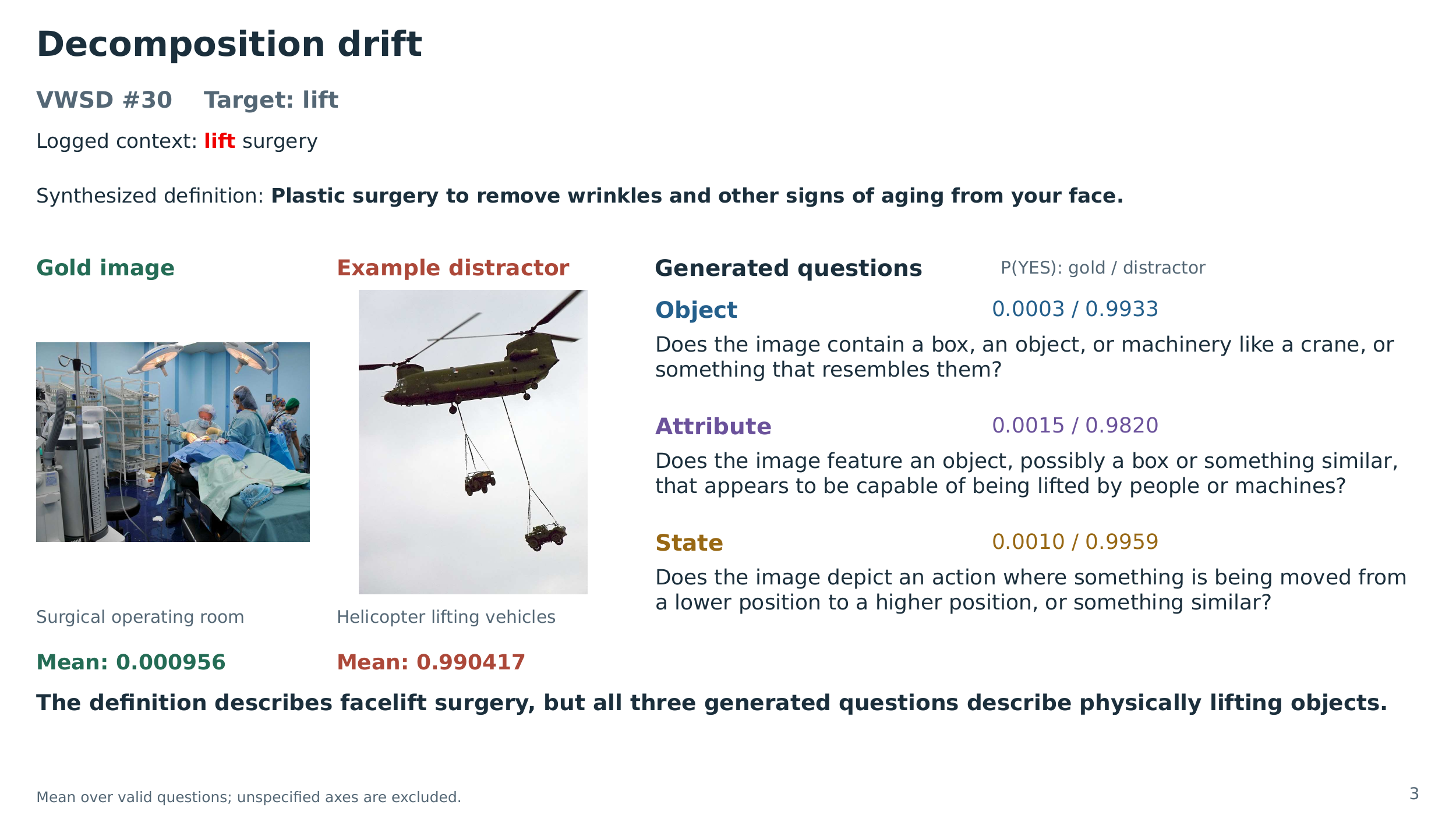}
  \caption{Decomposition drift for VWSD \#30 (\emph{lift}).  Questions derived
  from a facelift definition drift to the physical lifting sense.}
  \label{fig:cgr-decomposition-lift}
\end{figure*}

For VWSD \#211, the definition correctly identifies Milk River, and the Object
question gives nearly equal support to the river and milk bottle ($0.9820$ vs.
$0.9770$).  The Attribute question then introduces the dairy-specific cue
``white or creamy,'' assigning $0.7311$ to the bottle and $0.0000$ to the
river.  This drift reverses the valid-axis mean ($0.491010$ vs. $0.854041$).

\begin{figure*}[t]
  \centering
  \includegraphics[width=0.96\textwidth]{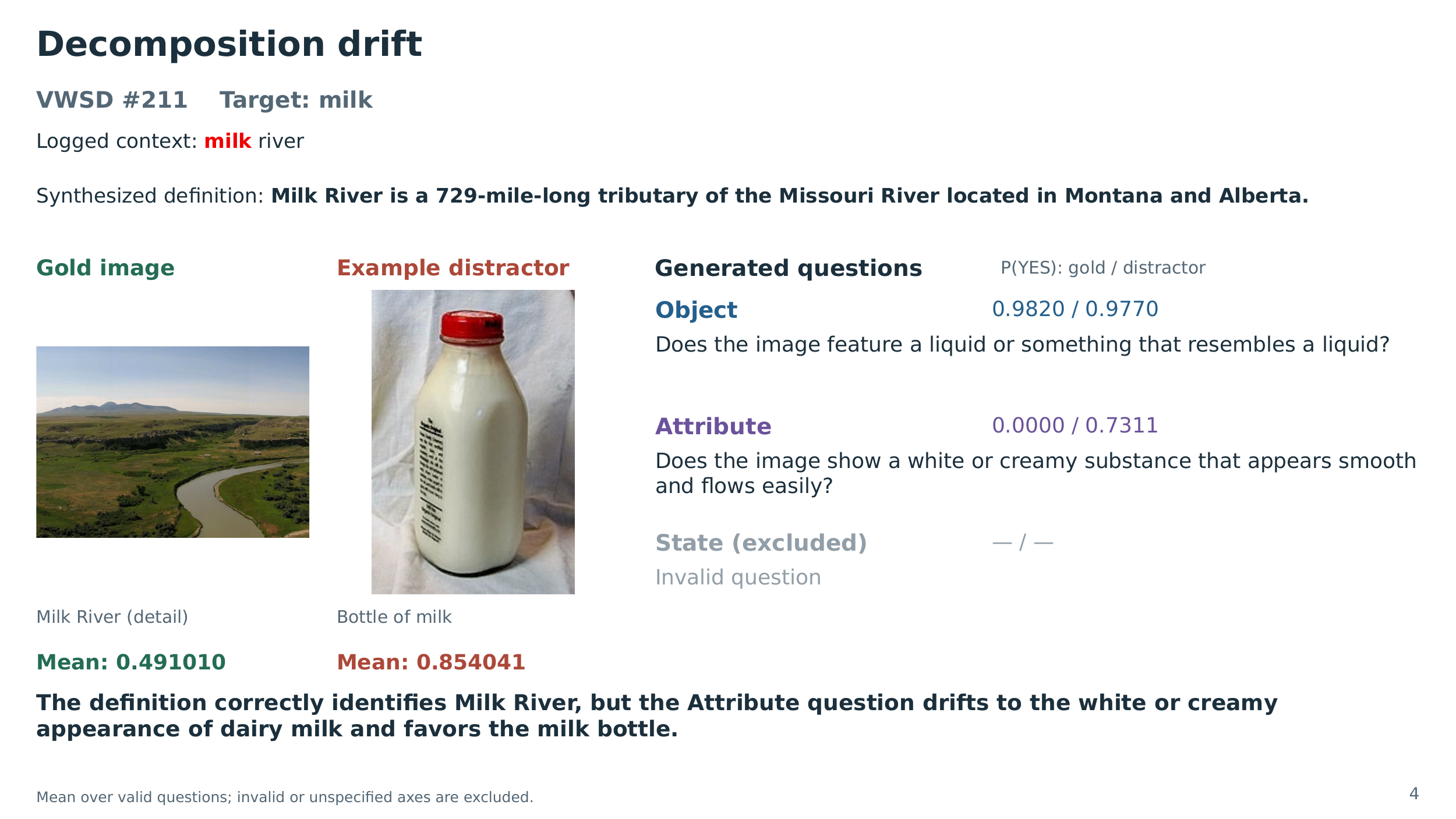}
  \caption{Decomposition drift for VWSD \#211 (\emph{milk}).  A
  dairy-specific Attribute constraint overwhelms the correctly identified
  river referent.}
  \label{fig:cgr-decomposition-milk}
\end{figure*}

\clearpage

\subsection{Cross-Axis Conflict}
\label{app:cgr-cross-axis}

Cross-axis conflicts arise when individually plausible constraints support
different candidates, and averaging them changes the final ranking.  For
WikiDiverse \#598, the Object question strongly favors the Phnom Penh Airport
logo ($0.9820$ vs. $0.1480$), while the Attribute question about parked
airplanes strongly favors an aerial photograph of another airport ($0.0032$
vs. $0.9770$).  The latter reverses the overall mean.

\begin{figure*}[t]
  \centering
  \includegraphics[width=0.96\textwidth]{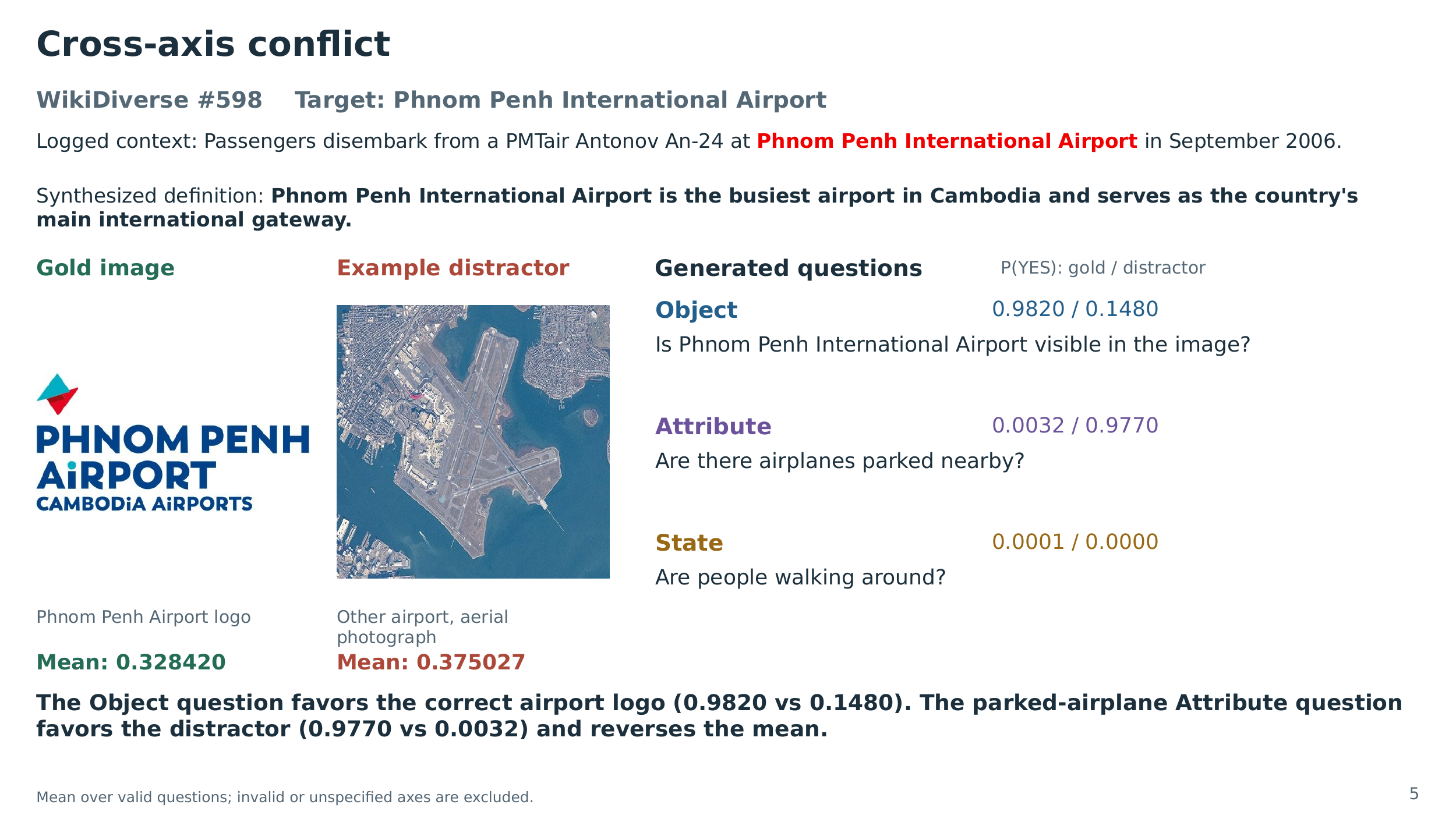}
  \caption{Cross-axis conflict for WikiDiverse \#598.  Target identity and
  airport-scene attributes favor different candidates.}
  \label{fig:cgr-cross-axis-airport}
\end{figure*}

VerSe \#956 yields an analogous conflict for \emph{fly}.  The State question
recognizes the airborne object in the gold image ($0.9948$ vs. $0.0001$), but
the Object question asks for a bird or animal and therefore favors the dog
distractor ($0.0000$ vs. $0.9975$).  The resulting means are $0.331745$ and
$0.332534$, respectively.

\begin{figure*}[t]
  \centering
  \includegraphics[width=0.96\textwidth]{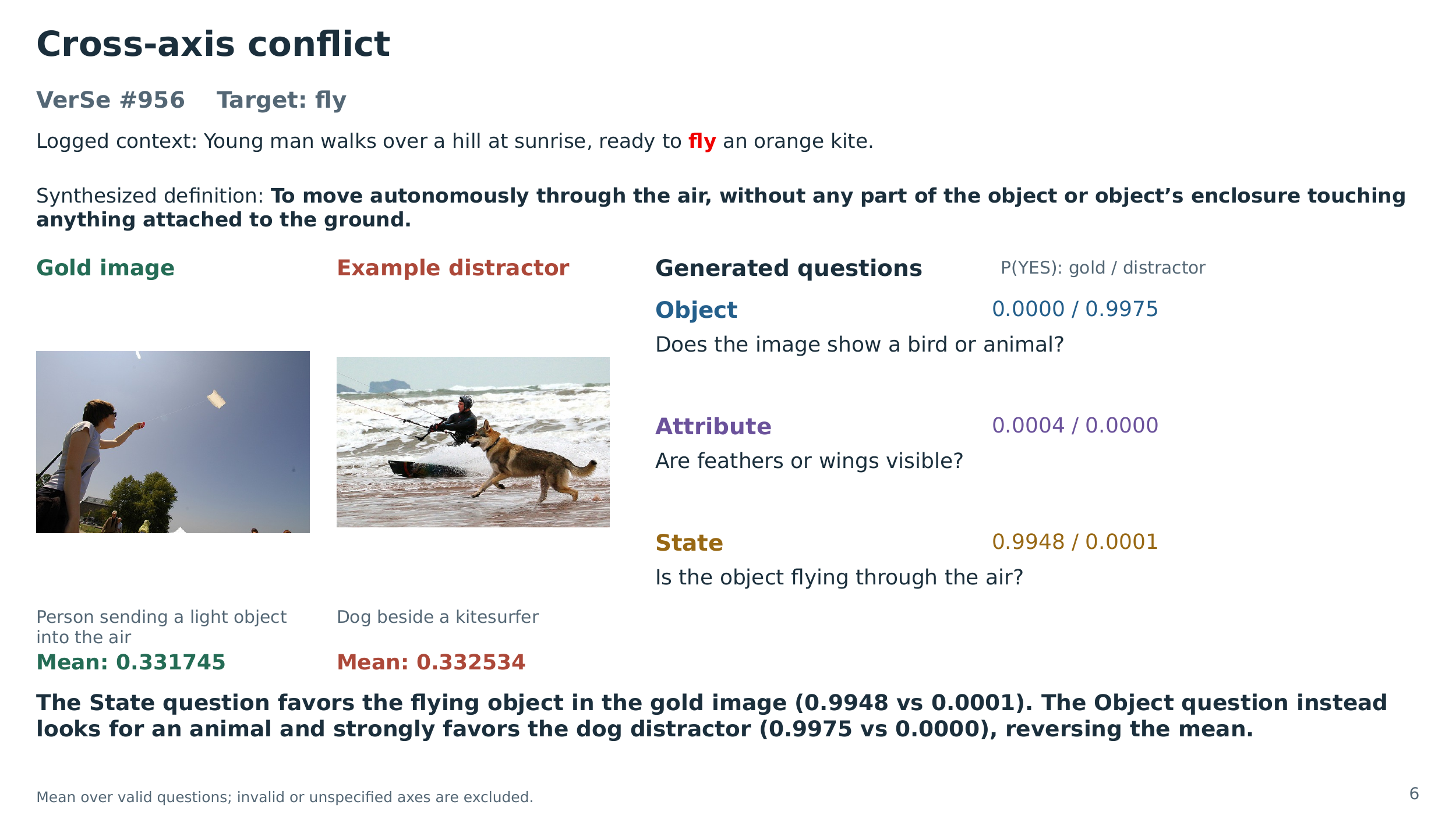}
  \caption{Cross-axis conflict for VerSe \#956 (\emph{fly}).  State evidence
  favors the gold image, whereas an animal-oriented Object constraint favors
  the distractor.}
  \label{fig:cgr-cross-axis-fly}
\end{figure*}

\clearpage

\subsection{Non-Discriminative Constraints}
\label{app:cgr-nondiscriminative}

These failures occur when the generated constraints describe properties shared
by the gold image and a distractor while omitting the evidence needed to
distinguish them.  For WikiDiverse \#217, both the U.S. and U.K. Libertarian
Party logos satisfy the generic party-sign question ($0.9968$ vs. $0.9981$).
The constraints omit the U.S. identity and the distinctive torch emblem, so
the valid-axis means are nearly tied ($0.498451$ vs. $0.499044$).

\begin{figure*}[t]
  \centering
  \includegraphics[width=0.96\textwidth]{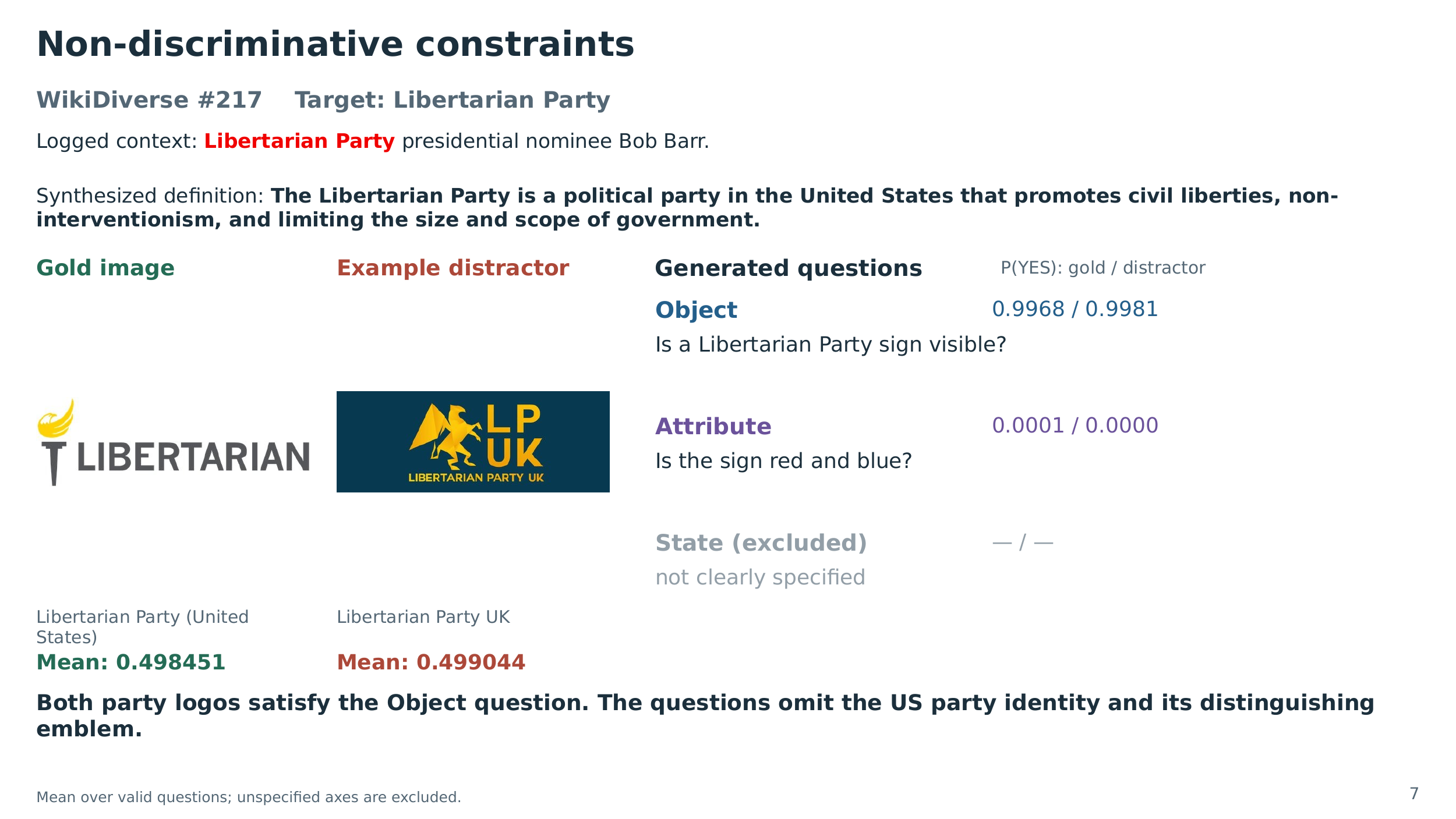}
  \caption{Non-discriminative constraints for WikiDiverse \#217.  Generic
  party-logo cues do not distinguish the U.S. Libertarian Party from its U.K.
  counterpart.}
  \label{fig:cgr-nondiscriminative-libertarian}
\end{figure*}

For WikiDiverse \#1002, both the Su-27 and F-16 satisfy the cues ``fighter jet''
and ``gray.''  The questions omit discriminative features such as engine count
and tail configuration, yielding means of $0.998850$ for the Su-27 and
$0.999100$ for the F-16 distractor.

\begin{figure*}[t]
  \centering
  \includegraphics[width=0.96\textwidth]{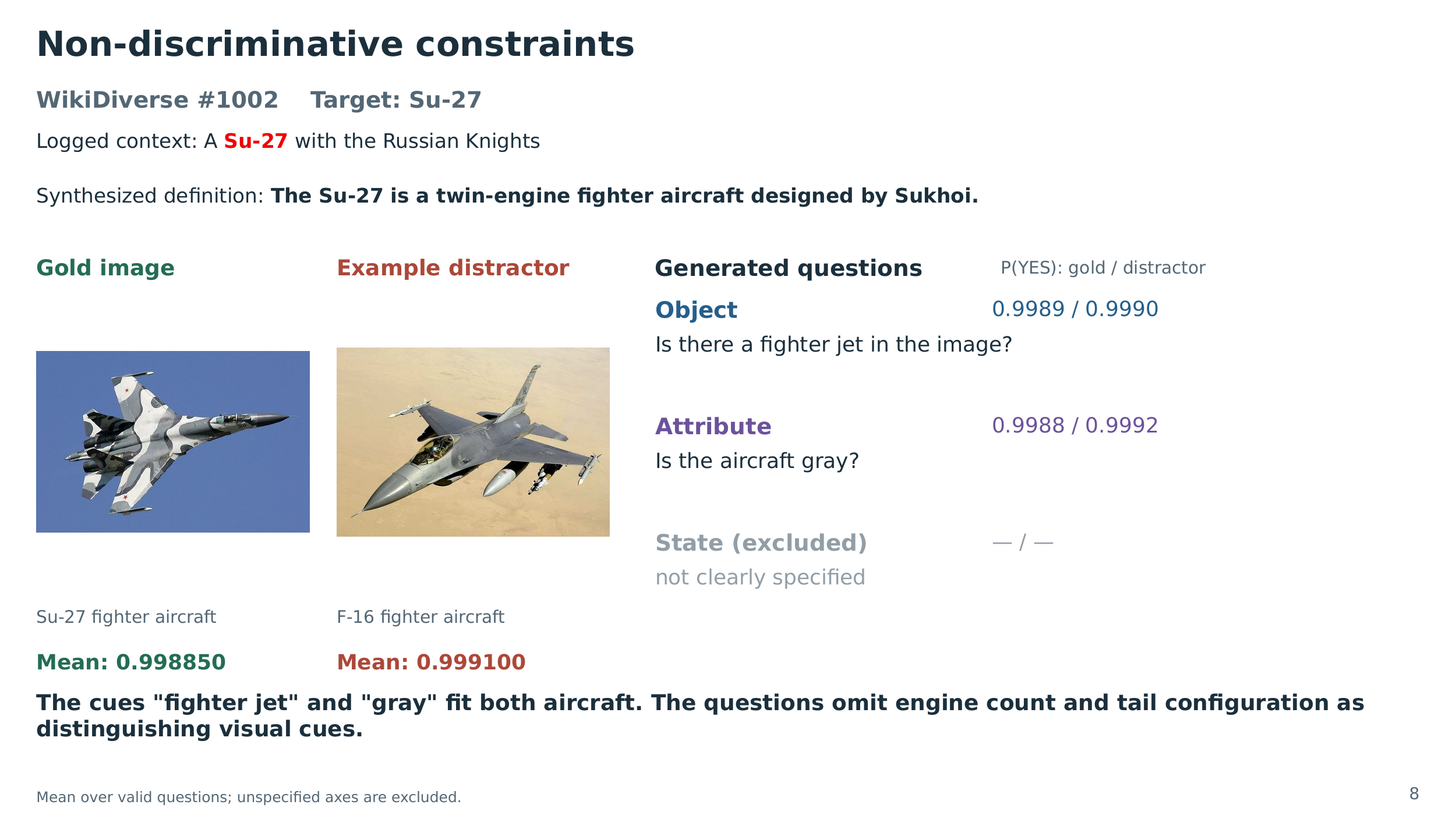}
  \caption{Non-discriminative constraints for WikiDiverse \#1002.  Broad
  fighter-aircraft cues fit both the Su-27 and F-16.}
  \label{fig:cgr-nondiscriminative-su27}
\end{figure*}

\clearpage

\subsection{Visual Verification Errors}
\label{app:cgr-verification}

Visual verification errors occur when a question is sufficiently specific,
but the VLM assigns a high probability despite contradictory visible evidence.
In VerSe \#494, the standing dog shows no visible biting or use of teeth, yet
the Object question receives $P(\mathrm{YES})=0.9047$ for the distractor and
accounts for nearly the entire score difference.

\begin{figure*}[t]
  \centering
  \includegraphics[width=0.96\textwidth]{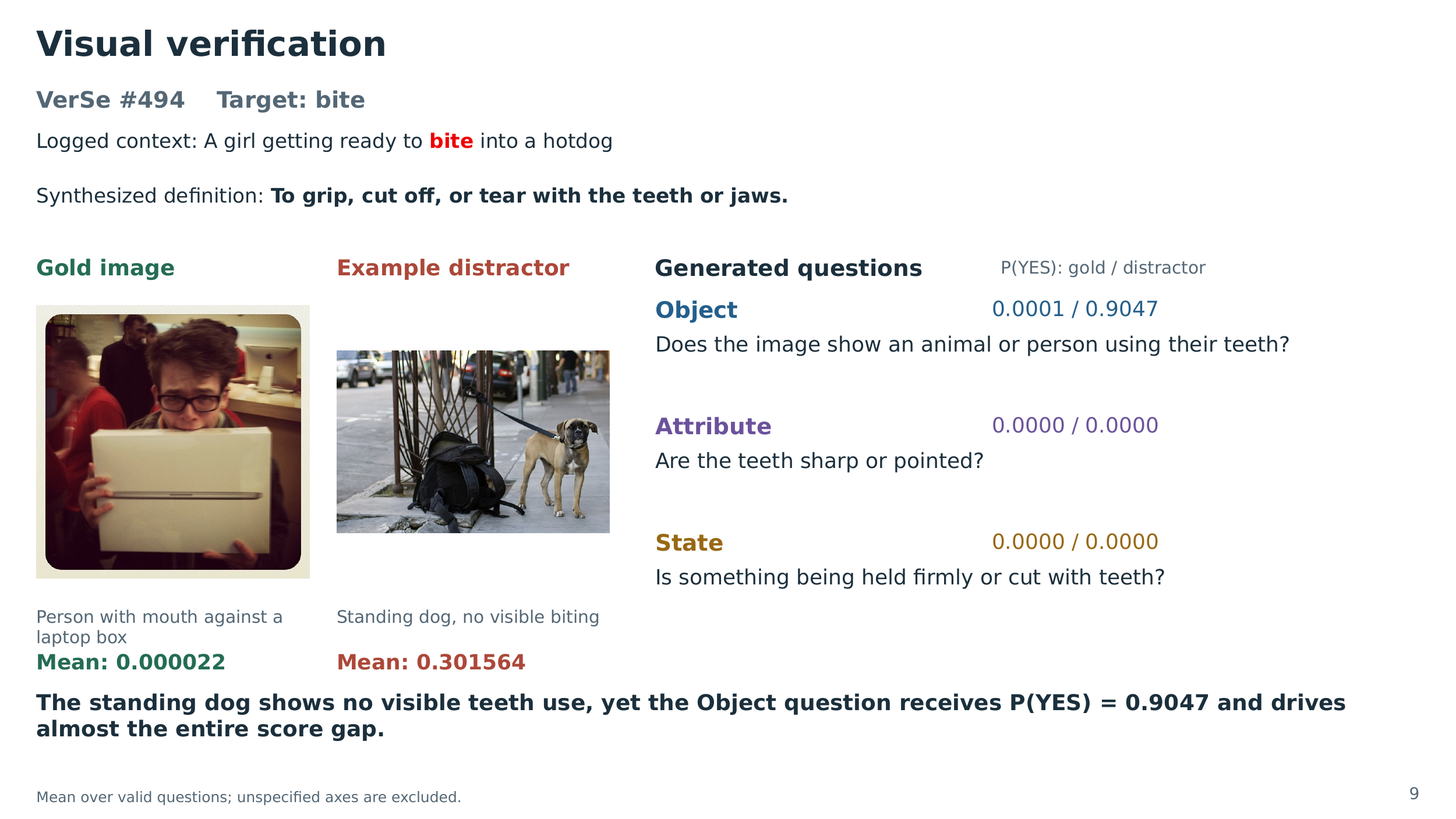}
  \caption{Visual verification error for VerSe \#494 (\emph{bite}).  The VLM
  affirms tooth use for a standing dog despite the absence of visible biting.}
  \label{fig:cgr-verification-bite}
\end{figure*}

WikiDiverse \#1013 provides a direct text-verification failure.  The distractor
sign visibly reads ``Moorgreen Hospital,'' but the question asking whether the
sign identifies Stobhill Hospital receives $P(\mathrm{YES})=0.9859$.  Together
with the Object score of $0.9933$, this produces a distractor mean of
$0.989622$ despite explicit contradictory signage.

\begin{figure*}[t]
  \centering
  \includegraphics[width=0.96\textwidth]{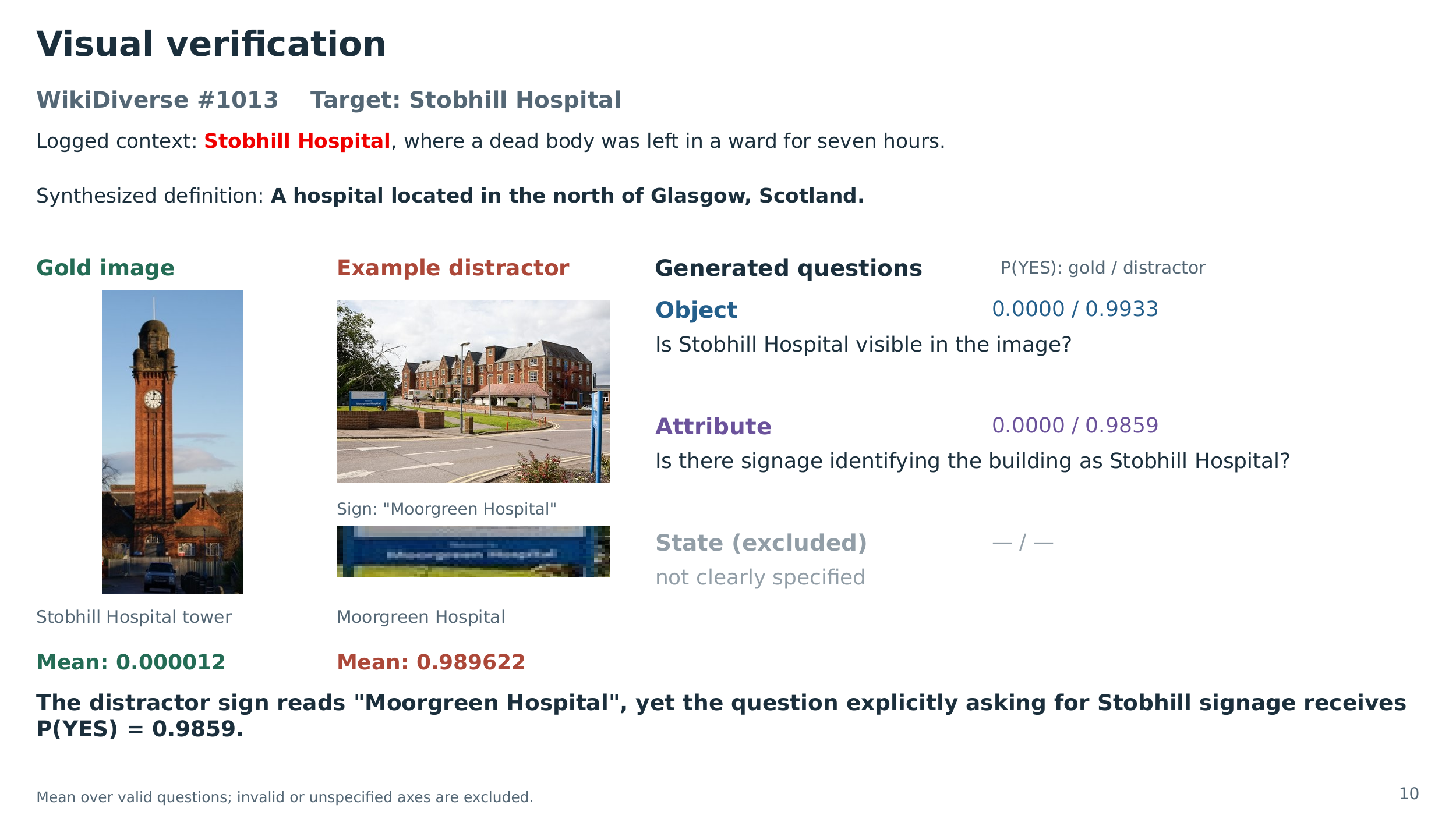}
  \caption{Visual verification error for WikiDiverse \#1013.  The model
  accepts the Stobhill constraint even though the visible sign names Moorgreen
  Hospital.}
  \label{fig:cgr-verification-stobhill}
\end{figure*}

\newpage

\section{Data Release Plan}
Upon publication, we will release the prepared benchmark package, including the fixed instances, ordered candidate lists, gold labels, benchmark filenames, source mappings, collection-time SHA-256 digests, and reconstruction and validation scripts. VerSe and VWSD images will be reconstructed from their official source packages. For WikiDiverse, the package provides the original source URLs and metadata needed to reconstruct the candidate images and detect missing files or checksum drift in upstream sources for licensing reasons. Because externally hosted images may be removed or modified over time, future reconstructions may not be pixel-identical to the experimental snapshot. Users should therefore report the reconstruction date and any missing, checksum-mismatched, substituted, or excluded images.

\section{Runtime Comparison}
\label{apx:runtime_comparison}

\begin{figure}[!h]
    \centering
    \includegraphics[width=.5\linewidth]{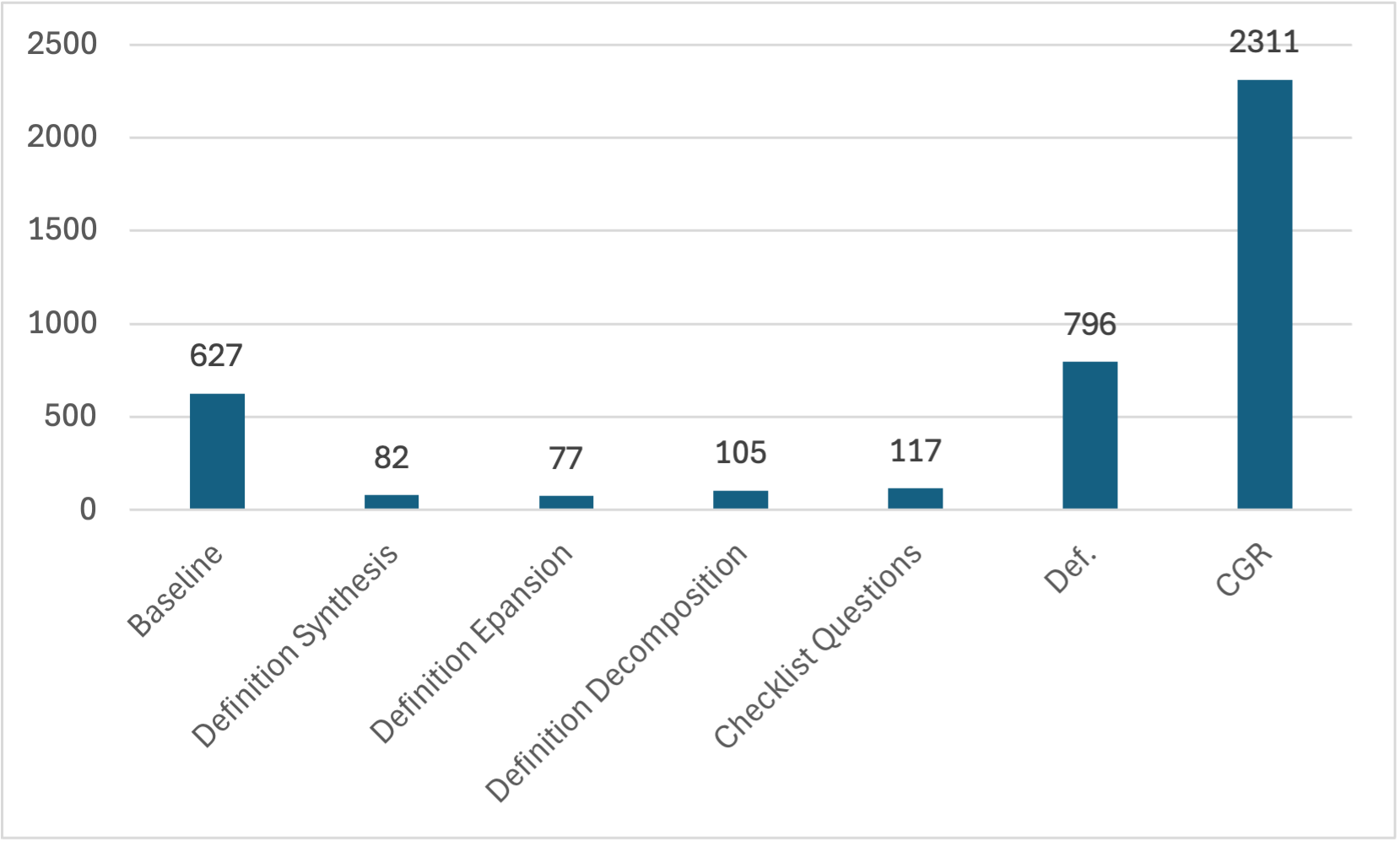}
    \caption{Pilot runtime profile on VWSD, measured in seconds.}
    \label{fig:runtime_profile}
\end{figure}

Figure~\ref{fig:runtime_profile} reports the pilot runtime breakdown on VWSD. Base pointwise scoring required 627 seconds, while the individual text-only stages required only 77--117 seconds each; the complete definition-based setting took 796 seconds. In contrast, CGR required 2,311 seconds, approximately 3.7× the baseline runtime, indicating that candidate-wise constraint evaluation accounts for most of GAR's computational overhead.

\end{document}